\documentclass{article}
\usepackage{arxiv}

\usepackage[utf8]{inputenc} 
\usepackage[T1]{fontenc} 
\usepackage{hyperref} 
\usepackage{url} 
\usepackage{booktabs} 
\usepackage{nicefrac} 
\usepackage{microtype} 
\usepackage{lipsum} 
\usepackage{natbib}
\usepackage{graphicx}
\usepackage{doi}
\usepackage{cancel}
\usepackage{multirow}
\usepackage{multicol}
\usepackage{enumitem}
\usepackage{placeins}
\usepackage{subcaption}
\usepackage{float}
\usepackage{wrapfig}
\usepackage[table]{xcolor}
\usepackage{algorithm}
\usepackage{algorithmicx}
\usepackage{algpseudocode}

\usepackage{amsmath,amsfonts,bm}

\def\eqref#1{equation~\ref{#1}}

\def\1{\bm{1}}

\DeclareMathAlphabet{\mathsfit}{\encodingdefault}{\sfdefault}{m}{sl}
\SetMathAlphabet{\mathsfit}{bold}{\encodingdefault}{\sfdefault}{bx}{n}

\DeclareMathOperator*{\argmax}{arg\,max}

\usepackage{amssymb}
\usepackage{mathtools}
\usepackage{amsthm}

\theoremstyle{plain}
\newtheorem{theorem}{Theorem}[section]

\theoremstyle{definition}
\newtheorem{definition}[theorem]{Definition}

\theoremstyle{remark}

\usepackage{titletoc}
\usepackage{tocloft}
\title{Concrete Subspace Learning Based Interference Elimination for Multi-Task Model Fusion}

\author{
  Anke Tang$^{1,\dagger}$,\ Xianglin Luo$^{2,\dagger}$,\ Li Shen$^{3}$,\ Yong Luo$^{1}$,\ Liang Ding$^{4}$,
  Han Hu$^{5}$,\ Bo Du$^{1}$,\ Dacheng Tao$^{6}$
  \\[0.6em]
  \begin{minipage}{0.9\textwidth}\raggedright\small
    $^{1}$School of Computer Science, Wuhan University, Wuhan 430072, China\\
    $^{2}$Chinese University of Hong Kong, Shenzhen 518172, China\\
    $^{3}$Sun Yat-sen University, Shenzhen 518107, China\\
    $^{4}$Alibaba Group, Hangzhou 310052, China\\
    $^{5}$School of Information and Electronics, Beijing Institute of Technology, Beijing 100081, China\\
    $^{6}$Nanyang Technological University, Singapore 639798\\[0.3em]
    $^{\dagger}$Equal contribution.
  \end{minipage}
}
\date{}
\renewcommand{\shorttitle}{Concrete Subspace Learning for Multi-Task Model Fusion}

\hypersetup{
  pdftitle={Concrete Subspace Learning Based Interference Elimination for Multi-Task Model Fusion},
  pdfauthor={Anke Tang, Xianglin Luo, Li Shen, Yong Luo, Liang Ding, Han Hu, Bo Du, Dacheng Tao},
  pdfkeywords={Model Fusion, Subspace Learning, Meta-Learning, Multi-Task Learning},
}

\begin{document}
\maketitle

\begin{abstract}
  Merging task-specific models via arithmetic operations on task vectors is a scalable strategy for multi-task learning.
  However, existing merging techniques typically resolve parameter conflicts by evaluating individual attributes like parameter magnitudes or signs, overlooking their collective impact on the overall functionality of the model.
  In this work, we propose the CONtinuous relaxation of disCRETE (Concrete) subspace learning method to identify a common low-dimensional subspace and utilize its shared information to tackle the interference problem without significantly sacrificing performance.
  Specifically, we model the problem as a bi-level optimization problem and introduce a meta-learning framework to find the Concrete subspace mask through gradient-based techniques.
  At the upper level, we focus on learning a shared Concrete mask to identify the subspace, while at the inner level, model merging is performed to maximize the performance of the merged model.
  We conduct extensive experiments in both vision and language domains, and the results demonstrate the effectiveness of our method. For example, our method yields an average accuracy improvement of up to 8.3\% compared to standard task arithmetic when merging eight ViT-B/32 models.
\end{abstract}


\section{Introduction}
\label{section:introduction}

Pre-trained large models are foundational to modern machine learning, adaptable to diverse tasks via fine-tuning.
Merging these specialized models into a unified one offers a scalable way to extract knowledge without accessing original training data~\cite{liDeepModelFusion2023:b,zhengLearningModelsFinetuning2025,Survery_ModelMerging_2024}.

Multi-task model merging becomes valuable when original training data is unavailable, despite the availability of fine-tuned models\cite{wuHeterogeneousModelReuse2019}.
Following related research, exemplified by task arithmetic~\cite{ilharcoEditingModelsTask2023:b}, a multitude of influential techniques have been introduced to edit pre-trained models and incorporate task-specific models.
These techniques leverage the knowledge captured within fine-tuned models without the need for extensive retraining or access to the original training data.

A notable challenge for multi-task model fusion is resolving task interference, a phenomenon that occurs when interactions among the parameters from task-specific models negatively impact the overall performance.
Existing methods attempt to mitigate interference through various heuristics.
For instance, Ties-Merging merges task vectors using magnitude trim, elect sign, and disjoint merge strategies instead of simple addition~\cite{yadavResolvingInterferenceWhen2023:b}.
Yu et al.~\cite{yuLanguageModelsAre2024} randomly drop parameters and rescale the remaining ones.
However, these techniques typically address interference by evaluating parameters in isolation, while often overlooking their collective impact on the overall functionality.

\begin{figure}[t]
  \begin{center}
    \includegraphics[width=\linewidth]{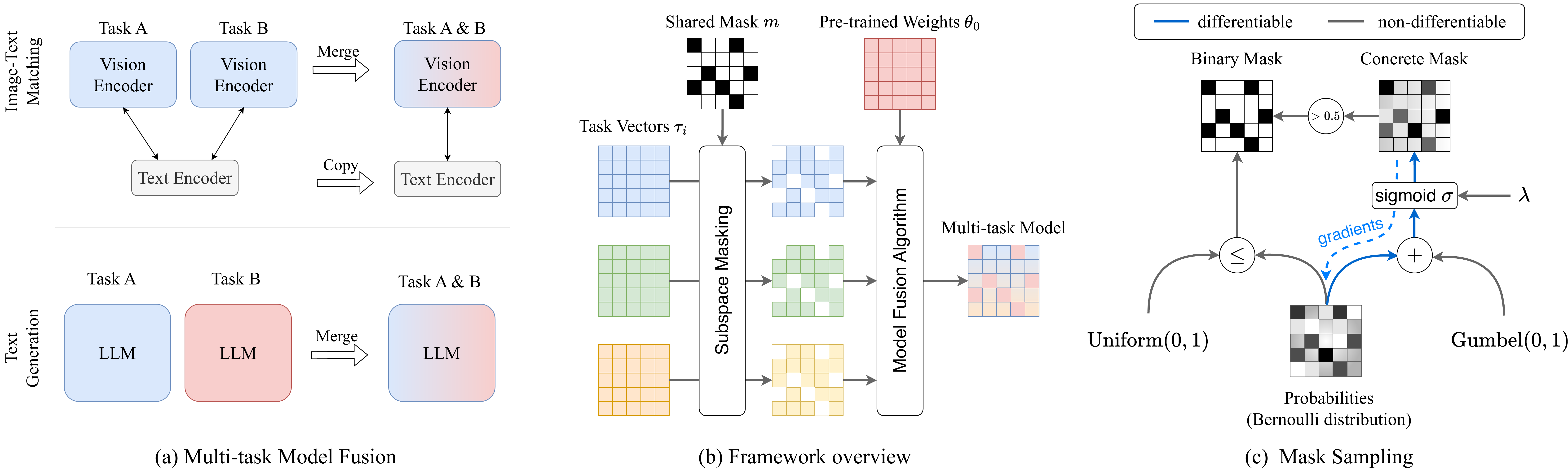}
    \caption{
      (a) \textbf{Multi-task model fusion}. Different types of expert models are merged into a unified model.
      (b) \textbf{Framework overview}. Our proposed framework comprises two main steps: first, establishing a shared subspace across various task vectors, and second, merging the models within this shared subspace.
      (c) \textbf{Mask sampling}. Here we illustrate the procedure for sampling discrete binary masks and our differentiable Concrete mask. While a Concrete mask can also be binarized, this binarization process is non-differentiable.
    }
    \label{fig:overview}
  \end{center}
\end{figure}

We introduce the CONtinuous relaxation of disCRETE (\textit{Concrete}) subspace learning method to identify a common low-dimensional subspace of the model parameter space and effectively leverage the shared information within it to address the issue of task interference, while minimizing the impact on overall performance, as shown in Figure~\ref{fig:overview}.
In addition, we present enhanced variations of Task Arithmetic, AdaMerging~\cite{yangAdaMergingAdaptiveModel2024}, and TSV-M~\cite{gargiuloTaskSingularVectors2025}, which we have termed Concrete Task Arithmetic, Concrete AdaMerging, and Concrete TSV-M, respectively.

Specifically, we model the problem as a bi-level optimization problem and introduce a meta-learning framework to find the Concrete subspace mask through gradient-based techniques.
At the upper level, we focus on learning a shared Concrete mask to identify the subspace, while at the inner level, an existing model merging algorithm is performed to maximize the performance of the merged model.

We conduct a range of experiments in both vision and language domains, including scenarios involving fully fine-tuned and LoRA fine-tuned encoder-decoder Transformer models, as well as experiments on multi-task model fusion and out-of-distribution generalization.

To summarize, our contributions are as follows:
\begin{itemize}[itemsep=0pt,parsep=0pt,topsep=0pt]
  \item We propose the Concrete subspace learning method to identify a common subspace of the model parameter space and effectively leverage the shared information within it to eliminate task interference while minimizing the negative impact on performance.
  \item We model the problem as a bi-level optimization problem and introduce a meta-learning framework to find the Concrete mask through gradient-based techniques.
  \item We introduce variations of three multi-task model fusion techniques known as Task Arithmetic, AdaMerging, and TSV-M.
  \item We conduct experiments in both vision and language domains, including fully fine-tuned and LoRA fine-tuned language models. Results demonstrate our method's effectiveness.
\end{itemize}

\section{Related Work}
\label{section:related_work}

\textbf{Multi-task model fusion} is an effective and scalable approach that combines knowledge from multiple task-specific models into a single unified model.
Here we discuss merging techniques into two main categories:
(1) \textit{Weight interpolation} is the most straightforward method for multi-task model fusion.
This technique merges multiple models into a unified model without requiring exhaustive computations or prior optimization.
Aggregation can be performed directly or within a subspace.
Approaches such as Model Soups~\cite{wortsmanModelSoupsAveraging2022}, Task Arithmetic~\cite{ilharcoEditingModelsTask2023:b}, Stochastic Weight Averaging~\cite{kaddourStopWastingMy2022:b,izmailovAveragingWeightsLeads2019:b}, Fisher Merging~\cite{matenaMergingModelsFisherWeighted2022}, Ties-Merging~\cite{yadavResolvingInterferenceWhen2023:b}, and TSV-M~\cite{gargiuloTaskSingularVectors2025} have been shown to enhance performance effectively.
However, these methods often assume that the models being merged share similar architectures and parameter distributions.
(2) \textit{Alignment-based methods}~\cite{liConvergentLearningDifferent2016:b,tatroOptimizingModeConnectivity2020:a} involve matching and averaging units across different models. This process narrows the mathematical distances, such as Euclidean distances, between models, harmonizing them for more effective fusion\cite{georgestoicaZipItMergingModels2023:b,jinDatalessKnowledgeFusion2023}.
These include methods such as activation matching and weight matching, extending to concepts like Re-basin~\cite{ainsworthGitReBasinMerging2023:b}, which utilizes permutation invariance to bring solutions into a single low-loss basin.

\textbf{Multi-task model fusion under PEFT setting.}
As models scale, full fine-tuning becomes costly, leading to the popularity of parameter-efficient fine-tuning methods.
Some approaches apply weight interpolation on the parameter space of adapter modules~\cite{chronopoulouAdapterSoupWeightAveraging2023,huangLoraHubEfficientCrossTask2024,zhangComposingParameterEfficientModules2023,dimitriadisParetoLowRankAdapters2024,prabhakarLoRASoupsMerging2025}.
Wu et al.~\cite{wuPiTuningTransferring2023:b} compute task similarity based on the Fisher information matrix and select adapter modules for interpolation.
Others improve the performance of model fusion by partially linearizing the PEFT modules~\cite{tangParameterEfficientMultitask2024}.

Unlike previous research, our study specifically addresses the problem of task interference.
Our proposed method acts as \textit{a highly adaptable plug-in} for existing model fusion techniques and can be applied to both fully fine-tuned models and parameter-efficiently fine-tuned models.

\section{Methodology}
\label{section:methodology}

This section introduces the problem setting and notation, then presents Concrete subspace learning for meta-learning a shared subspace across tasks and test-time adaptation.

\subsection{Preliminary}
\label{subsec:preliminary}

We start with a pre-trained neural network $f(\cdot, \theta_0)$ parameterized by $\theta_0 \in \mathbb{R}^d$, trained on an extensive dataset, and consider a set of $T$ downstream tasks $\mathcal{S} = \{s_i\}_{i=1}^T$.
Subsequently, the pre-trained model undergoes fine-tuning on each task $s_i$ to obtain a fine-tuned model parameterized by $\theta_i$.
The task vectors are $\tau_i = \theta_i - \theta_0$, encapsulating task-specific information.

Given access to both the pre-trained and fine-tuned model weights, our objective is to merge these models into a unified model capable of performing effectively on all tasks within $\mathcal{S}$, parameterized by $\theta^{\text{merged}}$:
\begin{equation}
  \theta^{\text{merged}}=\mathcal{A}(\theta_0, \{\tau_i\}_{i=1}^T,w_{\mathcal{A}}),
\end{equation}
where $\mathcal{A}$ is a fusion algorithm, and $w_{\mathcal{A}}$ represents $\mathcal{A}$'s algorithmic parameters.
This is illustrated in Figure~\ref{fig:overview}(a).
For example, one such algorithm is task-wise AdaMerging~\cite{yangAdaMergingAdaptiveModel2024}, which performs a task-wise linear combination of task vectors. Mathematically, this can be expressed as $\theta^{\text{merged}} = \theta_0 + \sum_{i=1}^T w_i \tau_i$, where $w_{\mathcal{A}}=\{w_i\}_{i=1}^T$ represents the weights assigned to each task vector $\tau_i$.

In our study, we propose a subspace learning-based method to enhance existing model fusion algorithms, designed to be compatible with them, which we refer to as Concrete subspace learning.
By incorporating Concrete subspace learning, we can mitigate task interference and improve the multi-task performance of the merged model.

\subsection{Concrete Subspace Learning}
\label{subsec:concrete_subspace_learning}

To address the aforementioned problem, we propose the Concrete (CONtinuous relaxation of disCRETE) subspace learning method to identify a common subspace across tasks.
We represent the shared subspace using a shared mask $m$, sampled from a Concrete mask $\mathbf{m}$, and denote the task vector after masking and rescaling as $\mathcal{M}(\tau, m)$, where $\mathcal{M}:\mathbb{R}^d\times\mathbb{R}^d\mapsto\mathbb{R}^d$ is the masking process. We describe the Concrete mask in detail in Section~\ref{section:the_concrete_masking_process}, including the sampling process and the masking process.
By applying the masking process $\mathcal{M}(\cdot, m)$ to the task vectors, the method effectively filters out task-specific noise and variations, focusing on the shared information and structure to tackle the interference problem. Subsequently, we merge the models within this shared subspace to obtain the merged model parameterized by $\mathcal{A}(\theta_0, \{\mathcal{M}(\tau_j, m)\}_{j=1}^T; w_\mathcal{A})$. Figure~\ref{fig:overview}(b) illustrates the framework of our proposed method. 

\textbf{Theoretical motivation.}
Existing methods often overlook the detailed interaction between task vectors in the parameter space, simply assuming that linear combinations will yield effective solutions. To strictly analyze this, we approximate the behavior of the merged model using the first-order Taylor expansion around the pre-trained weights $\theta_0$.
For a given input $x$, the output of the merged model $f(x, \theta^{\text{merged}})$ can be approximated as:
\begin{small}
  \begin{align}
    f(x,\theta^{\text{merged}}) & \approx f(x,\theta_0) + \nabla_{\theta}f(x,\theta_0)^\top (\theta^{\text{merged}} - \theta_0).
  \end{align}
\end{small}
Consider Task Arithmetic as a baseline where $\theta^{\text{merged}} = \theta_0 + \sum_{i=1}^T \tau_i$ (simplified with $w=1$). The change in output is driven by the term $\nabla_{\theta}f(x,\theta_0)^\top (\sum_{i=1}^T \tau_i) = \sum_{k=1}^d \frac{\partial f}{\partial \theta_k} (\sum_{i=1}^T \tau_{i,k})$.
Here, interference occurs at the parameter level. For a specific dimension $k$, if task vectors have conflicting signs (e.g., $\text{sgn}(\tau_{i,k}) \neq \text{sgn}(\tau_{j,k})$), their updates cancel each other out, leading to \emph{destructive interference}. This cancellation results in a merged parameter update that is diluted or neutralized, creating a model that is optimal for neither task. Conversely, if signs align, they result in \emph{constructive interference}, reinforcing the shared knowledge.
Simple summation or scalar re-weighting cannot resolve these dimensional conflicts. The proposed Concrete subspace learning introduces a flexible, dimension-wise mask $m \in [0, 1]^d$. The fusion becomes $\theta^{\text{merged}} = \theta_0 + m \odot (\sum_{i=1}^T \tau_i)$.
By optimizing $m$ to maximize multi-task performance (as formulated in Eq.~(\ref{eq:upper_optim})), the mask acts as a learnable feature selector that:
(1) Retains dimensions where tasks exhibit constructive interference (shared capabilities);
(2) Preserves dimensions critical for specific tasks if they do not strongly conflict with others (orthogonal capabilities); and
(3) Suppresses dimensions with high variance or conflicting directions across tasks (noise and interference pruning).
Thus, learning the mask is justified as finding an optimal projection into a subspace that minimizes the inner product loss caused by update conflicts, ensuring the merged model effectively navigates the trade-off landscape of multi-task optimization.
More explicitly, under mild conditions where task vectors exhibit orthogonality in non-conflicting dimensions, the interference term induced by a Concrete mask can be summarized as
\begin{small}
  \begin{align}
    \mathcal{I}(m)
    = \sum_{i \neq j}
    \left\langle
    \nabla_\theta f(x,\theta_0)^\top(\tau_i \circ m),\,
    \tau_j \circ m
    \right\rangle .
  \end{align}
\end{small}
The optimal mask $m^*$ that minimizes $\mathcal{I}(m)$ suppresses dimensions where task vectors conflict, while retaining dimensions where task vectors are mutually compatible or orthogonal to each other.

\textbf{Subspace learning as a bi-level optimization problem.}
Since the shared mask $m$ should benefit multiple tasks for arbitrary model fusion algorithms, we model identifying this common low-dimensional subspace as a bi-level optimization problem and design an algorithm to meta-learn the shared mask.
We can formulate the problem as follows:
\begin{align}
  \max_{m} \frac{1}{T} & \sum_{i=1}^{T} \mathcal{P}_i \left(\mathcal{A}(\theta_0, \{\mathcal{M}(\tau_j, m)\}_{j=1}^T; w^*), s_i \right) \label{eq:upper_optim} \\
  \text{s.t.}\, w^*    & = {\arg\min}_{w_{\mathcal{A}}}                                                                                          \nonumber     \\
  d                    & \left(\theta^{MTL},\mathcal{A} (\theta_0 , \{\mathcal{M}(\tau_j, m)\}_{j=1}^T; {w_{\mathcal{A}}}  )\right), \label{eq:lower_optim}
\end{align}
where $\mathcal{P}_i$ is the task-specific performance metric, $d$ is a distance metric between the merged model and the multi-task model $\theta^{MTL}$, which serves as an upper bound baseline of the performance.
The upper-level objective is to find a shared mask $m$ that maximizes the performance of the merged model $\mathcal{A}(\theta_0, \{\mathcal{M}(\tau_j, m)\}_{j=1}^T, w)$ across all tasks. The objective at the inner level is to find the optimal algorithmic parameters $w$ that maximize the performance of the merged model on each task.

It is worth noting that the process of obtaining $w^*$ depends on the specific model fusion algorithm $\mathcal{A}$ used, which is transparent to our approach.
Instead, our study focuses on optimizing the upper-level objective function, as defined in Eq.(\ref{eq:upper_optim}), without concern for the implementation details of Eq.(\ref{eq:lower_optim}).
In extreme cases, $w^*$ may be determined as a hyperparameter, enabling the inner optimization problem to be bypassed through hyperparameter tuning on the validation set, rather than iterative optimization, which remains conceptually aligned with our framework.

\begin{algorithm}[t]
  \caption{Meta-learn the Concrete subspace}
  \label{alg:meta-learn_mask}
  \small
  \begin{algorithmic}
    \State {\bfseries Input:} pre-trained weights $\theta_0 \in \mathbb{R}^d$, task vectors $\mathcal{T} = \{\tau\}_{i=1}^T$, tasks $\mathcal{S} = \{s_i\}_{i=1}^T$, learning rates $\alpha^{L}$ and $\beta$.
    \\ \vskip -6pt \hrulefill
    \State $\psi\gets 0$ \Comment{mask initialization}
    \Repeat
    \State sample $m \sim \text{Concrete}(\psi, \lambda)$ \Comment{sample mask from logits}
    \State $\mathcal{T}' \leftarrow \left\{ \tau'_i \mid \tau'_i = \mathcal{M}(\tau_i, m) \right\}_{j=1}^T$
    \If{$w_\mathcal{A}$ is iteratively optimizable}
    \State $w_\mathcal{A} \gets w_0$ \Comment{parameter initialization}
    \State sample a batch of $N$ unlabeled data $\mathcal{D}$
    \State $w^* \leftarrow w_{\mathcal{A}} - \alpha^{L} \nabla_{w_{\mathcal{A}}} \mathcal{L}_{\mathcal{A}}(\mathcal{A}(\theta_0, \mathcal{T}'; w); \mathcal{D})$
    \Else
    \State $w^* \gets w_0$ \Comment{hyperparameter selection}
    \EndIf
    \State $\theta^{\text{merged}} \leftarrow \mathcal{A}(\theta_0, \mathcal{T}'; w^*)$
    \State sample a batch of $N$ unlabeled data $\mathcal{D}$
    \State $\psi \leftarrow \psi - \beta \nabla_{\psi}\mathcal{L}_{\text{entropy}}(\theta^{\text{merged}};\mathcal{D})$
    \Until{convergence}
    \State {\bfseries Return:} Concrete mask $\mathbf{m}$ parameterized by $\psi$.
  \end{algorithmic}
\end{algorithm}

Algorithm~\ref{alg:meta-learn_mask} describes the meta-learning procedure for learning the shared mask.
Specifically, it iteratively refines a Concrete mask parameterized by its logits $\psi$.
Within the inner loop, we first optimize the algorithmic parameters $w_{\mathcal{A}}$ through single-step test-time adaptation training (e.g., weight parameters of AdaMerging).
Here, we use a large learning rate to ensure that $w_{\mathcal{A}}$ is updated quickly.
Alternatively, we can bypass this step by selecting hyperparameters based on performance on the validation set (e.g., Task Arithmetic or Ties-Merging).
We then update the merged model weights to $\mathcal{A}(\theta_0, \{\mathcal{M}(\tau_j, m)\}_{j=1}^T; w)$.
Then, in the outer loop, we optimize the logits $\psi$ of the Concrete mask $m$.
Our approach functions as a versatile plug-in method, seamlessly integrable with existing model fusion techniques.
For more details on combining our method with specific model fusion algorithms, please refer to Appendix~\ref{appendix:algorithm_details}.

\begin{table}[htb]
  \caption{Multi-task performance when merging CLIP-ViT-B/32 and CLIP-ViT-L/14 models on all eight tasks.}
  \label{table:multi-task_performance_clip-vit-b-32}
  \resizebox{\linewidth}{!}{%
    \centering
    \begin{tabular}{lccccccccc}
      \textit{CLIP-ViT-B/32}  &                                                                                                                                                          \\
      \toprule
      \textbf{Method}         & \textbf{SUN397} & \textbf{Cars} & \textbf{RESISC45} & \textbf{EuroSAT} & \textbf{SVHN} & \textbf{GTSRB} & \textbf{MNIST} & \textbf{DTD}  & \textbf{Avg.} \\
      \midrule
      Pre-trained             & 63.2            & 59.6          & 60.2              & 45.0             & 31.6          & 32.6           & 48.3           & 44.4          & 48.1          \\
      Individual              & \textbf{75.3}   & \textbf{77.7} & \textbf{96.1}     & \textbf{99.9}    & \textbf{97.5} & 98.7           & \textbf{99.7}  & \textbf{79.4} & \textbf{90.5} \\
      Traditional MTL         & 73.9            & 74.4          & 93.9              & 98.2             & 95.8          & \textbf{98.9}  & 99.5           & 77.9          & 88.9          \\
      \midrule
      Weight Averaging        & 65.3            & 63.3          & 71.4              & 73.6             & 64.2          & 52.8           & 87.5           & 50.1          & 66.0          \\
      Fisher Merging          & 68.6            & 69.2          & 70.7              & 66.4             & 72.9          & 51.1           & 87.9           & 59.9          & 68.3          \\
      RegMean                 & 65.3            & 63.5          & 75.6              & 78.6             & 78.1          & 67.4           & 93.7           & 52.0          & 71.8          \\
      \midrule
      \multicolumn{10}{c}{\textit{Task Arithmetic (TA)-Based}}                                                                                                                           \\
      Task Arithmetic         & 55.3            & 54.9          & 66.7              & 77.4             & 80.2          & 69.7           & 97.3           & 50.1          & 69.0          \\
      Ties-Merging            & \textbf{65.0}   & \textbf{64.3} & 74.7              & 76.8             & 81.3          & 69.4           & 96.5           & \textbf{54.3} & 72.8          \\
      \textbf{Concrete TA}    & 62.5            & 61.1          & \textbf{76.0}     & \textbf{95.7}    & \textbf{91.0} & \textbf{81.9}  & \textbf{98.5}  & 51.9          & \textbf{77.3} \\
      \midrule
      \multicolumn{10}{c}{\textit{Task-wise AdaMerging (TW AM)-Based}}                                                                                                                   \\
      TW AM                   & 58.3            & 53.2          & 71.8              & 80.1             & 81.6          & 84.4           & 93.4           & 42.7          & 70.7          \\
      TW AM++                 & 60.8            & 56.9          & 73.1              & 83.4             & 87.3          & 82.4           & 95.7           & \textbf{50.1} & 73.7          \\
      \textbf{TW Concrete AM} & \textbf{62.7}   & \textbf{58.9} & \textbf{74.5}     & \textbf{94.8}    & \textbf{91.1} & \textbf{95.0}  & \textbf{98.1}  & 34.6          & \textbf{76.2} \\
      \midrule
      \multicolumn{10}{c}{\textit{Layer-wise AdaMergin (LW AM)-Based}}                                                                                                                   \\
      LW AM                   & 64.2            & 69.5          & 82.4              & 92.5             & 86.5          & 93.7           & 97.7           & 61.1          & 80.9          \\
      LW AM++                 & 66.6            & 68.3          & 82.2              & 94.2             & 89.6          & 89.0           & 98.3           & 60.6          & 81.1          \\
      \textbf{LW Concrete AM} & \textbf{67.8}   & \textbf{70.0} & \textbf{87.5}     & \textbf{96.0}    & \textbf{91.6} & \textbf{96.7}  & \textbf{98.7}  & \textbf{63.8} & \textbf{84.0} \\
      \midrule
      \multicolumn{10}{c}{\textit{Task Singular Vector Merging (TSV-M)-Based}}                                                                                                           \\
      TSV-M                   & 64.6            & 68.7          & 80.5              & 91.1             & 90.9          & 90.6           & 99.1           & 64.5          & 81.2          \\
      \textbf{Concrete TSV-M} & \textbf{67.3}   & \textbf{71.1} & \textbf{88.2}     & \textbf{98.0}    & \textbf{94.7} & \textbf{95.0}  & \textbf{99.2}  & \textbf{64.9} & \textbf{84.8} \\
      \bottomrule
    \end{tabular}
  }
  \vskip 12pt
  \resizebox{\linewidth}{!}{%
    \centering
    \begin{tabular}{lccccccccc}
      \textit{CLIP-ViT-L/14}  &                                                                                                                                                          \\
      \toprule
      \textbf{Method}         & \textbf{SUN397} & \textbf{Cars} & \textbf{RESISC45} & \textbf{EuroSAT} & \textbf{SVHN} & \textbf{GTSRB} & \textbf{MNIST} & \textbf{DTD}  & \textbf{Avg.} \\
      \midrule
      Pre-trained             & 68.2            & 77.9          & 71.3              & 61.3             & 58.4          & 50.6           & 76.4           & 55.4          & 64.9          \\
      Individual              & \textbf{82.3}   & \textbf{92.4} & \textbf{97.4}     & \textbf{99.9}    & \textbf{98.1} & \textbf{99.2}  & \textbf{99.7}  & 84.1          & \textbf{94.1} \\
      Traditional MTL         & 80.8            & 90.6          & 96.3              & 96.3             & 97.6          & 99.1           & 99.6           & \textbf{84.4} & 93.5          \\
      \midrule
      Weight Averaging        & 72.1            & 81.6          & 82.6              & 91.4             & 78.2          & 70.6           & 97.0           & 62.8          & 79.5          \\
      Fisher Merging          & 69.2            & 88.6          & 87.5              & 93.5             & 80.6          & 74.8           & 93.3           & 70.0          & 82.2          \\
      RegMean                 & 73.3            & 81.8          & 86.1              & 97.0             & 88.0          & 84.2           & 98.5           & 60.8          & 83.7          \\
      \midrule
      \multicolumn{10}{c}{\textit{Task Arithmetic (TA)-Based}}                                                                                                                           \\
      Task Arithmetic         & 74.1            & 82.1          & 86.7              & 92.6             & 87.9          & 86.8           & 98.9           & 65.6          & 84.4          \\
      Ties-Merging            & \textbf{75.0}   & 84.5          & 88.0              & 94.3             & 85.7          & 82.1           & 98.7           & \textbf{67.7} & 84.5          \\
      \textbf{Concrete TA}    & 74.6            & \textbf{86.2} & \textbf{89.0}     & \textbf{96.7}    & \textbf{93.6} & \textbf{93.4}  & \textbf{99.1}  & 66.9          & \textbf{87.4} \\
      \midrule
      \multicolumn{10}{c}{\textit{Layer-wise AdaMergin (LW AM)-Based}}                                                                                                                   \\
      LW AM                   & 79.0            & 90.3          & 90.8              & 96.2             & 93.4          & \textbf{98.0}  & 99.0           & \textbf{79.9} & 90.8          \\
      LW AM++                 & \textbf{79.4}   & 90.3          & 91.6              & \textbf{97.4}    & 93.4          & 97.5           & 99.0           & 79.2          & 91.0          \\
      \textbf{LW Concrete AM} & 77.8            & \textbf{91.2} & \textbf{92.1}     & 97.0             & \textbf{94.4} & 97.9           & 99.0           & 79.5          & \textbf{91.1} \\
      \midrule
      \multicolumn{10}{c}{\textit{Task Singular Vector Merging (TSV-M)-Based}}                                                                                                           \\
      TSV-M                   & 78.3            & 89.9          & 94.0              & 98.5             & 95.3          & 96.2           & 99.5           & 79.1          & 91.3          \\
      \textbf{Concrete TSV-M} & \textbf{78.6}   & \textbf{91.2} & \textbf{94.8}     & \textbf{98.9}    & \textbf{96.5} & \textbf{97.9}  & \textbf{99.5}  & \textbf{80.3} & \textbf{92.2} \\
      \bottomrule
    \end{tabular}
  }
\end{table}

\textbf{Test-time adaptation training.}
We consider a scenario where the target dataset contains only unlabeled test data.
Therefore, to maximize Eq.(\ref{eq:upper_optim}), we use a test-time adaptation training approach~\cite{mounsavengBagTricksFully2024,liangComprehensiveSurveyTestTime2023}.
The objective is to enhance the model's predictive accuracy and its capacity to generalize to unseen, unlabeled examples during the testing phase.

Take classification tasks as an example.
We can use the merged model to make predictions on the unlabeled test data, and then use these predictions to optimize the merged model.
Specifically, we minimize the Shannon entropy of the predictions as a proxy objective to encourage the model to make confident predictions.
\begin{align}
  \label{eq:entropy_loss}
  \mathcal{L}_{\text{entropy}}
   & = \mathbb{E}_{x\sim s_i^{\text{test}},s_i\sim \mathcal{S}}[-p(\hat{y}\mid x)\log p(\hat{y}\mid x)] \\
   & \approx -\frac{1}{N}\sum_{i=1}^N \sum_{c=1}^C p(\hat{y}_c\mid x_i) \log p(\hat{y}_c\mid x_i),
\end{align}
where we sample a batch of unlabeled data from the test set of each task $s_i$, $N=BT$ is the number of unlabeled instances, $B$ is the batch size, $T$ is the number of tasks, $x_i$ is the $i$-th unlabeled instance, $p(\hat{y}_c\mid x_i)$ is the predicted probability of the $c$-th class, and $C$ is the number of classes.

\section{Experiments}
\label{section:experiments}


\textbf{Models and Tasks.}
We utilize CLIP~\cite{radfordLearningTransferableVisual2021} for image-text matching and Flan-T5~\cite{chungScalingInstructionFinetunedLanguage2024} for text generation.
Refer to Appendix~\ref{appendix:model_fine_tuning_details} for details.

\begin{table}[tb]
  \caption{Requirements of different model merging methods.}
  \label{table:model_merging_methods}
  \resizebox{\linewidth}{!}{%
    \centering
    \begin{tabular}{lccc}
      \toprule
      \textbf{Method}                   & \textbf{Labeled training data}             & \textbf{Validation data}                   & \textbf{Test-time adaptation} \\ \midrule
      Weight Averaging                  & No                                         & No                                         & No                            \\
      Fisher Merging                    & Yes (Estimate Fisher information matrix)   & No                                         & No                            \\
      RegMean                           & Yes (compute Gram Matrix)                  & No                                         & No                            \\
      Task Arithmetic                   & No                                         & Yes (select scaling factor)                & No                            \\
      Ties-Merging                      & No                                         & Yes (select scaling factor)                & No                            \\
      AdaMerging                        & No                                         & No                                         & Yes                           \\
      TSV-M                             & No                                         & Yes (select scaling factor)                & No                            \\ \midrule
      \textbf{Concrete + $\mathcal{A}$} & \textbf{No, unless $\mathcal{A}$ requires} & \textbf{No, unless $\mathcal{A}$ requires} & \textbf{Yes}                  \\ \bottomrule
    \end{tabular}
  }
\end{table}

\textbf{Baselines.}
We compare our approach with simple averaging~\cite{wortsmanModelSoupsAveraging2022}, Task Arithmetic~\cite{ilharcoEditingModelsTask2023:b}, Ties-Merging~\cite{yadavResolvingInterferenceWhen2023:b}, AdaMerging~\cite{yangAdaMergingAdaptiveModel2024} and TSV-M~\cite{gargiuloTaskSingularVectors2025}.
The requirements of these methods are summarized in Table~\ref{table:model_merging_methods}.
A more detailed description is provided in Appendix~\ref{appendix:model_fusion_details}.

\subsection{Open-Vocabulary Classification Tasks}

We perform open-vocabulary classification by matching images and text in a shared multi-modal embedding space.
Specifically, we use the pre-trained CLIP-ViT-B/32 and -L/14 models, which map images and text into a common embedding space where semantically related pairs are close to each other.
The text encoders are kept fixed, while the image encoders are fine-tuned on downstream tasks.
We evaluate this approach on eight diverse datasets: SUN397, Stanford Cars, RESISC45, EuroSAT, SVHN, GTSRB, MNIST, and DTD. These datasets span a wide range of domains, including scene recognition, object classification, remote sensing, digit recognition, traffic sign recognition, and texture classification. Performance is measured using Top-1 accuracy, which reports the percentage of correctly classified samples.


\begin{table}[htb]
  \caption{Generalization results on two unseen tasks when merging CLIP-ViT-B/32 models on six tasks.}
  \label{table:generalization_results_clip-vit-b-32}
  \resizebox{\linewidth}{!}{%
    \centering
    \begin{tabular}{l|ccccccc|ccc}
      \toprule
      \multirow{2}{*}{\textbf{Method}} & \multicolumn{7}{c|}{\textbf{Seen Tasks}} & \multicolumn{3}{c}{\textbf{Unseen Tasks}}                                                                                                                                 \\
                                       & SUN397                                   & Cars                                      & RESISC45      & DTD           & SVHN          & GTSRB         & \textbf{Avg.} & MNIST         & EuroSAT       & \textbf{Avg.} \\
      \midrule
      Task Arithmetic                  & 63.4                                     & 62.3                                      & 75.3          & 57.8          & 84.7          & 80.4          & 70.7          & 77.3          & 45.6          & 61.4          \\
      Ties-Merging                     & \textbf{67.8}                            & 66.2                                      & 77.0          & 56.2          & 77.2          & 71.0          & 69.2          & 75.9          & 43.1          & 59.5          \\
      \textbf{Concrete TA}             & 66.2                                     & \textbf{66.4}                             & \textbf{82.0} & \textbf{58.3} & \textbf{91.4} & \textbf{92.7} & \textbf{72.9} & \textbf{80.7} & \textbf{52.9} & \textbf{66.8} \\
      \midrule
      AdaMerging                       & 65.2                                     & 65.9                                      & 88.5          & 61.1          & 92.2          & 91.5          & 77.4          & \textbf{84.0} & \textbf{56.1} & \textbf{70.0} \\
      AdaMerging++                     & 68.2                                     & 67.6                                      & 86.3          & 63.6          & 92.6          & 89.8          & 78.0          & 83.9          & 53.5          & 68.7          \\
      \textbf{Concrete AM}             & \textbf{68.9}                            & \textbf{71.7}                             & \textbf{91.2} & \textbf{66.9} & \textbf{94.1} & \textbf{97.5} & \textbf{81.7} & 83.6          & 53.9          & 69.7          \\
      \bottomrule
    \end{tabular}
  }
\end{table}

\begin{table}[htb]
  \caption{Ablation study on the rescaling operation.}
  \label{table:ablation_rescaling}
  \resizebox{\linewidth}{!}{%
    \centering
    \begin{tabular}{lccccccccc}
      \toprule
      \textbf{Method}              & \textbf{SUN397} & \textbf{Cars} & \textbf{RESISC45} & \textbf{EuroSAT} & \textbf{SVHN} & \textbf{GTSRB} & \textbf{MNIST} & \textbf{DTD}  & \textbf{Avg.} \\
      \midrule
      Concrete TA (w/o rescaling)  & \textbf{65.0}   & \textbf{64.0} & 74.0              & 85.0             & 66.1          & 53.8           & 94.6           & 51.0          & 69.2          \\
      Concrete TA (with rescaling) & 62.5            & 61.1          & \textbf{76.0}     & \textbf{95.7}    & \textbf{91.0} & \textbf{81.9}  & \textbf{98.5}  & \textbf{51.9} & \textbf{77.3} \\
      \bottomrule
    \end{tabular}
  }
\end{table}
\begin{table}[htb]
  \caption{Multi-task performance when merging Flan-T5-base (LoRA fine-tuned) models on all eight tasks.}
  \label{table:flan-t5-base_lora}
  \resizebox{\linewidth}{!}{%
    \centering
    \begin{tabular}{lccccccccc}
      \toprule
      \textbf{Method}         & \textbf{CoLA} & \textbf{MNLI} & \textbf{MRPC} & \textbf{QNLI} & \textbf{QQP}  & \textbf{RTE} & \textbf{SST2} & \textbf{STSB} & \textbf{Avg.} \\
      \midrule
      Pre-trained             & 69.1          & 56.5          & 76.2          & 88.4          & 82.1          & 80.1         & 91.2          & 62.2          & 75.7          \\
      Individual              & 69.1          & 82.7          & 85.5          & 90.9          & 84.0          & 84.4         & 92.9          & 87.4          & 84.6          \\
      Weight Averaging        & 69.7          & 59.7          & 78.9          & 90.1          & 83.8          & 80.5         & 91.2          & 72.0          & 78.2          \\
      \midrule
      \multicolumn{10}{c}{\textit{Task Arithmetic (TA)-Based}}                                                                                                               \\
      Task Arithmetic         & 68.8          & 55.2          & 78.7          & 89.8          & 83.7          & 79.1         & 91.5          & 72.4          & 77.4          \\
      Ties-Merging            & 68.3          & 56.3          & \textbf{79.4} & 89.8          & 83.7          & 79.4         & 91.6          & 71.2          & 77.5          \\
      \textbf{Concrete TA}    & \textbf{69.1} & \textbf{58.1} & 78.4          & \textbf{89.9} & 83.5          & 79.4         & 91.6          & \textbf{73.4} & \textbf{78.0} \\
      \midrule
      \multicolumn{10}{c}{\textit{Layer-wise AdaMerging (LW AM)-Based}}                                                                                                      \\
      LW AM                   & \textbf{69.1} & \textbf{60.3} & 78.4          & \textbf{90.0} & \textbf{83.6} & 79.1         & 91.6          & 74.1          & 78.3          \\
      \textbf{LW Concrete AM} & 69.0          & 59.4          & \textbf{80.1} & 89.9          & 82.9          & 79.1         & \textbf{91.7} & \textbf{75.4} & \textbf{78.5} \\
      \bottomrule
    \end{tabular}
  }
\end{table}

\textbf{Multi-task model fusion experiments.}
Table~\ref{table:multi-task_performance_clip-vit-b-32} shows that our Concrete methods consistently outperform baselines, achieving the highest average performance in their respective groups.
Layer-wise TSV-M surpasses all other fusion methods, trailing only traditional multi-task learning.
Moreover, among all methods, the Concrete variants of existing model fusion techniques demonstrate superior performance in the majority of tasks and achieve the highest average accuracy.
These findings underscore the robust effectiveness of our proposed Concrete subspace learning approach.

\begin{figure}[tb]
  \captionsetup[subfigure]{font=footnotesize}
  \begin{center}
    \hfill
    \begin{subfigure}[b]{0.365\textwidth}
      \centering
      \includegraphics[height=3.5cm]{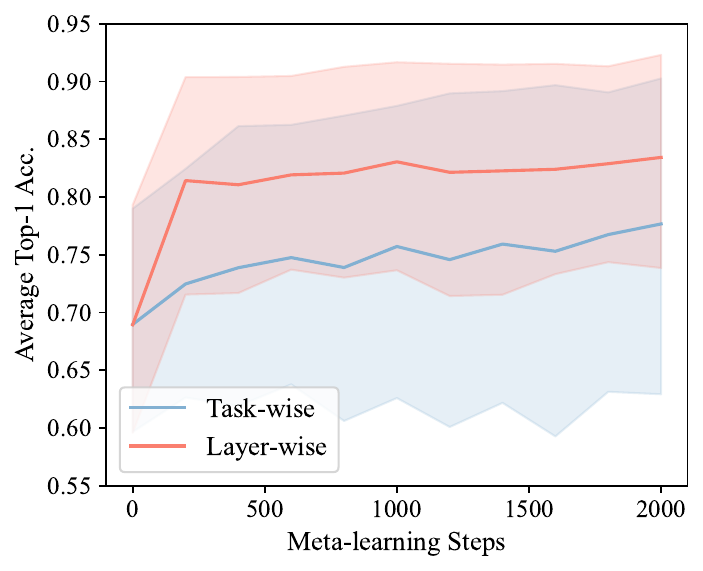}
      \caption{Meta-learn the Concrete mask}
      \label{fig:clip-vit-b-32_concrete_adamerging_meta}
    \end{subfigure}
    \hspace{0cm}
    \begin{subfigure}[b]{0.57\textwidth}
      \centering
      \includegraphics[height=3.5cm]{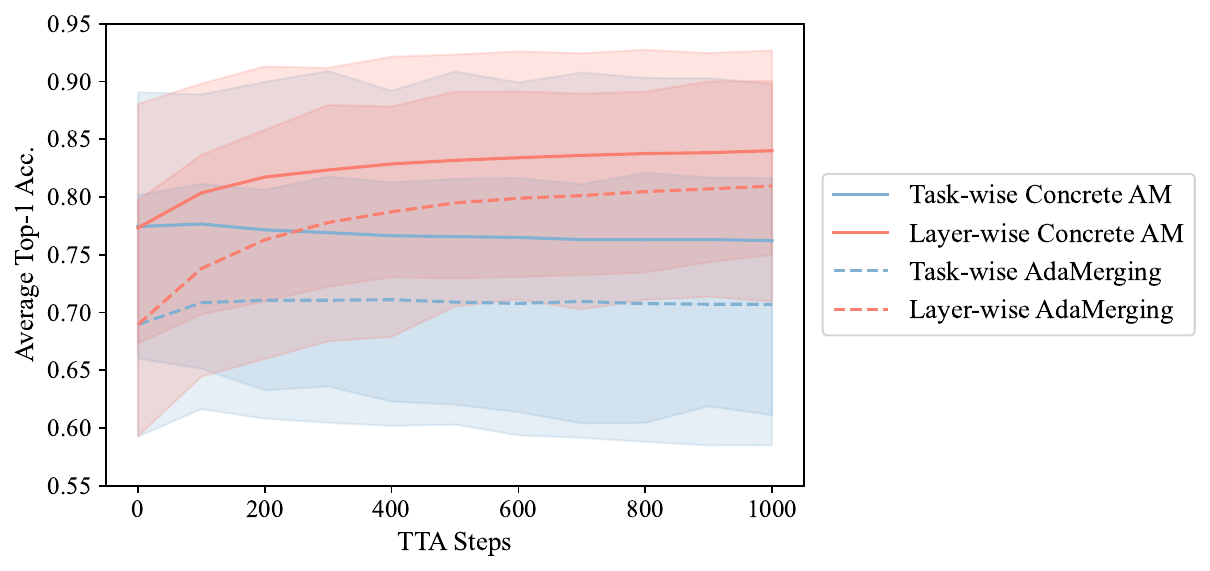}
      \caption{AdaMerging with/without Concrete mask}
      \label{fig:clip-vit-b-32_concrete_admerging_vs_adamerging_tta}
    \end{subfigure}
    \hfill
    \caption{
      Comparison of AdaMerging and Concrete AdaMerging on CLIP-ViT-B/32 models.
      (a) Performance evolution during the meta-learning phase (Algorithm~\ref{alg:meta-learn_mask}).
      (b) Comparison of AdaMerging with and without the Concrete mask.
    }
    \label{fig:clip-vit-b-32_concrete_adamerging_vs_adamerging}
  \end{center}
\end{figure}

Figure~\ref{fig:clip-vit-b-32_concrete_adamerging_vs_adamerging} compares AdaMerging with its Concrete counterpart in terms of multi-task performance during meta-learning and test-time adaptation.
Specifically, Figure~\ref{fig:clip-vit-b-32_concrete_adamerging_meta} shows steady performance gains during meta-learning, suggesting the mask captures shared task information.
Figure~\ref{fig:clip-vit-b-32_concrete_admerging_vs_adamerging_tta} confirms that the Concrete mask consistently enhances performance by mitigating task interference, outperforming the baseline without the mask.

\begin{table}[htb]
  \caption{
    Wall-clock cost of Concrete AdaMerging compared with AdaMerging TTA and multi-task learning.
  }
  \label{table:wall_time_cost}
  \centering
  \small
  \begin{tabular}{p{0.42\linewidth}p{0.46\linewidth}}
    \toprule
    \textbf{Method}                      & \textbf{Wall Time}                    \\
    \midrule
    AdaMerging TTA                       & 1{,}000 steps, $\sim$5 min            \\
    \hline
    \multirow{2}{*}{Concrete AdaMerging} & 400 steps meta-learning $+$           \\
                                         & 1{,}000 steps TTA, $\sim$5+5 = 10 min \\
    \hline
    Multi-task Learning                  & 8{,}000 steps, $\sim$1 hour           \\
    \bottomrule
  \end{tabular}
\end{table}

Table~\ref{table:wall_time_cost} reports the wall-clock cost under the same
single-GPU setting used in our rebuttal analysis.
Concrete AdaMerging adds about five minutes of mask meta-learning over
AdaMerging TTA, while remaining substantially cheaper than training a
multi-task model from scratch.
The learned mask is a one-time cost and adds only an element-wise multiplication
when applied during model fusion.


\textbf{Generalization experiments.}
Generalization experiments on CLIP-ViT-B/32 are conducted to evaluate the performance of the merged model on unseen tasks.
Here we set two tasks as unseen, while the remaining six tasks are considered seen.
The experiments involve merging the models fine-tuned on the six seen tasks, and evaluating the performance of the merged model on both the seen and unseen tasks.
The outcomes from two independent runs of the generalization experiments are detailed in Tables~\ref{table:generalization_results_clip-vit-b-32} and \ref{table:generalization_results_clip-vit-b-32_2}.
In Appendix~\ref{appendix:generalization_experiment_details} of the supplementary material, we provide additional details.

As shown in Tables~\ref{table:generalization_results_clip-vit-b-32} and \ref{table:generalization_results_clip-vit-b-32_2}, all Concrete methods demonstrate superior performance over their respective baselines.
In the first set of experiments, Concrete Task Arithmetic outperformed other Task Arithmetic-based methods in seven out of eight tasks, while in both sets, Concrete AdaMerging surpassed other methods in six out of eight tasks.
On unseen tasks, Concrete Task Arithmetic significantly outperforms other Task Arithmetic-based methods. This indicates that the learned Concrete mask can effectively capture the shared information between different tasks.
This is helpful for transferring knowledge from seen tasks to unseen tasks.

\begin{table}[htb]
  \caption{
    Ablations of the test data distribution on CLIP-ViT-B/32.
    We carry out comparison between $\mathcal{A}$ and its Concrete counterpart by bolding the better performance.
  }
  \label{table:ablation_data_distribution_vit_b_32}
  \resizebox{\linewidth}{!}{%
    \centering
    \begin{tabular}{l|cccc|cccc}
      \toprule
      \textbf{Method}              & \textbf{Cars}                                           & \textbf{EuroSAT}                                          & \textbf{RESISC45} & \textbf{GTSRB} & \textbf{Cars} & \textbf{EuroSAT} & \textbf{RESISC45} & \textbf{GTSRB} \\
      \midrule
                                   & \multicolumn{4}{c|}{{Clean Test Set}}                   & \multicolumn{4}{c}{{Corrupted Test Set (Motion Blur)}}                                                                                                                 \\
      Task Arithmetic              & 66.9                                                    & 94.7                                                      & 82.6              & 75.1           & 65.3          & 68.1             & 80.0              & 64.2           \\
      \textbf{Concrete TA}         & \textbf{68.9}                                           & \textbf{96.8}                                             & \textbf{83.4}     & \textbf{81.5}  & \textbf{66.9} & \textbf{73.9}    & \textbf{82.2}     & \textbf{68.1}  \\
      AdaMerging                   & 73.7                                                    & 96.1                                                      & 85.8              & 96.3           & 71.2          & 74.6             & 82.7              & 94.1           \\
      \textbf{Concrete AdaMerging} & \textbf{78.2}                                           & \textbf{97.3}                                             & \textbf{91.4}     & \textbf{98.0}  & \textbf{76.1} & \textbf{76.5}    & \textbf{89.6}     & \textbf{96.7}  \\
      \midrule
                                   & \multicolumn{4}{c|}{Corrupted Test Set (Impulse Noise)} & \multicolumn{4}{c}{Corrupted Test Set (Gaussian Noise)}                                                                                                                \\
      Task Arithmetic              & 62.1                                                    & \textbf{49.1}                                             & 72.7              & \textbf{40.4}  & 63.6          & \textbf{55.4}    & 75.9              & \textbf{49.4}  \\
      \textbf{Concrete TA}         & \textbf{65.2}                                           & 34.9                                                      & \textbf{74.8}     & 39.5           & \textbf{66.5} & 48.5             & \textbf{77.2}     & 47.2           \\
      AdaMerging                   & 67.2                                                    & \textbf{30.8}                                             & 75.9              & \textbf{77.5}  & 69.6          & \textbf{41.2}    & \textbf{90.9}     & 76.0           \\
      \textbf{Concrete AdaMerging} & \textbf{71.8}                                           & 21.1                                                      & \textbf{83.9}     & 70.1           & \textbf{72.6} & 20.8             & 86.6              & \textbf{76.7}  \\
      \midrule
                                   & \multicolumn{4}{c|}{Corrupted Test Set (Pixelate)}      & \multicolumn{4}{c}{Corrupted Test Set (Spatter)}                                                                                                                       \\
      Task Arithmetic              & \textbf{2.78}                                           & 41.5                                                      & \textbf{22.8}     & 66.6           & 63.3          & \textbf{60.1}    & 73.9              & \textbf{54.3}  \\
      \textbf{Concrete TA}         & 0.71                                                    & \textbf{42.0}                                             & 15.9              & \textbf{68.4}  & \textbf{65.7} & 59.0             & \textbf{75.5}     & 49.3           \\
      AdaMerging                   & \textbf{2.49}                                           & \textbf{53.8}                                             & \textbf{22.4}     & 90.6           & 69.9          & \textbf{43.6}    & 75.4              & 89.4           \\
      \textbf{Concrete AdaMerging} & 0.51                                                    & 37.8                                                      & 7.5               & \textbf{95.1}  & \textbf{70.9} & 22.0             & \textbf{82.7}     & \textbf{91.3}  \\
      \midrule
                                   & \multicolumn{4}{c|}{Corrupted Test Set (Contrast)}      & \multicolumn{4}{c}{Corrupted Test Set (JPEG Compression)}                                                                                                              \\
      Task Arithmetic              & 66.0                                                    & 62.9                                                      & 75.9              & 70.6           & 66.5          & 72.3             & 82.2              & 60.0           \\
      \textbf{Concrete TA}         & \textbf{67.9}                                           & \textbf{70.0}                                             & \textbf{78.4}     & \textbf{76.0}  & \textbf{68.7} & \textbf{74.8}    & \textbf{83.7}     & \textbf{62.1}  \\
      AdaMerging                   & 71.7                                                    & \textbf{69.8}                                             & 79.3              & 95.1           & 70.9          & \textbf{75.8}    & 83.6              & 90.1           \\
      \textbf{Concrete AdaMerging} & \textbf{75.4}                                           & 66.2                                                      & \textbf{87.2}     & \textbf{97.6}  & \textbf{76.8} & 54.4             & \textbf{90.2}     & \textbf{98.5}  \\
      \bottomrule
    \end{tabular}
  }
\end{table}

\textbf{Ablation study on the test data distribution.}
We conduct a series of ablation studies on the data distribution to assess the robustness of our methodology when faced with out-of-distribution (OOD) test data. Here, the unlabeled test data to which we aim to generalize may not originate from the same distribution as the training data. This discrepancy can pose significant challenges to performance and reliability.
Tables~\ref{table:ablation_data_distribution_vit_b_32} and \ref{table:ablation_data_distribution_vit_b_16} show the results of our experiments.
From the results, it is evident that our approach outperforms Task Arithmetic in half of the corruption types tested. For the other half, our method demonstrates comparable performance.
Please refer to Appendix~\ref{appendix:test_data_ablations} of the supplementary material for more details.

\textbf{Ablation study on the rescaling operation.}
When applying the masking process to task vectors, we rescale the masked task vectors. Here, we conduct a comparison between our method that involves rescaling and one without rescaling.
\begin{gather}
  \mathcal{M}^{w/}(\tau, m) = {\tau \circ m}/{\operatorname{Avg}(m)} \\
  \mathcal{M}^{w/o}(\tau, m) = \tau \circ m
\end{gather}
The results in Table~\ref{table:ablation_rescaling} suggest that rescaling is an essential component of our masking strategy, as it helps to amplify the effect of the masked task vectors and thereby improves the overall performance of the merged model.

\subsection{Text Generation Tasks}

For text generation tasks, we utilize the encoder-decoder Transformer structure Flan-T5-base and Flan-T5-large as our pre-trained models and conduct our experiments on eight NLP tasks from the GLUE benchmark~\cite{wangGLUEMultiTaskBenchmark2018}.
We report Spearman's $\rho$ for STSB and exact match accuracy for others.
The hyperparameter settings and other details about model fine-tuning, model fusion, and the prompt templates for each task are provided in Appendices~\ref{appendix:model_fine_tuning_details},~\ref{appendix:model_fusion_details} and \ref{appendix:preprocessed_examples_of_glue_benchmark} of the supplementary material.

\textbf{Multi-task model fusion experiments.}
Similar to the experiments on CLIP models, we conduct multi-task model fusion experiments on LoRA fine-tuned Flan-T5-base and Flan-T5-large.
The results are detailed in Tables~\ref{table:flan-t5-base_lora} and~\ref{table:flan-t5-large_lora}, respectively.
As shown in the tables, our proposed methodologies, the ``Concrete'' extension of Task Arithmetic and layer-wise AdaMerging, appear to offer incremental advantages.
We also observed that the effectiveness of our Concrete methods on LoRA fine-tuned language models was not significantly more pronounced when compared to other methods. This can be attributed to two factors:

(1) Pre-trained large language models inherently possess a remarkable ability to excel across a wide range of tasks due to their advanced architecture and training on vast amounts of diverse data.
However, this robust performance may limit the potential for substantial improvement through specialized methods. Given the already impressive capabilities of these models, further enhancements may yield only marginal gains.
(2) Additionally, in comparison to full fine-tuning, LoRA fine-tuning itself is a form of subspace fine-tuning.
Consequently, the potential performance gain of seeking an even smaller subspace within the LoRA fine-tuning may be limited, as the model's performance is already being efficiently adapted through the LoRA fine-tuning process.




\section{Conclusion}
\label{section:conclusion}

In this work, we propose Concrete subspace learning to resolve task interference in multi-task model fusion by identifying a shared low-dimensional subspace.
We also present enhanced extensions of two techniques, Task Arithmetic and AdaMerging, with our Concrete mask, which we have termed Concrete Task Arithmetic and Concrete AdaMerging, respectively.
Compared to previous algorithms, our method offers the advantage of effectively resolving task interference by considering the collective impact of parameters from task-specific models, rather than evaluating individual attributes like the parameters' magnitude or sign.

Our experiments in both the vision and language domains demonstrate the effectiveness of our method across a wide range of tasks.
While our method effectively addresses task interference and maintains overall performance, it requires solving a bi-level optimization problem.
This may increase computational time.
Future work could further explore the use of zero-order optimization techniques to reduce the memory footprint of our algorithm, thereby enhancing its scalability to larger models.




\bibliographystyle{unsrtnat}
\bibliography{references}

@inproceedings{ainsworthGitReBasinMerging2023:b,
  author     = {Ainsworth, Samuel and Hayase, Jonathan and Srinivasa, Siddhartha},
  booktitle  = {The {{Eleventh International Conference}} on {{Learning Representations}}},
  langid     = {english},
  month      = {September},
  shorttitle = {Git {{Re-Basin}}},
  title      = {Git {{Re-Basin}}: {{Merging Models}} modulo {{Permutation Symmetries}}},
  url        = {https://openreview.net/forum?id=CQsmMYmlP5T},
  year       = {2023}
}

@inproceedings{bentonLossSurfaceSimplexes2021:b,
  author       = {Benton, Gregory and Maddox, Wesley and Lotfi, Sanae and Wilson, Andrew Gordon Gordon},
  booktitle    = {International Conference on Machine Learning},
  organization = {PMLR},
  pages        = {769--779},
  title        = {Loss surface simplexes for mode connecting volumes and fast ensembling},
  year         = {2021}
}

@article{chengRemoteSensingImage2017,
  author    = {Cheng, Gong and Han, Junwei and Lu, Xiaoqiang},
  journal   = {Proceedings of the IEEE},
  number    = {10},
  pages     = {1865--1883},
  publisher = {IEEE},
  title     = {Remote sensing image scene classification: Benchmark and state of the art},
  volume    = {105},
  year      = {2017}
}

@inproceedings{chronopoulouAdapterSoupWeightAveraging2023,
  author    = {Alexandra Chronopoulou and
               Matthew E. Peters and
               Alexander Fraser and
               Jesse Dodge},
  editor    = {Andreas Vlachos and
               Isabelle Augenstein},
  title     = {AdapterSoup: Weight Averaging to Improve Generalization of Pretrained
               Language Models},
  booktitle = {Findings of the Association for Computational Linguistics: {EACL}
               2023, Dubrovnik, Croatia, May 2-6, 2023},
  pages     = {2009--2018},
  publisher = {Association for Computational Linguistics},
  year      = {2023}
}

@article{chungScalingInstructionFinetunedLanguage2024,
  author  = {Hyung Won Chung and
             Le Hou and
             Shayne Longpre and
             Barret Zoph and
             Yi Tay and
             William Fedus and
             Yunxuan Li and
             Xuezhi Wang and
             Mostafa Dehghani and
             Siddhartha Brahma and
             Albert Webson and
             Shixiang Shane Gu and
             Zhuyun Dai and
             Mirac Suzgun and
             Xinyun Chen and
             Aakanksha Chowdhery and
             Alex Castro{-}Ros and
             Marie Pellat and
             Kevin Robinson and
             Dasha Valter and
             Sharan Narang and
             Gaurav Mishra and
             Adams Yu and
             Vincent Y. Zhao and
             Yanping Huang and
             Andrew M. Dai and
             Hongkun Yu and
             Slav Petrov and
             Ed H. Chi and
             Jeff Dean and
             Jacob Devlin and
             Adam Roberts and
             Denny Zhou and
             Quoc V. Le and
             Jason Wei},
  title   = {Scaling Instruction-Finetuned Language Models},
  journal = {J. Mach. Learn. Res.},
  volume  = {25},
  pages   = {70:1--70:53},
  year    = {2024}
}

@inproceedings{cimpoiDescribingTexturesWild2014,
  author    = {Cimpoi, Mircea and Maji, Subhransu and Kokkinos, Iasonas and Mohamed, Sammy and Vedaldi, Andrea},
  booktitle = {Proceedings of the IEEE conference on computer vision and pattern recognition},
  pages     = {3606--3613},
  title     = {Describing textures in the wild},
  year      = {2014}
}

@inproceedings{danielfreemanTopologyGeometryHalfrectified2017,
  author     = {Daniel Freeman, C. and Bruna, Joan},
  shorttitle = {Topology and Geometry of Half-Rectified Network Optimization},
  title      = {Topology and Geometry of Half-Rectified Network Optimization: 5th {{International Conference}} on {{Learning Representations}}, {{ICLR}} 2017},
  year       = {2017}
}

@inproceedings{draxlerEssentiallyNoBarriers2019:b,
  author       = {Draxler, Felix and Veschgini, Kambis and Salmhofer, Manfred and Hamprecht, Fred},
  booktitle    = {International conference on machine learning},
  organization = {PMLR},
  pages        = {1309--1318},
  title        = {Essentially no barriers in neural network energy landscape},
  year         = {2018}
}

@inproceedings{entezariRolePermutationInvariance2022,
  author    = {Rahim Entezari and
               Hanie Sedghi and
               Olga Saukh and
               Behnam Neyshabur},
  title     = {The Role of Permutation Invariance in Linear Mode Connectivity of
               Neural Networks},
  booktitle = {The Tenth International Conference on Learning Representations, {ICLR}
               2022, Virtual Event, April 25-29, 2022},
  publisher = {OpenReview.net},
  year      = {2022}
}

@inproceedings{frankleLinearModeConnectivity2020:b,
  author       = {Frankle, Jonathan and Dziugaite, Gintare Karolina and Roy, Daniel and Carbin, Michael},
  booktitle    = {International Conference on Machine Learning},
  organization = {PMLR},
  pages        = {3259--3269},
  title        = {Linear mode connectivity and the lottery ticket hypothesis},
  year         = {2020}
}

@article{garipovLossSurfacesMode2018:b,
  author  = {Garipov, Timur and Izmailov, Pavel and Podoprikhin, Dmitrii and Vetrov, Dmitry P and Wilson, Andrew G},
  journal = {Advances in neural information processing systems},
  title   = {Loss surfaces, mode connectivity, and fast ensembling of dnns},
  volume  = {31},
  year    = {2018}
}

@inproceedings{georgestoicaZipItMergingModels2023:b,
  author    = {George Stoica and
               Daniel Bolya and
               Jakob Bjorner and
               Pratik Ramesh and
               Taylor Hearn and
               Judy Hoffman},
  title     = {ZipIt! Merging Models from Different Tasks without Training},
  booktitle = {The Twelfth International Conference on Learning Representations,
               {ICLR} 2024, Vienna, Austria, May 7-11, 2024},
  publisher = {OpenReview.net},
  year      = {2024}
}

@techreport{gumbelStatisticalTheoryExtreme1954,
  author      = {Gumbel, E. J.},
  institution = {{National Bureau of Standards, Washington, D. C. Applied Mathematics Div.}},
  number      = {PB175818},
  title       = {Statistical {{Theory}} of {{Extreme Values}} and {{Some Practical Applications}}. {{A Series}} of {{Lectures}}.},
  year        = {1954}
}

@inproceedings{helber2018introducing,
  author       = {Helber, Patrick and Bischke, Benjamin and Dengel, Andreas and Borth, Damian},
  booktitle    = {IGARSS 2018-2018 IEEE International Geoscience and Remote Sensing Symposium},
  organization = {IEEE},
  pages        = {204--207},
  title        = {Introducing EuroSAT: A Novel Dataset and Deep Learning Benchmark for Land Use and Land Cover Classification},
  year         = {2018}
}

@inproceedings{huangLoraHubEfficientCrossTask2024,
  author     = {Huang, Chengsong and Liu, Qian and Lin, Bill Yuchen and Pang, Tianyu and Du, Chao and Lin, Min},
  booktitle  = {First {{Conference}} on {{Language Modeling}}},
  langid     = {english},
  month      = {August},
  shorttitle = {{{LoraHub}}},
  title      = {{{LoraHub}}: {{Efficient Cross-Task Generalization}} via {{Dynamic LoRA Composition}}},
  url        = {https://openreview.net/forum?id=TrloAXEJ2B},
  year       = {2024}
}

@inproceedings{huLoRALowRankAdaptation2021a,
  author     = {Hu, Edward J. and Shen, Yelong and Wallis, Phillip and {Allen-Zhu}, Zeyuan and Li, Yuanzhi and Wang, Shean and Wang, Lu and Chen, Weizhu},
  booktitle  = {International {{Conference}} on {{Learning Representations}}},
  langid     = {english},
  month      = {October},
  shorttitle = {{{LoRA}}},
  title      = {{{LoRA}}: {{Low-Rank Adaptation}} of {{Large Language Models}}},
  url        = {https://openreview.net/forum?id=nZeVKeeFYf9},
  year       = {2021}
}

@inproceedings{ilharcoEditingModelsTask2023:b,
  author    = {Ilharco, Gabriel and Ribeiro, Marco Tulio and Wortsman, Mitchell and Schmidt, Ludwig and Hajishirzi, Hannaneh and Farhadi, Ali},
  booktitle = {The {{Eleventh International Conference}} on {{Learning Representations}}},
  langid    = {english},
  month     = {September},
  title     = {Editing Models with Task Arithmetic},
  url       = {https://openreview.net/forum?id=6t0Kwf8-jrj},
  year      = {2023}
}

@article{Survery_ModelMerging_2024,
  title   = {Model Merging in LLMs, MLLMs, and Beyond: Methods, Theories, Applications and Opportunities},
  author  = {Yang, Enneng and Shen, Li and Guo, Guibing and Wang, Xingwei and Cao, Xiaochun and Zhang, Jie and Tao, Dacheng},
  journal = {ACM Computing Surveys},
  year    = {2025}
}

@inproceedings{izmailovAveragingWeightsLeads2019:b,
  address   = {Monterey, California, USA},
  author    = {Izmailov, Pavel and Podoprikhin, Dmitrii and Garipov, Timur and Vetrov, Dmitry and Wilson, Andrew Gordon},
  booktitle = {Conference on {{Uncertainty}} in {{Artificial Intelligence}}},
  langid    = {english},
  title     = {Averaging {{Weights Leads}} to {{Wider Optima}} and {{Better Generalization}}},
  url       = {https://www.auai.org/uai2018/proceedings/papers/313.pdf},
  year      = {2018}
}

@inproceedings{jangCategoricalReparameterizationGumbelSoftmax2017,
  author    = {Eric Jang and
               Shixiang Gu and
               Ben Poole},
  title     = {Categorical Reparameterization with Gumbel-Softmax},
  booktitle = {{ICLR} (Poster)},
  publisher = {OpenReview.net},
  year      = {2017}
}

@inproceedings{jinDatalessKnowledgeFusion2023,
  author    = {Xisen Jin and
               Xiang Ren and
               Daniel Preotiuc{-}Pietro and
               Pengxiang Cheng},
  title     = {Dataless Knowledge Fusion by Merging Weights of Language Models},
  booktitle = {{ICLR}},
  publisher = {OpenReview.net},
  year      = {2023}
}

@inproceedings{kaddourStopWastingMy2022:b,
  author    = {Kaddour, Jean},
  booktitle = {Has It {{Trained Yet}}? {{NeurIPS}} 2022 {{Workshop}}},
  langid    = {english},
  month     = {October},
  title     = {Stop {{Wasting My Time}}! {{Saving Days}} of {{ImageNet}} and {{BERT Training}} with {{Latest Weight Averaging}}},
  url       = {https://openreview.net/forum?id=0OrABUHZuz},
  year      = {2022}
}

@inproceedings{krause3DObjectRepresentations2013,
  author    = {Krause, Jonathan and Stark, Michael and Deng, Jia and Fei-Fei, Li},
  booktitle = {Proceedings of the IEEE international conference on computer vision workshops},
  pages     = {554--561},
  title     = {3d object representations for fine-grained categorization},
  year      = {2013}
}

@article{lecunGradientbasedLearningApplied1998,
  author    = {LeCun, Yann and Bottou, L{\'e}on and Bengio, Yoshua and Haffner, Patrick},
  journal   = {Proceedings of the IEEE},
  number    = {11},
  pages     = {2278--2324},
  publisher = {Ieee},
  title     = {Gradient-based learning applied to document recognition},
  volume    = {86},
  year      = {1998}
}

@article{liangComprehensiveSurveyTestTime2023,
  author  = {Jian Liang and
             Ran He and
             Tieniu Tan},
  title   = {A Comprehensive Survey on Test-Time Adaptation Under Distribution
             Shifts},
  journal = {Int. J. Comput. Vis.},
  volume  = {133},
  number  = {1},
  pages   = {31--64},
  year    = {2025}
}

@inproceedings{liConvergentLearningDifferent2016:b,
  author    = {Yixuan Li and
               Jason Yosinski and
               Jeff Clune and
               Hod Lipson and
               John E. Hopcroft},
  title     = {Convergent Learning: Do different neural networks learn the same representations?},
  booktitle = {{ICLR}},
  year      = {2016}
}

@article{liDeepModelFusion2023:b,
  author  = {Weishi Li and
             Yong Peng and
             Miao Zhang and
             Liang Ding and
             Han Hu and
             Li Shen},
  title   = {Deep Model Fusion: {A} Survey},
  journal = {CoRR},
  volume  = {abs/2309.15698},
  year    = {2023}
}

@misc{dimitriadisParetoLowRankAdapters2024,
  title         = {Pareto {{Low-Rank Adapters}}: {{Efficient Multi-Task Learning}} with {{Preferences}}},
  shorttitle    = {Pareto {{Low-Rank Adapters}}},
  author        = {Dimitriadis, Nikolaos and Frossard, Pascal and Fleuret, Francois},
  year          = 2024,
  month         = jul,
  number        = {arXiv:2407.08056},
  eprint        = {2407.08056},
  primaryclass  = {cs},
  publisher     = {arXiv},
  doi           = {10.48550/arXiv.2407.08056},
  url           = {http://arxiv.org/abs/2407.08056},
  archiveprefix = {arXiv}
}

@inproceedings{gargiuloTaskSingularVectors2025,
  title      = {Task {{Singular Vectors}}: {{Reducing Task Interference}} in {{Model Merging}}},
  shorttitle = {Task {{Singular Vectors}}},
  booktitle  = {Proceedings of the {{IEEE}}/{{CVF Conference}} on {{Computer Vision}} and {{Pattern Recognition}}},
  author     = {Gargiulo, Antonio Andrea and Crisostomi, Donato and Bucarelli, Maria Sofia and Scardapane, Simone and Silvestri, Fabrizio and Rodol{\`a}, Emanuele},
  year       = 2025,
  pages      = {18695--18705},
  url        = {https://openaccess.thecvf.com/content/CVPR2025/html/Gargiulo_Task_Singular_Vectors_Reducing_Task_Interference_in_Model_Merging_CVPR_2025_paper.html},
  langid     = {english}
}

@book{luceIndividualChoiceBehavior1959,
  address   = {{Oxford, England}},
  author    = {Luce, R. Duncan},
  pages     = {xii, 153},
  publisher = {{John Wiley}},
  series    = {Individual Choice Behavior},
  title     = {Individual Choice Behavior},
  year      = {1959}
}

@article{maddison2014sampling,
  author  = {Maddison, Chris J and Tarlow, Daniel and Minka, Tom},
  journal = {Advances in neural information processing systems},
  title   = {A* sampling},
  volume  = {27},
  year    = {2014}
}

@inproceedings{maddisonConcreteDistributionContinuous2017,
  author    = {Chris J. Maddison and
               Andriy Mnih and
               Yee Whye Teh},
  title     = {The Concrete Distribution: {A} Continuous Relaxation of Discrete Random
               Variables},
  booktitle = {{ICLR} (Poster)},
  publisher = {OpenReview.net},
  year      = {2017}
}

@inproceedings{matenaMergingModelsFisherWeighted2022,
  author    = {Michael Matena and
               Colin Raffel},
  title     = {Merging Models with Fisher-Weighted Averaging},
  booktitle = {NeurIPS},
  year      = {2022}
}

@inproceedings{mounsavengBagTricksFully2024,
  author    = {Saypraseuth Mounsaveng and
               Florent Chiaroni and
               Malik Boudiaf and
               Marco Pedersoli and
               Ismail Ben Ayed},
  title     = {Bag of Tricks for Fully Test-Time Adaptation},
  booktitle = {{WACV}},
  pages     = {1925--1934},
  publisher = {{IEEE}},
  year      = {2024}
}

@inproceedings{nagarajanUniformConvergenceMay2019:a,
  author    = {Nagarajan, Vaishnavh and Kolter, J. Zico},
  booktitle = {Advances in {{Neural Information Processing Systems}}},
  publisher = {{Curran Associates, Inc.}},
  title     = {Uniform Convergence May Be Unable to Explain Generalization in Deep Learning},
  volume    = {32},
  year      = {2019}
}

@article{netzer_reading_2021,
  author   = {Netzer, Yuval and Wang, Tao and Coates, Adam and Bissacco, Alessandro and Wu, Bo and Ng, Andrew Y},
  language = {en},
  title    = {Reading {Digits} in {Natural} {Images} with {Unsupervised} {Feature} {Learning}},
  year     = {2021}
}

@inproceedings{prabhakarLoRASoupsMerging2025,
  address    = {Abu Dhabi, UAE},
  author     = {Prabhakar, Akshara and Li, Yuanzhi and Narasimhan, Karthik and Kakade, Sham and Malach, Eran and Jelassi, Samy},
  booktitle  = {Proceedings of the 31st {{International Conference}} on {{Computational Linguistics}}: {{Industry Track}}},
  editor     = {Rambow, Owen and Wanner, Leo and Apidianaki, Marianna and {Al-Khalifa}, Hend and Eugenio, Barbara Di and Schockaert, Steven and Darwish, Kareem and Agarwal, Apoorv},
  month      = {January},
  pages      = {644--655},
  publisher  = {Association for Computational Linguistics},
  shorttitle = {{{LoRA Soups}}},
  title      = {{{LoRA Soups}}: {{Merging LoRAs}} for {{Practical Skill Composition Tasks}}},
  url        = {https://aclanthology.org/2025.coling-industry.55/},
  year       = {2025}
}

@inproceedings{radford2021learning,
  author       = {Radford, Alec and Kim, Jong Wook and Hallacy, Chris and Ramesh, Aditya and Goh, Gabriel and Agarwal, Sandhini and Sastry, Girish and Askell, Amanda and Mishkin, Pamela and Clark, Jack and others},
  booktitle    = {International conference on machine learning},
  organization = {PMLR},
  pages        = {8748--8763},
  title        = {Learning transferable visual models from natural language supervision},
  year         = {2021}
}

@inproceedings{radfordLearningTransferableVisual2021,
  author    = {Alec Radford and
               Jong Wook Kim and
               Chris Hallacy and
               Aditya Ramesh and
               Gabriel Goh and
               Sandhini Agarwal and
               Girish Sastry and
               Amanda Askell and
               Pamela Mishkin and
               Jack Clark and
               Gretchen Krueger and
               Ilya Sutskever},
  title     = {Learning Transferable Visual Models From Natural Language Supervision},
  booktitle = {{ICML}},
  series    = {Proceedings of Machine Learning Research},
  volume    = {139},
  pages     = {8748--8763},
  publisher = {{PMLR}},
  year      = {2021}
}

@article{stallkamp_man_2012,
  author     = {Stallkamp, J. and Schlipsing, M. and Salmen, J. and Igel, C.},
  doi        = {10.1016/j.neunet.2012.02.016},
  issn       = {0893-6080},
  journal    = {Neural Networks},
  month      = {August},
  pages      = {323--332},
  series     = {Selected {Papers} from {IJCNN} 2011},
  shorttitle = {Man vs. computer},
  title      = {Man vs. computer: {Benchmarking} machine learning algorithms for traffic sign recognition},
  volume     = {32},
  year       = {2012}
}

@inproceedings{tangParameterEfficientMultitask2024,
  author    = {Tang, Anke and Shen, Li and Luo, Yong and Zhan, Yibing and Hu, Han and Du, Bo and Chen, Yixin and Tao, Dacheng},
  booktitle = {The {{Twelfth International Conference}} on {{Learning Representations}}},
  langid    = {english},
  month     = {October},
  title     = {Parameter-{{Efficient Multi-Task Model Fusion}} with {{Partial Linearization}}},
  url       = {https://openreview.net/forum?id=iynRvVVAmH},
  year      = {2024}
}

@inproceedings{tatroOptimizingModeConnectivity2020:a,
  author    = {Tatro, Norman and Chen, Pin-Yu and Das, Payel and Melnyk, Igor and Sattigeri, Prasanna and Lai, Rongjie},
  booktitle = {Advances in {{Neural Information Processing Systems}}},
  pages     = {15300--15311},
  publisher = {{Curran Associates, Inc.}},
  title     = {Optimizing {{Mode Connectivity}} via {{Neuron Alignment}}},
  volume    = {33},
  year      = {2020}
}

@inproceedings{wangGLUEMultiTaskBenchmark2018,
  address    = {{Brussels, Belgium}},
  author     = {Wang, Alex and Singh, Amanpreet and Michael, Julian and Hill, Felix and Levy, Omer and Bowman, Samuel},
  booktitle  = {Proceedings of the 2018 {{EMNLP Workshop BlackboxNLP}}: {{Analyzing}} and {{Interpreting Neural Networks}} for {{NLP}}},
  doi        = {10.18653/v1/W18-5446},
  langid     = {english},
  pages      = {353--355},
  publisher  = {{Association for Computational Linguistics}},
  shorttitle = {{{GLUE}}},
  title      = {{{GLUE}}: {{A Multi-Task Benchmark}} and {{Analysis Platform}} for {{Natural Language Understanding}}},
  url        = {http://aclweb.org/anthology/W18-5446},
  year       = {2018}
}

@inproceedings{wortsmanModelSoupsAveraging2022,
  author    = {Mitchell Wortsman and
               Gabriel Ilharco and
               Samir Yitzhak Gadre and
               Rebecca Roelofs and
               Raphael Gontijo Lopes and
               Ari S. Morcos and
               Hongseok Namkoong and
               Ali Farhadi and
               Yair Carmon and
               Simon Kornblith and
               Ludwig Schmidt},
  title     = {Model soups: averaging weights of multiple fine-tuned models improves
               accuracy without increasing inference time},
  booktitle = {{ICML}},
  series    = {Proceedings of Machine Learning Research},
  volume    = {162},
  pages     = {23965--23998},
  publisher = {{PMLR}},
  year      = {2022}
}

@article{wuHeterogeneousModelReuse2019,
  author     = {Wu, Xi Zhu and Liu, Song and Zhou, Zhi Hua},
  isbn       = {9781510886988},
  journal    = {36th International Conference on Machine Learning, ICML 2019},
  pages      = {11862--11871},
  shorttitle = {{{HMR}}},
  title      = {Heterogeneous Model Reuse via Optimizing Multiparty Multiclass Margin},
  volume     = {2019-June},
  year       = {2019}
}

@inproceedings{wuPiTuningTransferring2023:b,
  author       = {Wu, Chengyue and Wang, Teng and Ge, Yixiao and Lu, Zeyu and Zhou, Ruisong and Shan, Ying and Luo, Ping},
  booktitle    = {International Conference on Machine Learning},
  organization = {PMLR},
  pages        = {37713--37727},
  title        = {$\pi$-Tuning: Transferring Multimodal Foundation Models with Optimal Multi-task Interpolation},
  year         = {2023}
}

@inproceedings{xiao_sun_2010,
  address    = {San Francisco, CA, USA},
  author     = {Xiao, Jianxiong and Hays, James and Ehinger, Krista A. and Oliva, Aude and Torralba, Antonio},
  booktitle  = {2010 {IEEE} {Computer} {Society} {Conference} on {Computer} {Vision} and {Pattern} {Recognition}},
  doi        = {10.1109/CVPR.2010.5539970},
  isbn       = {978-1-4244-6984-0},
  language   = {en},
  month      = {June},
  pages      = {3485--3492},
  publisher  = {IEEE},
  shorttitle = {{SUN} database},
  title      = {{SUN} database: {Large}-scale scene recognition from abbey to zoo},
  url        = {http://ieeexplore.ieee.org/document/5539970/},
  year       = {2010}
}

@article{yadavResolvingInterferenceWhen2023:b,
  author     = {Yadav, Prateek and Tam, Derek and Choshen, Leshem and Raffel, Colin A. and Bansal, Mohit},
  journal    = {Advances in Neural Information Processing Systems},
  langid     = {english},
  month      = {December},
  pages      = {7093--7115},
  shorttitle = {{{TIES-Merging}}},
  title      = {{{TIES-Merging}}: {{Resolving Interference When Merging Models}}},
  url        = {https://proceedings.neurips.cc/paper_files/paper/2023/hash/1644c9af28ab7916874f6fd6228a9bcf-Abstract-Conference.html},
  volume     = {36},
  year       = {2023}
}

@article{yangAdaMergingAdaptiveModel2024,
  author  = {Yang, Enneng and Wang, Zhenyi and Shen, Li and Liu, Shiwei and Guo, Guibing and Wang, Xingwei and Tao, Dacheng},
  journal = {The Twelfth International Conference on Learning Representations},
  title   = {AdaMerging: Adaptive Model Merging for Multi-Task Learning},
  year    = {2024}
}

@inproceedings{yuLanguageModelsAre2024,
  author     = {Yu, Le and Yu, Bowen and Yu, Haiyang and Huang, Fei and Li, Yongbin},
  booktitle  = {Proceedings of the 41st {{International Conference}} on {{Machine Learning}}},
  issn       = {2640-3498},
  langid     = {english},
  month      = {July},
  pages      = {57755--57775},
  publisher  = {PMLR},
  shorttitle = {Language {{Models}} Are {{Super Mario}}},
  title      = {Language {{Models}} Are {{Super Mario}}: {{Absorbing Abilities}} from {{Homologous Models}} as a {{Free Lunch}}},
  url        = {https://proceedings.mlr.press/v235/yu24p.html},
  year       = {2024}
}

@inproceedings{yunisConvexityLinearMode2022:b,
  author    = {Yunis, David and Patel, Kumar Kshitij and Savarese, Pedro Henrique Pamplona and Vardi, Gal and Frankle, Jonathan and Walter, Matthew and Livescu, Karen and Maire, Michael},
  booktitle = {OPT 2022: Optimization for Machine Learning (NeurIPS 2022 Workshop)},
  title     = {On convexity and linear mode connectivity in neural networks},
  year      = {2022}
}

@article{zhangComposingParameterEfficientModules2023,
  author  = {Zhang, Jinghan and Chen, Shiqi and Liu, Junteng and He, Junxian},
  journal = {Advances in Neural Information Processing Systems},
  langid  = {english},
  month   = {December},
  pages   = {12589--12610},
  title   = {Composing {{Parameter-Efficient Modules}} with {{Arithmetic Operation}}},
  url     = {https://proceedings.neurips.cc/paper_files/paper/2023/hash/299a08ee712d4752c890938da99a77c6-Abstract-Conference.html},
  volume  = {36},
  year    = {2023}
}

@article{zhengLearningModelsFinetuning2025,
  author  = {Zheng, Hongling and Shen, Li and Tang, Anke and Luo, Yong and Hu, Han and Du, Bo and Wen, Yonggang and Tao, Dacheng},
  doi     = {10.1038/s42256-024-00961-0},
  issn    = {2522-5839},
  journal = {Nature Machine Intelligence},
  langid  = {english},
  month   = {January},
  number  = {1},
  pages   = {6--17},
  title   = {Learning from Models beyond Fine-Tuning},
  url     = {https://www.nature.com/articles/s42256-024-00961-0},
  volume  = {7},
  year    = {2025}
}

\appendix
\newpage
\startcontents[sections]
\printcontents[sections]{}{1}{\setcounter{tocdepth}{2}}
\vskip 0.2in \hrule
\section{Gumbel-Max Trick and Concrete Categorical Distribution}
\label{appendix:concrete_masking}

\subsection{The Gumbel-Max Trick}

Consider a discrete categorical distribution parameterized by logits $\psi \in \mathbb{R}^{n}$, where $\psi_i$ is the logit of the $i$-th category.
The Gumbel-Max trick \cite{gumbelStatisticalTheoryExtreme1954,luceIndividualChoiceBehavior1959,maddison2014sampling} provides a reparameterization trick to sample from the categorical distribution. This is achieved by sampling from the standard Gumbel distribution $g\sim\text{Gumbel}(\mu=0,\beta=1)$ and taking the argmax of the sum of the Gumbel random variables $g$ and the logits $\psi$.

This trick proceeds as follows:
sample $n$ Gumbel random variables $g_1, \dots, g_n$ independently from the standard Gumbel distribution $\text{Gumbel}(\mu=0,\beta=1)$~\footnote{We can draw a random sample $u$ from a unifrom distribution on the interval $(0,1)$ and then transform it into a Gumbel-distributed variable $g$ using the formula $g=-\log(-\log u)$.}, find the index $i$ of that maximizes $\psi_i + g_i$, then we have
\begin{equation}
  \label{eq:gumbel_max_trick}
  \argmax_{i\in[n]} (\psi_i + g_i) \sim \text{Categorical}(\text{softmax}(\psi)).
\end{equation}
If we represent the categorical distribution as a one-hot vector $\mathbf{y} = (y_1, \dots, y_n) \in \{0,1\}^n$, where $y_i=1$ indicates that the $i$-th category is sampled and for all $j\neq i$, $y_j=0$, then we have
\begin{align}
  \mathbb{P}(y_k=1) & = \mathbb{P}\left(\argmax_{i\in[n]} (\psi_i + g_i) = k\right)                          \\
                    & = \frac{\exp(\psi_k)}{\sum_{i=1}^n \exp(\psi_i)}.  \label{eq:gumbel_max_trick_one_hot}
\end{align}

\subsection{Continuous Relaxation of the discrete Categorical Distribution}

Since the derivative of the argmax function is not defined, we cannot backpropagate the gradients through it.
To address this issue, \cite{maddisonConcreteDistributionContinuous2017} proposed to use a CONtinuous relaxation of the disCRETE (Concrete) categorical distribution.
A Concrete random variable relax the condition that the one-hot vector $y$ must be located at the vertices of the $(n-1)$-dimensional simplex $\Delta^{n-1}$, and instead, it allows $y$ to be located anywhere inside the simplex $\Delta^{n-1}$, i.e.\ $\left\{ y\in \mathbb{R}^n \mid y_i \in [0,1], \sum_{i=1}^n y_i =1 \right\}$.

To sample a Concrete random variable $y$ from a distribution that is parameterized by a temperature hyperparameter $\lambda > 0$ and a real-value logit vector  $\psi \in \mathbb{R}^{n}$, we have
\begin{gather}
  \label{eq:concrete_sample}
  g\sim\text{Gumbel}(\mu=0,\beta=1),\\
  y = \text{softmax}\left(\frac{\psi + g}{\lambda}\right), \\
  y_i = \frac{\exp\left((\psi_i + g_i)/{\lambda}\right)}{\sum_{j=1}^n \exp\left(({\psi_j + g_j})/{\lambda}\right)} \quad \text{for} \,\, i\in[n].
\end{gather}

\subsection{The Concrete Mask}
\label{section:the_concrete_masking_process}

\textbf{The Concrete mask and the sampling process.}
A binary mask can be drawn from a Bernoulli distribution denoted by $\text{Bernoulli}(p=\sigma(\mathbf{\psi}))$, where $p$ is the probability (and $\psi$ denotes the logits) of each parameter being activated.
The left portion of Figure~\ref{fig:overview}(c) illustrates the process of generating discrete binary masks by sampling from a uniform distribution and comparing the samples with the parameterized probabilities.
Alternatively, the Gumbel-Max trick can be employed to treat the Bernoulli distribution as a 2-class categorical distribution for sampling.
However, the discrete Bernoulli distribution is not differentiable, preventing the backpropagation of gradients through it to optimize the parameters $\psi$.

\begin{definition}[Concrete mask]
  A Concrete mask, denoted as $\mathbf{m}$, is a real-valued random vector from sample space $[0,1]^d$.
  It is parameterized by logits $\psi \in \mathbb{R}^d$ or probabilities $p = \sigma(\psi)$, where $d$ is the number of parameters in a neural network.
  Each entry $m_i$ of the mask is a random variable sampled from a Concrete distribution as Eq.(\ref{eq:concrete_mask_sampling}).
\end{definition}

Inspired by \cite{maddisonConcreteDistributionContinuous2017,jangCategoricalReparameterizationGumbelSoftmax2017}, we use a CONtinuous relaxation of the disCRETE Bernoulli distribution to sample the shared mask $m$ for identifying the common subspace.
Figure~\ref{fig:overview}(c) illustrates the sampling procedure of the differentiable Concrete mask.
The Bernoulli random variables can be considered as a special 2-class case of the Gumbel-Softmax random variables, which are continuous relaxations of the discrete Categorical random variables.
In Appendix~\ref{appendix:concrete_masking} of the supplementary material, we provide a more detailed explanation of the Gumbel-Max trick, Concrete random variables, and Gumbel-Softmax distribution.

Consider a Bernoulli random variable, where $p_0$ and $p_1$ represent the unnormalized probabilities of the outcomes 0 and 1, respectively. Here, $\psi$ stands for the logits. In this context, the probability of the occurrence $m=1$ is
\begin{equation}
  \mathbb{P}^+ = \mathbb{P}(m=1) = \frac{p_1}{p_0 + p_1} = \sigma(\psi),
\end{equation}
where $\sigma$ denotes the sigmoid function.
In the context of the Gumbel-Max trick, the occurrence of the event $m=1$ is determined by the condition $g_1 + \log p_1 > g_0 + \log p_0$, where $g_0$ and $g_1$ are two independent standard Gumbel random variables.
Consequently, we can express this as:
\begin{align}
  \label{eq:P_m_1}
  \mathbb{P}^+ & = \mathbb{P}(g_1 + \log p_1 > g_0 + \log p_0) \nonumber          \\
               & = \mathbb{P}\left((g_1 - g_0) + (\log p_1 - \log p_0)> 0\right).
\end{align}
Because the difference of two standard Gumbel random variables is a Logistic random variable, we can replace $g_1 - g_0$ by $\log u - \log(1-u)$ where $u$ is a random variable sampled from a uniform distribution on the interval $(0,1)$.
Substituting this into Eq.(\ref{eq:P_m_1}) and expressing the probability in terms of the logits $\psi$ to simplify the expression, we have
\begin{flalign}
  \label{eq:bernoulli_hard}
  \mathbb{P}^+  = \mathbb{P}  \left(\log \frac{u}{1-u}  + \log \frac{\sigma(\psi)}{1-\sigma(\psi)} > 0\right), \nonumber \\
  \quad\quad\quad\quad\quad u \sim \text{Uniform}(0,1).
\end{flalign}

\begin{figure}
  \begin{center}
    \includegraphics[width=0.45\linewidth]{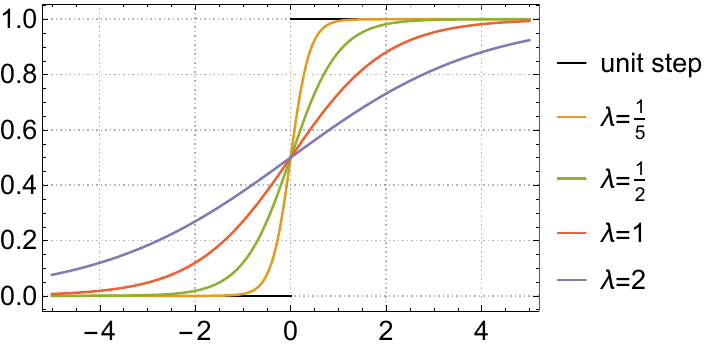}
    \caption{The sigmoid function with different temperatures $\lambda$.}
    \label{fig:sigmoid}
  \end{center}
  \vskip -0.15in
\end{figure}

The binary Concrete distribution offers a continuous relaxation of the discrete Bernoulli random variables, which is beneficial for gradient-based optimization as it allows for the backpropagation of gradients even through the sampling process.
Instead of making a hard decision as in Eq.(\ref{eq:bernoulli_hard}), we use a temperature parameter $\lambda$ to control the steepness of the sigmoid function, and hence control how close our `soft' decisions are to being `hard' decisions. The continuous version of the Bernoulli random variable is then given by
\begin{equation}
  \label{eq:concrete_mask_sampling}
  m = \sigma\left(\left(\log \frac{u}{1 - u} + \log \frac{\sigma(\psi)}{1 - \sigma(\psi)}\right) / \lambda\right).
\end{equation}

As the temperature $\lambda$ approaches zero, the sigmoid function becomes a step function, and the Concrete random variable $\hat{m}$ becomes a Bernoulli random variable, as shown in Figure~\ref{fig:sigmoid}. In the limit when $\lambda \to 0$, this results in sampling $m=1$ if $\log \frac{\sigma(\psi)}{1 - \sigma(\psi)} > -\log \frac{u}{1 - u}$, consistent with the original Gumbel-Max trick.
The binary Concrete distribution thus provides a differentiable approximation to Bernoulli random variables.
In addition to the Concrete mask sampling, we can further binarize the Concrete mask by setting the entries with values greater than 0.5 to 1 and the rest to 0.

\textbf{The masking process.}
For a neural network, we can use a mask $m$ to identify a subspace of the parameter space by setting the parameters that are not in the subspace to zero, i.e., $\mathbf{\theta} \circ m$, where $\circ$ denotes the element-wise product.
However, as we focus on the task-specific knowledge of fine-tuned models, we perform this operation on the task vector $\tau_i \circ m$. Following this operation, we rescale the remaining parameters of the task vector. Specifically, we divide them by the mean value of the mask entries to obtain the final task vector
\begin{equation}
  \mathcal{M}(\tau, m)= \frac{\tau_i \circ m}{Avg(m)}.
\end{equation}

This rescaling trick has also been employed in \cite{yuLanguageModelsAre2024}, which has been shown to be effective in maintaining the performance of the pruned models.
Without the rescaling, the pruned models would be significantly degraded, as the mask would be too sparse to retain the knowledge.

\section{Algorithms Details}
\label{appendix:algorithm_details}

In this section, we present a more detailed and complete pseudocode of the proposed algorithm. This pseudocode provides a step-by-step description of the algorithm's process, making it easier to understand how the algorithm works.

\begin{algorithm}[!tbp]
  \caption{Meta-Learning a shared Concrete mask across tasks}
  \label{alg:meta-learn_mask-long}
  \small
  \begin{algorithmic}
    \State {\bfseries Input:}
    \State \hspace{1.5em} a pre-trained model $f$ parameterized by $\theta_0$,
    \State \hspace{1.5em} a set of fine-tuned task vectors $\mathcal{T} = \{\tau_i\}_{i=1}^n$,
    \State \hspace{1.5em} a set of target tasks $\mathcal{S}^{\text{test}}$ (unlabeled data).
    \State {\bfseries Output:} a shared Concrete mask $\mathbf{m}$ parameterized by logits $\mathbf{x}$.
    \State Initialize the logits $\mathbf{x}$ to zeros
    \Repeat
    \State mask task vectors $\mathcal{T}$ with $\mathbf{m}$ to get $\mathcal{T}'$
    \State rescale the masked task vectors $\mathcal{T}'$ to get $\mathcal{T}''$
    \State initialize parameter $w$ associated with model fusion algorithm
    \State $\theta \leftarrow \text{MergeWeight}(\theta_0, \mathcal{T}''; w)$ \Comment{such as Task Arithmetic, AdaMerging or ModelSoups}
    \If{$w$ is optimizable}
    \For {each task $s_i \in \mathcal{S}^{\text{test}}$}
    \State sample a batch of unlabeled data $\mathcal{D}_i$ from $s_i$
    \State $l_i \leftarrow \mathcal{L}_i(f(\theta), \mathcal{D}_i)$ \Comment{unsupervised loss such as entropy loss for TTA}
    \EndFor
    \State $w' \leftarrow w - \alpha \nabla_w \left(\sum_{i=1}^n  l_i\right)$
    \State $\theta \leftarrow \text{MergeWeight}(\theta_0,\mathcal{T}''; w')$ \Comment{update merged weights with the updated $w'$}
    \EndIf
    \For {each task $s_i \in \mathcal{S}$}
    \State sample a batch of unlabeled data $\mathcal{D}_i$ from $s_i$
    \State $l_i \leftarrow \mathcal{L}_i(f(\theta), \mathcal{D}_i)$  \Comment{unsupervised loss such as entropy loss for TTA}
    \EndFor
    \State $\mathbf{x} \leftarrow \mathbf{x} - \beta \nabla_{\mathbf{x}}\left(\sum_{i=1}^n l_i\right)$
    \Until{convergence}
    \State {\bfseries Return:} the Concrete mask paramemetized by logits $\mathbf{x}$.
  \end{algorithmic}
\end{algorithm}

Algorithm~\ref{alg:meta-learn_mask-long} is the expand version of Algorithm~\ref{alg:meta-learn_mask} in the main paper.

\begin{algorithm}[!tbp]
  \caption{Concrete Task Arithmetic}
  \label{alg:concrete_task_arithmetic}
  \small
  \begin{algorithmic}
    \State {\bfseries Input:} pre-trained weights $\theta_0$,
    fine-tuned task vectors $\mathcal{T}=\{\tau_i\}_{i=1}^T$, a single scaling factor $w$.
    \\ \vskip -6pt \hrulefill
    \State $\mathbf{m} \gets \text{MetaLearnMask}(\theta_0, \mathcal{T})$ (Algorithm~\ref{alg:meta-learn_mask})
    \State Sample a mask $m$ from $\mathbf{m}$
    \State $\mathcal{T}' \leftarrow \left\{ \tau'_i \mid \tau'_i = \mathcal{M}(\tau_i, m) \right\}_{j=1}^T$
    \State $\theta^{\text{merged}} \leftarrow \theta_0 + w \sum_{i=1}^{T} \tau_i'$ \Comment{Task Arithmetic}
    \State \textbf{Return:} merged weights $\theta^{\text{merged}}$.
  \end{algorithmic}
\end{algorithm}
\begin{algorithm}[!tbp]
  \caption{Concrete AdaMerging}
  \label{alg:concrete_adamerging}
  \small
  \begin{algorithmic}
    \State {\bfseries Input:} pre-trained weights $\theta_0$,
    fine-tuned task vectors $\mathcal{T}=\{\tau_i\}_{i=1}^T$, AdaMerging weights $w$, learning rate $\alpha$.
    \\ \vskip -6pt \hrulefill
    \State $\mathbf{m} \gets \text{MetaLearnMask}(\theta_0, \mathcal{T})$ (Algorithm~\ref{alg:meta-learn_mask})
    \State Sample a mask $m$ from $\mathbf{m}$
    \State $\mathcal{T}' \leftarrow \left\{ \tau'_i \mid \tau'_i = \mathcal{M}(\tau_i, m) \right\}_{j=1}^T$
    \Repeat
    \State sample a batch of $N$ unlabeled data $\mathcal{D}$
    \State $\theta^{\text{merged}} \leftarrow \mathcal{A}_{\text{AdaMerging}}(\theta_0, \mathcal{T}'; w)$ \Comment{Eq.(\ref{eq:adamerging})}
    \State $w \leftarrow w - \alpha \nabla_{w} \mathcal{L}_{\text{entropy}}(\theta^{\text{merged}} ; \mathcal{D})$
    \Until {convergence}
    \State $\theta^{\text{merged}} \leftarrow \mathcal{A}_{\text{AdaMerging}}(\theta_0, \mathcal{T}'; w)$
    \State \textbf{Return:} merged weights $\theta^{\text{merged}}$.
  \end{algorithmic}
\end{algorithm}

Algorithm~\ref{alg:concrete_task_arithmetic} and Algorithm~\ref{alg:concrete_adamerging} demonstrate how our approach can be combined with Task Arithmetic and AdaMerging, respectively.
The combination with our method as ``Concrete Task Arithmetic'' and ``Concrete AdaMerging'', highlighting the specific adaptations and enhancements resulting from the incorporation of our approach.

\section{Model Fine-Tuning Details}
\label{appendix:model_fine_tuning_details}


\textbf{Model, Evaluation Tasks, and Metrics}.
We conduct experiments on diverse tasks spanning both vision and natural language domains.
The model, evaluation datasets and metrics settings are as follows:
\begin{itemize}
  \item For vision tasks, we utilize CLIP~\cite{radford2021learning} as our pre-trained models.
        The downstream tasks are SUN397~\cite{xiao_sun_2010}, Stanford Cars~\cite{krause3DObjectRepresentations2013}, RESISC45~\cite{chengRemoteSensingImage2017}, EuroSAT~\cite{helber2018introducing}, SVHN~\cite{netzer_reading_2021}, GTSRB~\cite{stallkamp_man_2012}, MNIST~\cite{lecunGradientbasedLearningApplied1998} and DTD~\cite{cimpoiDescribingTexturesWild2014}.
        We report top-1 accuracy as a performance metric.
  \item For NLP tasks, we utilize Flan-T5 \cite{chungScalingInstructionFinetunedLanguage2024} as our pre-trained language model.
        For fine-tuning, we employ the flan-t5 models on eight tasks derived from the GLUE benchmark \cite{wangGLUEMultiTaskBenchmark2018} with the same random seed $42$ to initialize the parameter-efficient models. These tasks are CoLA, MNLI, MRPC, QNLI, QQP, RTE, SST2, and STSB.
        We report Spearman's $\rho$ for STSB and accuracy for others.
\end{itemize}

The Flan-T5 models are LoRA fine-tuned with hyperparameter $r=16$ and $\alpha=32$~\cite{huLoRALowRankAdaptation2021a}.
We set the learning rate to $4e-5$ and maintained a consistent batch size of $16$ across all tasks.
The fine-tuning process for Flan-T5-base and Flan-T5-large involved 2000 steps on each downstream task.

\begin{figure}[!tb]
  \begin{center}
    \begin{subfigure}[b]{0.8\linewidth}
      \centering
      \includegraphics[width=10cm]{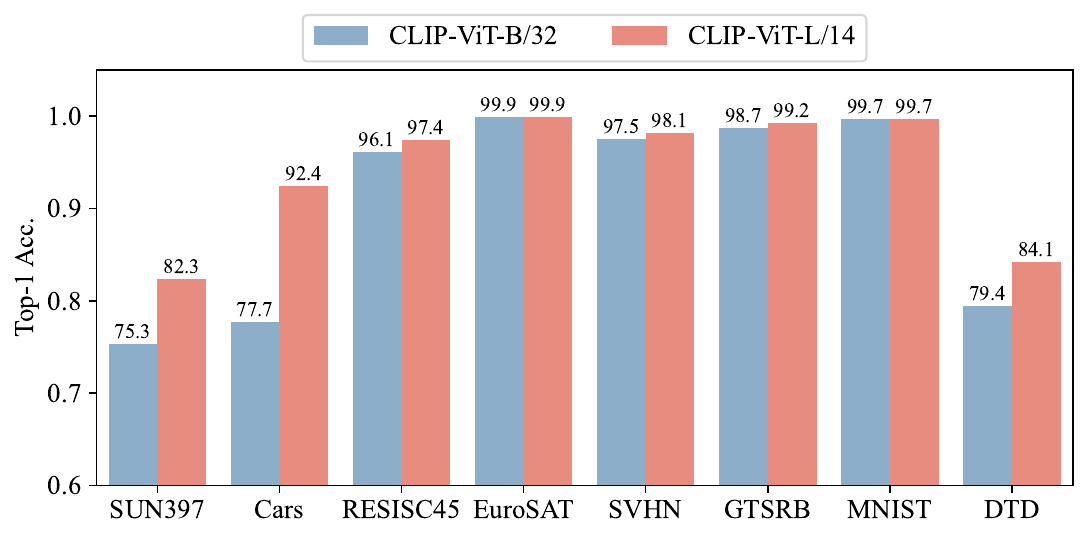}
      \caption{Individual performance of fine-tuned CLIP-ViT-B/32 and CLIP-ViT-L/14 models.}
      \label{fig:clip_individuals}
    \end{subfigure}
    \begin{subfigure}[b]{0.8\linewidth}
      \centering
      \includegraphics[width=10cm]{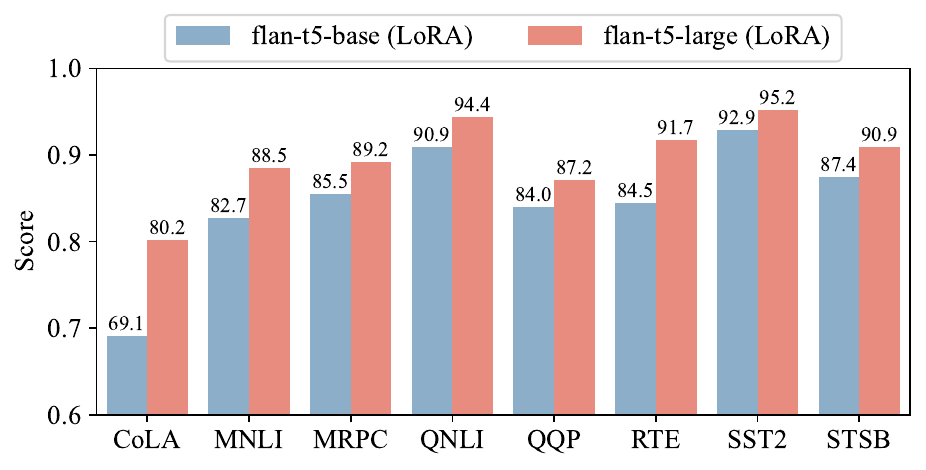}
      \caption{Individual performance of LoRA fine-tuned Flan-T5-base and Flan-T5-large models.}
      \label{fig:flan-t5_individuals}
    \end{subfigure}
    \caption{
      \textbf{Individual performance of fine-tuned models.}
      Here we present a visual comparison of the performance of different fine-tuned models.
      The top figure compares the performance of the CLIP-ViT-B/32 and CLIP-ViT-L/14 models on image classification tasks,
      while the bottom figure compares the performance of the LoRA fine-tuned Flan-T5-base and Flan-T5-large models on tasks from GLUE benchmark.
    }
  \end{center}
\end{figure}

Tables~\ref{table:clip-vit-b-32_individuals} and~\ref{table:clip-vit-l-14_individuals}, alongside Figure~\ref{fig:clip_individuals}, present detailed performance metrics for the CLIP-ViT-B/32 and CLIP-ViT-L/14 models, respectively. These visual and tabular data representations dissect the performance impact that both pre-training and subsequent fine-tuning have on the models' task execution.
The data spans a variety of downstream tasks, offering a comprehensive view of the models' capabilities.
Similarly, for the realm of natural language processing models, Tables~\ref{table:flan-t5-base_individuals} and \ref{table:flan-t5-large_individuals}, alongside Figure~\ref{fig:flan-t5_individuals}, provide an insight into the individual performance transformations that occur in the Flan-T5-base and Flan-T5-large models post LoRA fine-tuning.
As evidenced by the highlighted diagonal cells, it is apparent that the fine-tuning process has a significant impact on the models' abilities to perform specific tasks on both vision and language tasks.

\begin{table}[!tb]
  \caption{Individual performance of pre-trained and fine-tuned CLIP-ViT-B/32 models.}
  \label{table:clip-vit-b-32_individuals}
  \resizebox{\linewidth}{!}{%
    \centering
    \begin{tabular}{l|cccccccc}
      \toprule
      \textbf{Model} & \textbf{SUN397}                                                 & \textbf{Cars}                                                   & \textbf{RESISC45}                                               & \textbf{EuroSAT}                                                & \textbf{SVHN}                                                   & \textbf{GTSRB}                                                  & \textbf{MNIST}                                                  & \textbf{DTD}                                                    \\\midrule
      pre-trained    & \cellcolor[HTML]{20a386}\textcolor[HTML]{f1f1f1}{63.2}          & \cellcolor[HTML]{1f998a}\textcolor[HTML]{f1f1f1}{59.6}          & \cellcolor[HTML]{24878e}\textcolor[HTML]{f1f1f1}{60.2}          & \cellcolor[HTML]{2c718e}\textcolor[HTML]{f1f1f1}{45.0}          & \cellcolor[HTML]{453781}\textcolor[HTML]{f1f1f1}{31.6}          & \cellcolor[HTML]{404688}\textcolor[HTML]{f1f1f1}{32.6}          & \cellcolor[HTML]{443983}\textcolor[HTML]{f1f1f1}{48.3}          & \cellcolor[HTML]{31668e}\textcolor[HTML]{f1f1f1}{44.4}          \\
      SUN397         & \textbf{\cellcolor[HTML]{fde725}\textcolor[HTML]{000000}{75.3}} & \cellcolor[HTML]{39568c}\textcolor[HTML]{f1f1f1}{49.2}          & \cellcolor[HTML]{2d718e}\textcolor[HTML]{f1f1f1}{54.2}          & \cellcolor[HTML]{287d8e}\textcolor[HTML]{f1f1f1}{49.4}          & \cellcolor[HTML]{482979}\textcolor[HTML]{f1f1f1}{28.3}          & \cellcolor[HTML]{443b84}\textcolor[HTML]{f1f1f1}{29.7}          & \cellcolor[HTML]{433e85}\textcolor[HTML]{f1f1f1}{49.1}          & \cellcolor[HTML]{3c508b}\textcolor[HTML]{f1f1f1}{40.1}          \\
      Cars           & \cellcolor[HTML]{31678e}\textcolor[HTML]{f1f1f1}{55.9}          & \textbf{\cellcolor[HTML]{fde725}\textcolor[HTML]{000000}{77.7}} & \cellcolor[HTML]{31668e}\textcolor[HTML]{f1f1f1}{51.2}          & \cellcolor[HTML]{34618d}\textcolor[HTML]{f1f1f1}{39.6}          & \cellcolor[HTML]{472d7b}\textcolor[HTML]{f1f1f1}{29.4}          & \cellcolor[HTML]{433d84}\textcolor[HTML]{f1f1f1}{30.2}          & \cellcolor[HTML]{3e4a89}\textcolor[HTML]{f1f1f1}{51.8}          & \cellcolor[HTML]{3e4989}\textcolor[HTML]{f1f1f1}{38.8}          \\
      RESISC45       & \cellcolor[HTML]{414487}\textcolor[HTML]{f1f1f1}{52.2}          & \cellcolor[HTML]{3f4788}\textcolor[HTML]{f1f1f1}{47.2}          & \textbf{\cellcolor[HTML]{fde725}\textcolor[HTML]{000000}{96.1}} & \cellcolor[HTML]{21918c}\textcolor[HTML]{f1f1f1}{56.3}          & \cellcolor[HTML]{481769}\textcolor[HTML]{f1f1f1}{24.2}          & \cellcolor[HTML]{482071}\textcolor[HTML]{f1f1f1}{22.5}          & \cellcolor[HTML]{424086}\textcolor[HTML]{f1f1f1}{49.6}          & \cellcolor[HTML]{46307e}\textcolor[HTML]{f1f1f1}{34.7}          \\
      EuroSAT        & \cellcolor[HTML]{433d84}\textcolor[HTML]{f1f1f1}{51.6}          & \cellcolor[HTML]{443983}\textcolor[HTML]{f1f1f1}{45.3}          & \cellcolor[HTML]{471164}\textcolor[HTML]{f1f1f1}{32.5}          & \textbf{\cellcolor[HTML]{fde725}\textcolor[HTML]{000000}{99.9}} & \cellcolor[HTML]{440154}\textcolor[HTML]{f1f1f1}{19.3}          & \cellcolor[HTML]{472e7c}\textcolor[HTML]{f1f1f1}{26.1}          & \cellcolor[HTML]{440154}\textcolor[HTML]{f1f1f1}{37.9}          & \cellcolor[HTML]{453882}\textcolor[HTML]{f1f1f1}{35.9}          \\
      SVHN           & \cellcolor[HTML]{482475}\textcolor[HTML]{f1f1f1}{49.3}          & \cellcolor[HTML]{470d60}\textcolor[HTML]{f1f1f1}{40.2}          & \cellcolor[HTML]{450457}\textcolor[HTML]{f1f1f1}{30.3}          & \cellcolor[HTML]{440154}\textcolor[HTML]{f1f1f1}{12.7}          & \textbf{\cellcolor[HTML]{fde725}\textcolor[HTML]{000000}{97.5}} & \cellcolor[HTML]{424186}\textcolor[HTML]{f1f1f1}{31.4}          & \cellcolor[HTML]{6ccd5a}\textcolor[HTML]{000000}{85.7}          & \cellcolor[HTML]{460a5d}\textcolor[HTML]{f1f1f1}{28.7}          \\
      GTSRB          & \cellcolor[HTML]{440154}\textcolor[HTML]{f1f1f1}{46.4}          & \cellcolor[HTML]{440154}\textcolor[HTML]{f1f1f1}{38.9}          & \cellcolor[HTML]{440154}\textcolor[HTML]{f1f1f1}{29.5}          & \cellcolor[HTML]{482677}\textcolor[HTML]{f1f1f1}{22.0}          & \cellcolor[HTML]{33638d}\textcolor[HTML]{f1f1f1}{43.9}          & \textbf{\cellcolor[HTML]{fde725}\textcolor[HTML]{000000}{98.7}} & \cellcolor[HTML]{460a5d}\textcolor[HTML]{f1f1f1}{39.5}          & \cellcolor[HTML]{46075a}\textcolor[HTML]{f1f1f1}{28.5}          \\
      MNIST          & \cellcolor[HTML]{482374}\textcolor[HTML]{f1f1f1}{49.2}          & \cellcolor[HTML]{470d60}\textcolor[HTML]{f1f1f1}{40.2}          & \cellcolor[HTML]{481769}\textcolor[HTML]{f1f1f1}{33.5}          & \cellcolor[HTML]{482173}\textcolor[HTML]{f1f1f1}{20.7}          & \cellcolor[HTML]{2b748e}\textcolor[HTML]{f1f1f1}{49.2}          & \cellcolor[HTML]{440154}\textcolor[HTML]{f1f1f1}{15.3}          & \textbf{\cellcolor[HTML]{fde725}\textcolor[HTML]{000000}{99.7}} & \cellcolor[HTML]{440154}\textcolor[HTML]{f1f1f1}{27.4}          \\
      DTD            & \cellcolor[HTML]{472f7d}\textcolor[HTML]{f1f1f1}{50.4}          & \cellcolor[HTML]{38588c}\textcolor[HTML]{f1f1f1}{49.4}          & \cellcolor[HTML]{423f85}\textcolor[HTML]{f1f1f1}{41.9}          & \cellcolor[HTML]{3c508b}\textcolor[HTML]{f1f1f1}{33.9}          & \cellcolor[HTML]{472c7a}\textcolor[HTML]{f1f1f1}{28.9}          & \cellcolor[HTML]{482173}\textcolor[HTML]{f1f1f1}{22.8}          & \cellcolor[HTML]{453781}\textcolor[HTML]{f1f1f1}{47.8}          & \textbf{\cellcolor[HTML]{fde725}\textcolor[HTML]{000000}{79.4}} \\\bottomrule
    \end{tabular}
  }
\end{table}

\begin{table}[!tb]
  \caption{Individual performance of pre-trained and fine-tuned CLIP-ViT-L/14 models.}
  \label{table:clip-vit-l-14_individuals}
  \resizebox{\linewidth}{!}{%
    \centering
    \begin{tabular}{l|cccccccc}
      \toprule
      \textbf{Model} & \textbf{SUN397}                                                 & \textbf{Cars}                                                   & \textbf{RESISC45}                                               & \textbf{EuroSAT}                                                & \textbf{SVHN}                                                   & \textbf{GTSRB}                                                  & \textbf{MNIST}                                                  & \textbf{DTD}                                                    \\\midrule
      pre-trained    & \cellcolor[HTML]{463480}\textcolor[HTML]{f1f1f1}{68.2}          & \cellcolor[HTML]{31678e}\textcolor[HTML]{f1f1f1}{77.9}          & \cellcolor[HTML]{228c8d}\textcolor[HTML]{f1f1f1}{71.3}          & \cellcolor[HTML]{297a8e}\textcolor[HTML]{f1f1f1}{61.3}          & \cellcolor[HTML]{424086}\textcolor[HTML]{f1f1f1}{58.4}          & \cellcolor[HTML]{482979}\textcolor[HTML]{f1f1f1}{50.6}          & \cellcolor[HTML]{306a8e}\textcolor[HTML]{f1f1f1}{76.4}          & \cellcolor[HTML]{414287}\textcolor[HTML]{f1f1f1}{55.4}          \\
      SUN397         & \textbf{\cellcolor[HTML]{fde725}\textcolor[HTML]{000000}{82.3}} & \cellcolor[HTML]{46075a}\textcolor[HTML]{f1f1f1}{71.2}          & \cellcolor[HTML]{2e6d8e}\textcolor[HTML]{f1f1f1}{64.7}          & \cellcolor[HTML]{34618d}\textcolor[HTML]{f1f1f1}{54.6}          & \cellcolor[HTML]{481a6c}\textcolor[HTML]{f1f1f1}{52.5}          & \cellcolor[HTML]{471365}\textcolor[HTML]{f1f1f1}{46.9}          & \cellcolor[HTML]{34618d}\textcolor[HTML]{f1f1f1}{75.1}          & \cellcolor[HTML]{482475}\textcolor[HTML]{f1f1f1}{51.9}          \\
      Cars           & \cellcolor[HTML]{482173}\textcolor[HTML]{f1f1f1}{67.2}          & \textbf{\cellcolor[HTML]{fde725}\textcolor[HTML]{000000}{92.4}} & \cellcolor[HTML]{277f8e}\textcolor[HTML]{f1f1f1}{68.4}          & \cellcolor[HTML]{31688e}\textcolor[HTML]{f1f1f1}{56.4}          & \cellcolor[HTML]{433d84}\textcolor[HTML]{f1f1f1}{57.8}          & \cellcolor[HTML]{481c6e}\textcolor[HTML]{f1f1f1}{48.4}          & \cellcolor[HTML]{39558c}\textcolor[HTML]{f1f1f1}{73.7}          & \cellcolor[HTML]{414487}\textcolor[HTML]{f1f1f1}{55.6}          \\
      RESISC45       & \cellcolor[HTML]{470e61}\textcolor[HTML]{f1f1f1}{66.3}          & \cellcolor[HTML]{470d60}\textcolor[HTML]{f1f1f1}{71.5}          & \textbf{\cellcolor[HTML]{fde725}\textcolor[HTML]{000000}{97.4}} & \cellcolor[HTML]{2e6d8e}\textcolor[HTML]{f1f1f1}{57.7}          & \cellcolor[HTML]{481b6d}\textcolor[HTML]{f1f1f1}{52.7}          & \cellcolor[HTML]{481d6f}\textcolor[HTML]{f1f1f1}{48.5}          & \cellcolor[HTML]{297b8e}\textcolor[HTML]{f1f1f1}{78.9}          & \cellcolor[HTML]{482677}\textcolor[HTML]{f1f1f1}{52.1}          \\
      EuroSAT        & \cellcolor[HTML]{440154}\textcolor[HTML]{f1f1f1}{65.7}          & \cellcolor[HTML]{440154}\textcolor[HTML]{f1f1f1}{70.8}          & \cellcolor[HTML]{440154}\textcolor[HTML]{f1f1f1}{46.9}          & \textbf{\cellcolor[HTML]{fde725}\textcolor[HTML]{000000}{99.9}} & \cellcolor[HTML]{440154}\textcolor[HTML]{f1f1f1}{49.1}          & \cellcolor[HTML]{471164}\textcolor[HTML]{f1f1f1}{46.6}          & \cellcolor[HTML]{34608d}\textcolor[HTML]{f1f1f1}{75.0}          & \cellcolor[HTML]{440154}\textcolor[HTML]{f1f1f1}{48.4}          \\
      SVHN           & \cellcolor[HTML]{482979}\textcolor[HTML]{f1f1f1}{67.6}          & \cellcolor[HTML]{440154}\textcolor[HTML]{f1f1f1}{70.8}          & \cellcolor[HTML]{2f6b8e}\textcolor[HTML]{f1f1f1}{64.4}          & \cellcolor[HTML]{471063}\textcolor[HTML]{f1f1f1}{37.0}          & \textbf{\cellcolor[HTML]{fde725}\textcolor[HTML]{000000}{98.1}} & \cellcolor[HTML]{481668}\textcolor[HTML]{f1f1f1}{47.2}          & \cellcolor[HTML]{60ca60}\textcolor[HTML]{000000}{91.1}          & \cellcolor[HTML]{482475}\textcolor[HTML]{f1f1f1}{51.9}          \\
      GTSRB          & \cellcolor[HTML]{471164}\textcolor[HTML]{f1f1f1}{66.5}          & \cellcolor[HTML]{472c7a}\textcolor[HTML]{f1f1f1}{73.4}          & \cellcolor[HTML]{2e6d8e}\textcolor[HTML]{f1f1f1}{64.8}          & \cellcolor[HTML]{440154}\textcolor[HTML]{f1f1f1}{34.5}          & \cellcolor[HTML]{3a538b}\textcolor[HTML]{f1f1f1}{61.6}          & \textbf{\cellcolor[HTML]{fde725}\textcolor[HTML]{000000}{99.2}} & \cellcolor[HTML]{1f968b}\textcolor[HTML]{f1f1f1}{82.9}          & \cellcolor[HTML]{482979}\textcolor[HTML]{f1f1f1}{52.5}          \\
      MNIST          & \cellcolor[HTML]{443983}\textcolor[HTML]{f1f1f1}{68.5}          & \cellcolor[HTML]{482576}\textcolor[HTML]{f1f1f1}{73.0}          & \cellcolor[HTML]{2d718e}\textcolor[HTML]{f1f1f1}{65.5}          & \cellcolor[HTML]{46307e}\textcolor[HTML]{f1f1f1}{43.4}          & \cellcolor[HTML]{2e6e8e}\textcolor[HTML]{f1f1f1}{66.5}          & \cellcolor[HTML]{440154}\textcolor[HTML]{f1f1f1}{44.1}          & \textbf{\cellcolor[HTML]{fde725}\textcolor[HTML]{000000}{99.7}} & \cellcolor[HTML]{472a7a}\textcolor[HTML]{f1f1f1}{52.6}          \\
      DTD            & \cellcolor[HTML]{46075a}\textcolor[HTML]{f1f1f1}{66.0}          & \cellcolor[HTML]{443a83}\textcolor[HTML]{f1f1f1}{74.4}          & \cellcolor[HTML]{287c8e}\textcolor[HTML]{f1f1f1}{68.0}          & \cellcolor[HTML]{31668e}\textcolor[HTML]{f1f1f1}{55.7}          & \cellcolor[HTML]{471164}\textcolor[HTML]{f1f1f1}{51.3}          & \cellcolor[HTML]{481a6c}\textcolor[HTML]{f1f1f1}{47.8}          & \cellcolor[HTML]{440154}\textcolor[HTML]{f1f1f1}{64.3}          & \textbf{\cellcolor[HTML]{fde725}\textcolor[HTML]{000000}{84.1}} \\
      \bottomrule
    \end{tabular}
  }
\end{table}

\begin{table}[!tb]
  \caption{Individual performance of pre-trained-tuned Flan-tuned Flan-T5-base models.}
  \label{table:flan-t5-base_individuals}
  \resizebox{\linewidth}{!}{%
    \centering
    \begin{tabular}{l|cccccccc}
      \toprule
      \textbf{Model} & \textbf{CoLA}                                                   & \textbf{MNLI}                                                   & \textbf{MRPC}                                                   & \textbf{QNLI}                                                   & \textbf{QQP}                                                    & \textbf{RTE}                                                    & \textbf{SST2}                                                   & \textbf{STSB}                                                   \\\midrule
      pre-trained    & \cellcolor[HTML]{dde318}\textcolor[HTML]{000000}{69.1}          & \cellcolor[HTML]{3aba76}\textcolor[HTML]{f1f1f1}{56.5}          & \cellcolor[HTML]{228d8d}\textcolor[HTML]{f1f1f1}{76.2}          & \cellcolor[HTML]{44bf70}\textcolor[HTML]{f1f1f1}{88.4}          & \cellcolor[HTML]{1e9c89}\textcolor[HTML]{f1f1f1}{82.1}          & \cellcolor[HTML]{4ec36b}\textcolor[HTML]{000000}{80.1}          & \cellcolor[HTML]{22a785}\textcolor[HTML]{f1f1f1}{91.2}          & \cellcolor[HTML]{3c508b}\textcolor[HTML]{f1f1f1}{62.2}          \\
      CoLA           & \cellcolor[HTML]{dde318}\textcolor[HTML]{000000}{69.1}          & \cellcolor[HTML]{23898e}\textcolor[HTML]{f1f1f1}{39.9}          & \cellcolor[HTML]{27808e}\textcolor[HTML]{f1f1f1}{75.2}          & \cellcolor[HTML]{70cf57}\textcolor[HTML]{000000}{89.1}          & \cellcolor[HTML]{365d8d}\textcolor[HTML]{f1f1f1}{81.1}          & \cellcolor[HTML]{93d741}\textcolor[HTML]{000000}{81.9}          & \cellcolor[HTML]{228d8d}\textcolor[HTML]{f1f1f1}{90.7}          & \cellcolor[HTML]{440154}\textcolor[HTML]{f1f1f1}{54.0}          \\
      MNLI           & \textbf{\cellcolor[HTML]{fde725}\textcolor[HTML]{000000}{69.4}} & \textbf{\cellcolor[HTML]{fde725}\textcolor[HTML]{000000}{82.7}} & \cellcolor[HTML]{2f6c8e}\textcolor[HTML]{f1f1f1}{73.8}          & \cellcolor[HTML]{7ad151}\textcolor[HTML]{000000}{89.3}          & \cellcolor[HTML]{1f968b}\textcolor[HTML]{f1f1f1}{82.0}          & \cellcolor[HTML]{38b977}\textcolor[HTML]{f1f1f1}{79.4}          & \cellcolor[HTML]{1f9a8a}\textcolor[HTML]{f1f1f1}{90.9}          & \cellcolor[HTML]{287d8e}\textcolor[HTML]{f1f1f1}{68.1}          \\
      MRPC           & \cellcolor[HTML]{440154}\textcolor[HTML]{f1f1f1}{64.0}          & \cellcolor[HTML]{1f988b}\textcolor[HTML]{f1f1f1}{44.9}          & \textbf{\cellcolor[HTML]{fde725}\textcolor[HTML]{000000}{85.5}} & \cellcolor[HTML]{440154}\textcolor[HTML]{f1f1f1}{82.6}          & \cellcolor[HTML]{375a8c}\textcolor[HTML]{f1f1f1}{81.0}          & \cellcolor[HTML]{440154}\textcolor[HTML]{f1f1f1}{69.0}          & \cellcolor[HTML]{440154}\textcolor[HTML]{f1f1f1}{88.6}          & \cellcolor[HTML]{20a486}\textcolor[HTML]{f1f1f1}{73.6}          \\
      QNLI           & \cellcolor[HTML]{c5e021}\textcolor[HTML]{000000}{68.9}          & \cellcolor[HTML]{29af7f}\textcolor[HTML]{f1f1f1}{52.7}          & \cellcolor[HTML]{20938c}\textcolor[HTML]{f1f1f1}{76.7}          & \textbf{\cellcolor[HTML]{fde725}\textcolor[HTML]{000000}{90.9}} & \cellcolor[HTML]{4cc26c}\textcolor[HTML]{000000}{82.8}          & \cellcolor[HTML]{42be71}\textcolor[HTML]{f1f1f1}{79.8}          & \cellcolor[HTML]{38b977}\textcolor[HTML]{f1f1f1}{91.5}          & \cellcolor[HTML]{25838e}\textcolor[HTML]{f1f1f1}{68.9}          \\
      QQP            & \cellcolor[HTML]{433d84}\textcolor[HTML]{f1f1f1}{65.0}          & \cellcolor[HTML]{31b57b}\textcolor[HTML]{f1f1f1}{54.6}          & \cellcolor[HTML]{24868e}\textcolor[HTML]{f1f1f1}{75.7}          & \cellcolor[HTML]{69cd5b}\textcolor[HTML]{000000}{89.0}          & \textbf{\cellcolor[HTML]{fde725}\textcolor[HTML]{000000}{84.0}} & \cellcolor[HTML]{84d44b}\textcolor[HTML]{000000}{81.6}          & \cellcolor[HTML]{228d8d}\textcolor[HTML]{f1f1f1}{90.7}          & \cellcolor[HTML]{2ab07f}\textcolor[HTML]{f1f1f1}{75.3}          \\
      RTE            & \cellcolor[HTML]{453882}\textcolor[HTML]{f1f1f1}{64.9}          & \cellcolor[HTML]{26ad81}\textcolor[HTML]{f1f1f1}{51.8}          & \cellcolor[HTML]{482677}\textcolor[HTML]{f1f1f1}{69.4}          & \cellcolor[HTML]{75d054}\textcolor[HTML]{000000}{89.2}          & \cellcolor[HTML]{440154}\textcolor[HTML]{f1f1f1}{79.8}          & \textbf{\cellcolor[HTML]{fde725}\textcolor[HTML]{000000}{84.5}} & \cellcolor[HTML]{24868e}\textcolor[HTML]{f1f1f1}{90.6}          & \cellcolor[HTML]{228c8d}\textcolor[HTML]{f1f1f1}{70.1}          \\
      SST2           & \cellcolor[HTML]{73d056}\textcolor[HTML]{000000}{68.3}          & \cellcolor[HTML]{3aba76}\textcolor[HTML]{f1f1f1}{56.6}          & \cellcolor[HTML]{238a8d}\textcolor[HTML]{f1f1f1}{76.0}          & \cellcolor[HTML]{48c16e}\textcolor[HTML]{f1f1f1}{88.5}          & \cellcolor[HTML]{95d840}\textcolor[HTML]{000000}{83.4}          & \cellcolor[HTML]{42be71}\textcolor[HTML]{f1f1f1}{79.8}          & \textbf{\cellcolor[HTML]{fde725}\textcolor[HTML]{000000}{92.9}} & \cellcolor[HTML]{3a538b}\textcolor[HTML]{f1f1f1}{62.6}          \\
      STSB           & \cellcolor[HTML]{34608d}\textcolor[HTML]{f1f1f1}{65.7}          & \cellcolor[HTML]{440154}\textcolor[HTML]{f1f1f1}{1.7}           & \cellcolor[HTML]{440154}\textcolor[HTML]{f1f1f1}{67.4}          & \cellcolor[HTML]{7ad151}\textcolor[HTML]{000000}{89.3}          & \cellcolor[HTML]{481b6d}\textcolor[HTML]{f1f1f1}{80.1}          & \cellcolor[HTML]{42be71}\textcolor[HTML]{f1f1f1}{79.8}          & \cellcolor[HTML]{20938c}\textcolor[HTML]{f1f1f1}{90.8}          & \textbf{\cellcolor[HTML]{fde725}\textcolor[HTML]{000000}{87.4}} \\\bottomrule
    \end{tabular}
  }
\end{table}

\begin{table}[!tb]
  \caption{Individual performance of pre-trained and LoRA fine-tuned Flan-T5-large models.}
  \label{table:flan-t5-large_individuals}
  \centering
  \resizebox{\linewidth}{!}{%
    \centering
    \begin{tabular}{l|cccccccc}
      \toprule
      \textbf{Model} & \textbf{CoLA}                                                   & \textbf{MNLI}                                                   & \textbf{MRPC}                                                   & \textbf{QNLI}                                                   & \textbf{QQP}                                                    & \textbf{RTE}                                                    & \textbf{SST2}                                                   & \textbf{STSB}                                                   \\\midrule
      pre-trained    & \cellcolor[HTML]{471063}\textcolor[HTML]{f1f1f1}{73.7}          & \cellcolor[HTML]{2e6d8e}\textcolor[HTML]{f1f1f1}{56.6}          & \cellcolor[HTML]{2eb37c}\textcolor[HTML]{f1f1f1}{82.4}          & \cellcolor[HTML]{481b6d}\textcolor[HTML]{f1f1f1}{91.1}          & \cellcolor[HTML]{20928c}\textcolor[HTML]{f1f1f1}{85.5}          & \cellcolor[HTML]{3a548c}\textcolor[HTML]{f1f1f1}{85.6}          & \cellcolor[HTML]{2b748e}\textcolor[HTML]{f1f1f1}{94.3}          & \cellcolor[HTML]{34608d}\textcolor[HTML]{f1f1f1}{87.5}          \\
      CoLA           & \textbf{\cellcolor[HTML]{fde725}\textcolor[HTML]{000000}{80.2}} & \cellcolor[HTML]{355e8d}\textcolor[HTML]{f1f1f1}{53.9}          & \cellcolor[HTML]{22a785}\textcolor[HTML]{f1f1f1}{81.4}          & \cellcolor[HTML]{440154}\textcolor[HTML]{f1f1f1}{90.8}          & \cellcolor[HTML]{3f4788}\textcolor[HTML]{f1f1f1}{84.5}          & \cellcolor[HTML]{482071}\textcolor[HTML]{f1f1f1}{84.1}          & \cellcolor[HTML]{453581}\textcolor[HTML]{f1f1f1}{93.9}          & \cellcolor[HTML]{3e4a89}\textcolor[HTML]{f1f1f1}{87.1}          \\
      MNLI           & \cellcolor[HTML]{471063}\textcolor[HTML]{f1f1f1}{73.7}          & \textbf{\cellcolor[HTML]{fde725}\textcolor[HTML]{000000}{88.5}} & \cellcolor[HTML]{287c8e}\textcolor[HTML]{f1f1f1}{77.9}          & \cellcolor[HTML]{26818e}\textcolor[HTML]{f1f1f1}{92.4}          & \cellcolor[HTML]{277e8e}\textcolor[HTML]{f1f1f1}{85.2}          & \cellcolor[HTML]{1f958b}\textcolor[HTML]{f1f1f1}{87.7}          & \cellcolor[HTML]{24878e}\textcolor[HTML]{f1f1f1}{94.4}          & \cellcolor[HTML]{46307e}\textcolor[HTML]{f1f1f1}{86.7}          \\
      MRPC           & \cellcolor[HTML]{33638d}\textcolor[HTML]{f1f1f1}{75.6}          & \cellcolor[HTML]{38588c}\textcolor[HTML]{f1f1f1}{52.6}          & \textbf{\cellcolor[HTML]{fde725}\textcolor[HTML]{000000}{89.2}} & \cellcolor[HTML]{228d8d}\textcolor[HTML]{f1f1f1}{92.6}          & \cellcolor[HTML]{433d84}\textcolor[HTML]{f1f1f1}{84.4}          & \cellcolor[HTML]{2f6c8e}\textcolor[HTML]{f1f1f1}{86.3}          & \cellcolor[HTML]{2b748e}\textcolor[HTML]{f1f1f1}{94.3}          & \cellcolor[HTML]{481769}\textcolor[HTML]{f1f1f1}{86.3}          \\
      QNLI           & \cellcolor[HTML]{450559}\textcolor[HTML]{f1f1f1}{73.5}          & \cellcolor[HTML]{33628d}\textcolor[HTML]{f1f1f1}{54.5}          & \cellcolor[HTML]{37b878}\textcolor[HTML]{f1f1f1}{82.8}          & \textbf{\cellcolor[HTML]{fde725}\textcolor[HTML]{000000}{94.4}} & \cellcolor[HTML]{22a884}\textcolor[HTML]{f1f1f1}{85.8}          & \cellcolor[HTML]{3f4889}\textcolor[HTML]{f1f1f1}{85.2}          & \cellcolor[HTML]{440154}\textcolor[HTML]{f1f1f1}{93.7}          & \cellcolor[HTML]{3e4989}\textcolor[HTML]{f1f1f1}{87.1}          \\
      QQP            & \cellcolor[HTML]{481f70}\textcolor[HTML]{f1f1f1}{74.0}          & \cellcolor[HTML]{355e8d}\textcolor[HTML]{f1f1f1}{53.8}          & \cellcolor[HTML]{37b878}\textcolor[HTML]{f1f1f1}{82.8}          & \cellcolor[HTML]{23898e}\textcolor[HTML]{f1f1f1}{92.5}          & \textbf{\cellcolor[HTML]{fde725}\textcolor[HTML]{000000}{87.2}} & \cellcolor[HTML]{3a548c}\textcolor[HTML]{f1f1f1}{85.6}          & \cellcolor[HTML]{1f998a}\textcolor[HTML]{f1f1f1}{94.5}          & \cellcolor[HTML]{24878e}\textcolor[HTML]{f1f1f1}{88.3}          \\
      RTE            & \cellcolor[HTML]{33638d}\textcolor[HTML]{f1f1f1}{75.6}          & \cellcolor[HTML]{2d718e}\textcolor[HTML]{f1f1f1}{57.5}          & \cellcolor[HTML]{440154}\textcolor[HTML]{f1f1f1}{69.9}          & \cellcolor[HTML]{1e9d89}\textcolor[HTML]{f1f1f1}{92.8}          & \cellcolor[HTML]{440154}\textcolor[HTML]{f1f1f1}{83.8}          & \textbf{\cellcolor[HTML]{fde725}\textcolor[HTML]{000000}{91.7}} & \cellcolor[HTML]{25ac82}\textcolor[HTML]{f1f1f1}{94.6}          & \cellcolor[HTML]{440154}\textcolor[HTML]{f1f1f1}{86.0}          \\
      SST2           & \cellcolor[HTML]{460b5e}\textcolor[HTML]{f1f1f1}{73.6}          & \cellcolor[HTML]{31668e}\textcolor[HTML]{f1f1f1}{55.3}          & \cellcolor[HTML]{2ab07f}\textcolor[HTML]{f1f1f1}{82.1}          & \cellcolor[HTML]{404588}\textcolor[HTML]{f1f1f1}{91.6}          & \cellcolor[HTML]{20938c}\textcolor[HTML]{f1f1f1}{85.5}          & \cellcolor[HTML]{3f4889}\textcolor[HTML]{f1f1f1}{85.2}          & \textbf{\cellcolor[HTML]{fde725}\textcolor[HTML]{000000}{95.2}} & \cellcolor[HTML]{433d84}\textcolor[HTML]{f1f1f1}{86.9}          \\
      STSB           & \cellcolor[HTML]{440154}\textcolor[HTML]{f1f1f1}{73.4}          & \cellcolor[HTML]{440154}\textcolor[HTML]{f1f1f1}{39.3}          & \cellcolor[HTML]{2ab07f}\textcolor[HTML]{f1f1f1}{82.1}          & \cellcolor[HTML]{21918c}\textcolor[HTML]{f1f1f1}{92.6}          & \cellcolor[HTML]{3fbc73}\textcolor[HTML]{f1f1f1}{86.1}          & \cellcolor[HTML]{440154}\textcolor[HTML]{f1f1f1}{83.4}          & \cellcolor[HTML]{3d4d8a}\textcolor[HTML]{f1f1f1}{94.0}          & \textbf{\cellcolor[HTML]{fde725}\textcolor[HTML]{000000}{90.9}} \\\bottomrule
    \end{tabular}
  }
\end{table}

\section{Multi-Task Model Fusion Details}
\label{appendix:model_fusion_details}

This section provides further implementation and configuration details for the baseline methods compared in our experiments alongside a straightforward taxonomy of these methods.
Subsequently, Appendix~\ref{appendix:task_arithmetic_based_methods} and Appendix~\ref{appendix:adamerging_based_methods} serve to thoroughly explicate task arithmetic-based methods and AdaMerging-based methods.
In these supplementary sections, we also provide a comprehensive analysis of the experimental results to afford clarity and facilitate a deeper understanding of the effectiveness of each multi-task strategy.

\subsection{Details About Baseline Methods}
\label{appendix:baseline_details}

\begin{itemize}
  \item \textbf{Simple Weight Average}:
        We average the weights of the models fine-tuned on different tasks, this method is also referred to as ModelSoups~\cite{wortsmanModelSoupsAveraging2022} in the literature.
        The averaged model is then evaluated on the validation set of each task.
        For full fine-tuned models, the weights of the models are averaged directly (i.e., let $\theta = {1/n \sum_{i=1}^{n} \theta_i}$). For LoRA fine-tuning, we average the weights of merged models, i.e., let $\theta = \theta_0 + 1/n\sum_i A_i B_i$.
  \item \textbf{Task Arithmetic}~\cite{ilharcoEditingModelsTask2023:b}:
        As defined in Section~\ref{subsec:preliminary}, the task vector is computed on the set of model parameters.
        We compute the task vector for each task and then add them to construct a multi-task vector.
        The multi-task vector is then multiplied by a scaling coefficient $\lambda$ element-wisely and added to the initial parameters of the pre-trained model to obtain a multi-task model, i.e. $\theta = \theta_0 + \lambda \sum_{i} (\theta_i - \theta_0)$ for full fine-tuning and $\phi = \phi_0 + \lambda \sum_{i} A_i B_i$ for LoRA fine-tuning, where $\lambda$ is a hyperparameter that the best-performing model is chosen on validation set.
        In our study,  $\lambda$ is chosen to be $0.3$.
  \item \textbf{Ties-Merging}~\cite{yadavResolvingInterferenceWhen2023:b}:
        Ties-merging is a method for merging multiple task-specific models into a single multi-task model.
        This algorithm follows three steps (trim, elect sign of parameters, and disjoint merge) to obtain a merged task vector $\nu$. Given the final merged task vector $\tau$, the final model is chosen in a similar way as task arithmetic, i.e. $\theta = \theta_0 + \lambda \tau$, where $\lambda$ is a hyperparameter that the best-performing model is chosen on the validation set.
        In our study, $\lambda$ is chosen to be $0.3$.
  \item \textbf{AdaMerging}~\cite{yangAdaMergingAdaptiveModel2024}:
        AdaMerging is an adaptive model merging method where it autonomously learns the coefficients for merging either on a task-wise or layer-wise basis, using entropy minimization on unlabeled test samples as a surrogate objective function to refine the merging coefficients.
        The task-wise AdaMerging is formulated as $\theta = \theta_0 + \sum_{i=1}^{n} \lambda_i \tau_i$ where $\lambda_k$ is the merging coefficient for the $k$-th task and $\tau_k$ is the task vector for the $k$-th task.
        The layer-wise AdaMerging is formulated as $\theta^l = \theta_0^l + \sum_{i=1}^{n} \lambda^{l}_{i} \tau^{l}_{i}$.
        In our study, we initialize the $\lambda$ to be $0.3$ for all tasks and layers.
\end{itemize}

\textbf{A simple taxonomy of model merging methods.} In this paper, we categorize the methods briefly into three groups: task arithmetic-based methods, AdaMerging-based methods, and others.
Task arithmetic-based methods involve constructing a merged task vector, which is then scaled by a single factor and added to the pre-trained model.
While AdaMerging-based methods assign separate task-wise or layer-wise weights to each task vector.
In the following subsections, we will present some details about the experimental results of Task arithmetic-based methods and AdaMerging-based methods.

\begin{table}[!tb]
  \caption{Multi-task performance when merging Flan-T5-large (LoRA fine-tuned) models on all eight tasks.}
  \label{table:flan-t5-large_lora}
  \resizebox{\linewidth}{!}{%
    \centering
    \begin{tabular}{lccccccccc}
      \toprule
      \textbf{Method}         & \textbf{CoLA} & \textbf{MNLI} & \textbf{MRPC} & \textbf{QNLI} & \textbf{QQP}  & \textbf{RTE}  & \textbf{SST2} & \textbf{STSB} & \textbf{Avg.} \\
      \midrule
      Pre-trained             & 73.7          & 56.6          & 82.4          & 91.1          & 85.5          & 85.6          & 94.3          & 87.5          & 82.1          \\
      Individual              & 80.2          & 88.5          & 89.2          & 94.4          & 87.2          & 91.7          & 95.2          & 90.9          & 89.6          \\
      Weight Averaging        & 74.6          & 84.3          & 84.1          & 92.8          & 86.3          & 87.4          & 94.8          & 88.0          & 86.5          \\
      \midrule
      \multicolumn{10}{c}{\textit{Task Arithmetic (TA)-Based}}                                                                                                                \\
      Task Arithmetic         & 76.9          & 85.4          & 85.3          & 93.9          & 85.8          & 88.1          & 95.2          & 87.8          & 87.3          \\
      Ties-Merging            & \textbf{77.1} & 85.1          & \textbf{86.3} & 93.9          & \textbf{86.0} & 87.7          & 95.1          & \textbf{88.0} & 87.4          \\
      \textbf{Concrete TA}    & 76.6          & \textbf{86.4} & 86.0          & 93.9          & 85.9          & \textbf{88.4} & 95.2          & 87.9          & \textbf{87.5} \\
      \midrule
      \multicolumn{10}{c}{\textit{Layer-wise AdaMerging (LW AM)-Based}}                                                                                                       \\
      LW AM                   & \textbf{76.7} & 87.6          & 84.8          & 93.8          & 85.9          & 88.1          & 95.2          & \textbf{88.6} & \textbf{87.6} \\
      \textbf{LW Concrete AM} & 76.1          & \textbf{87.7} & \textbf{85.5} & 93.8          & 85.9          & 88.1          & \textbf{95.4} & 87.1          & 87.5          \\
      \bottomrule
    \end{tabular}
  }
\end{table}

\subsection{Task Arithmetic-Based Methods}
\label{appendix:task_arithmetic_based_methods}

The core procedure in task arithmetic-based methods includes the generation of a composite task vector.
This composite task vector is an amalgamation of information relevant to each task, carefully constructed to preserve the distinctive features that define them.
When this vector is ready, it goes through a scaling step. After being scaled, this adjusted task vector is merged right into the pre-trained model's existing parameters.
We describe this process mathematically as:
\begin{equation}
  \theta = \theta_0 + \lambda \mathcal{A}(\mathcal{T}),
\end{equation}
where $\mathcal{A}: \mathbb{R}^{d\times n} \mapsto \mathbb{R}^d$ represents a merging algorithm for multiple tasks, $\mathcal{T}$ is a collection of task vectors $\{\tau_i\}_{i=1}^n$, and $d$ indicates the size of the model parameters.
It's akin to performing arithmetic on the knowledge of the model—adding new insights or amplifying existing ones.

One of the critical advantages of task arithmetic-based methods is the simplicity of the concept and its implementation.
By leveraging a single scaling coefficient, the method offers a direct and computationally efficient way to transfer knowledge and adapt models to handle various tasks.

Despite their elegant simplicity, these methods require careful consideration in the determination of the scaling coefficient.
The right balance ensures that the merged task vector harmoniously enhances the pre-trained model's knowledge without overshadowing the valuable insights the model has already acquired.
Balancing this interplay is crucial for the success of the task arithmetic approach.

Task arithmetic-based model merging thus stands out for its straightforwardness, ease of implementation, and potential to enable pre-trained models to excel across a wider array of tasks through a calculated, additive enhancement of their capabilities.
In the following paragraphs, we will dive deeper into the experimental results garnered from employing these methods, offering a clear insight into their practical implications and the extent of their applicability.

\begin{figure}[!tb]
  \begin{center}
    \begin{subfigure}[b]{0.35\textwidth}
      \centering
      \includegraphics[height=4.5cm]{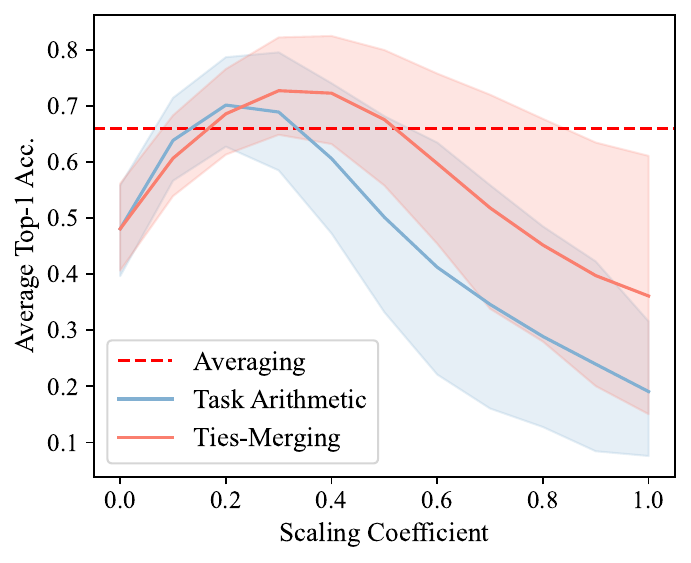}
      \caption{CLIP-ViT-B/32}
      \label{fig:clip-vit-b-32_task-arithmetic-based}
    \end{subfigure}
    \hspace{0.5cm}
    \begin{subfigure}[b]{0.35\textwidth}
      \centering
      \includegraphics[height=4.5cm]{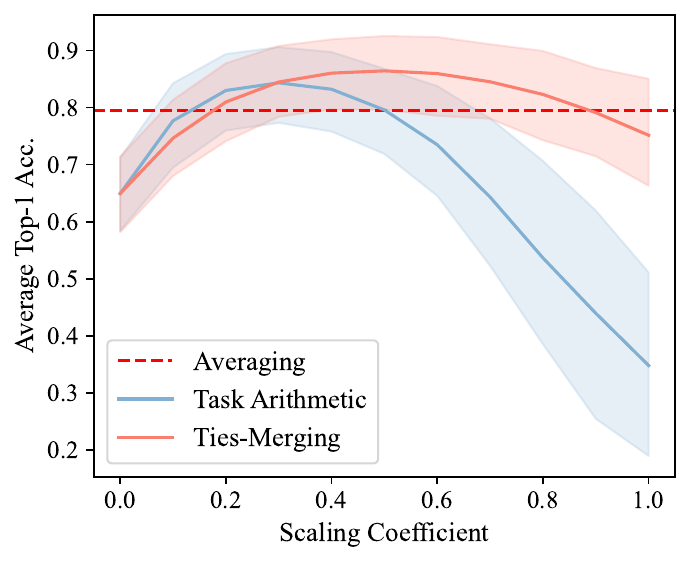}
      \caption{CLIP-ViT-L/14}
      \label{fig:clip-vit-l-14_task-arithmetic-based}
    \end{subfigure}\\
    \begin{subfigure}[b]{0.35\textwidth}
      \centering
      \includegraphics[height=4.5cm]{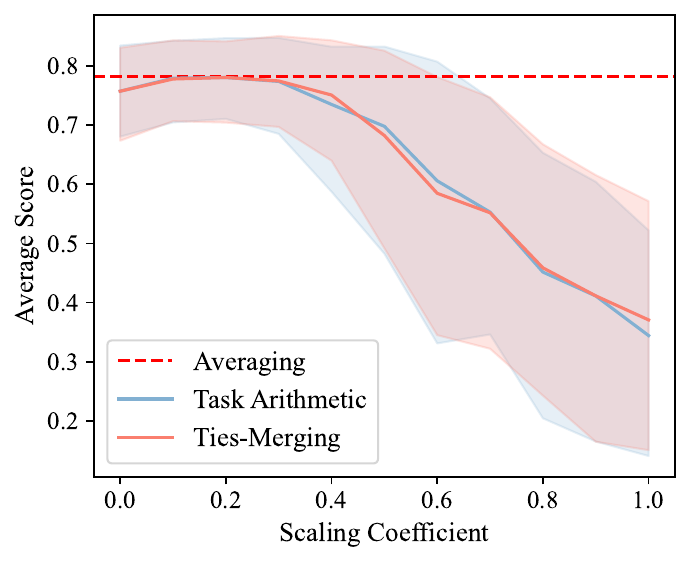}
      \caption{Flan-T5-base (LoRA fine-tuned)}
      \label{fig:flan-t5-base-lora_task-arithmetic-based}
    \end{subfigure}
    \hspace{0.5cm}
    \begin{subfigure}[b]{0.35\textwidth}
      \centering
      \includegraphics[height=4.5cm]{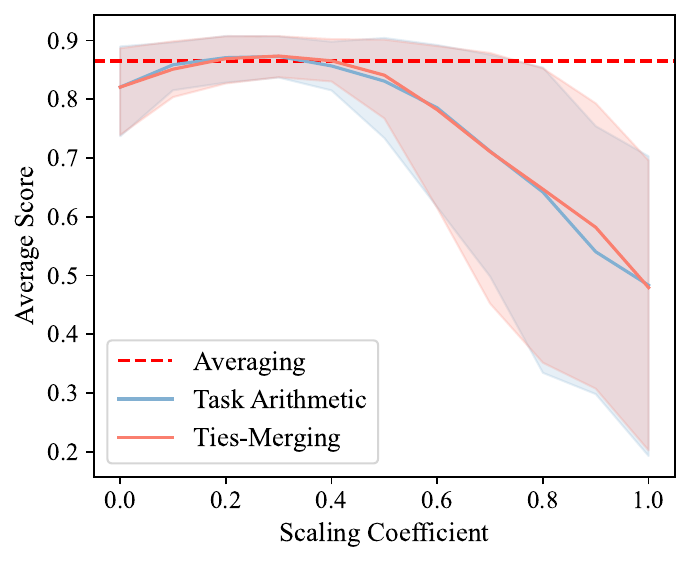}
      \caption{Flan-T5-large (LoRA fine-tuned)}
      \label{fig:flan-t5-large-lora_task-arithmetic-based}
    \end{subfigure}
    \caption{
      \textbf{Task Arithmetic and Ties-Merging.}
      Here we illustrate the average performance of models merged using Task Arithmetic and Ties-Merging methods, with varying scaling coefficients. The subfigures represent different models: CLIP-ViT-B/32, CLIP-ViT-L/14, Flan-T5-base (LoRA fine-tuned), and Flan-T5-large (LoRA fine-tuned). }
    \label{fig:average_score_of_task_arithmetic_and_ties_merging}
  \end{center}
\end{figure}

In Figure~\ref{fig:average_score_of_task_arithmetic_and_ties_merging}, we show the average performance of Task Arithmetic and Ties-Merging merged models as the scaling coefficient varies. Figures~\ref{fig:clip-vit-b-32_task-arithmetic-based}, \ref{fig:clip-vit-l-14_task-arithmetic-based}, \ref{fig:flan-t5-base-lora_task-arithmetic-based}, and \ref{fig:flan-t5-large-lora_task-arithmetic-based} show the results of CLIP-ViT-B/32, CLIP-ViT-L/14, Flan-T5-base (LoRA fine-tuned), and Flan-T5-large (LoRA fine-tuned), respectively.
It is evident that the merged multi-task model hits a peak in average performance across various tasks when the scaling coefficient is set around 0.3.
This value was empirically selected as the scaling coefficient in our experiments.
As we increase the scaling coefficient beyond this point, the average performance of the model begins to decline, eventually even falling below the level of the pre-trained model's original performance.
This suggests that too high of a scaling coefficient can have a negative impact on the knowledge that the pre-trained model initially possessed, emphasizing the importance of calibrating the scaling coefficient parameter $\lambda$ to avoid diminishing the model's existing strengths.

\begin{figure}[!tb]
  \begin{center}
    \begin{subfigure}[b]{0.4\textwidth}
      \centering
      \includegraphics[height=4.5cm]{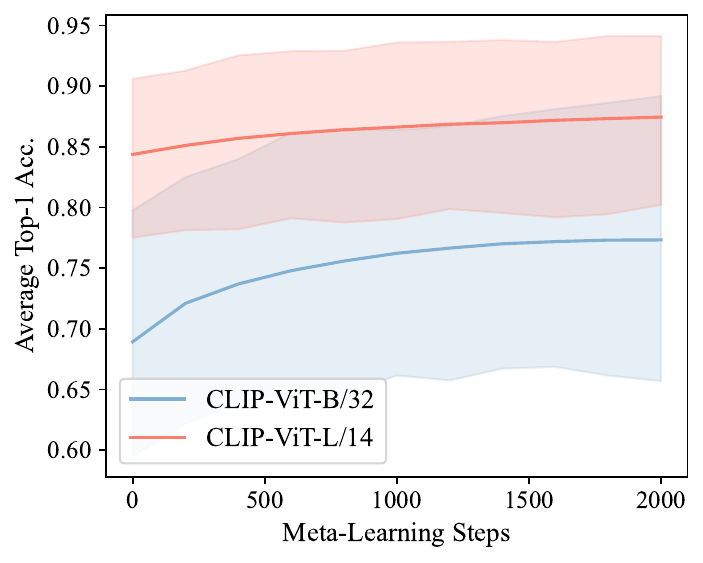}
      \caption{CLIP models}
      \label{fig:clip_concrete_task_arithmetic}
    \end{subfigure}
    \hspace{0.5cm}
    \begin{subfigure}[b]{0.45\textwidth}
      \centering
      \includegraphics[height=4.5cm]{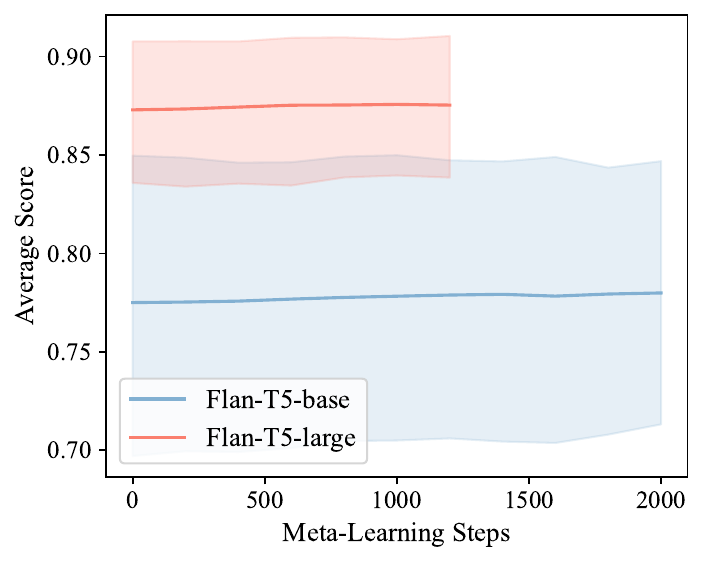}
      \caption{Flan-T5 models}
      \label{fig:flan-t5_concrete_task_arithmetic}
    \end{subfigure}
    \caption{
      \textbf{Concrete Task Arithmetic.}
      The average performance of Concrete Task Arithmetic merged models as the learning step increases.
    }
    \label{fig:average_performance_of_concrete_task_arithmetic}
  \end{center}
\end{figure}

The pseudo-code for combining our Concrete Subspace Learning with task arithmetic is illustrated in Algorithm~\ref{alg:concrete_task_arithmetic}, namely Concrete Task Arithmetic.
This algorithm outlines the steps involved in integrating our Concrete Subspace Learning approach with the task arithmetic method.
Our procedure begins by employing Algorithm~\ref{alg:meta-learn_mask} to determine the Concrete mask.
Throughout our experimental procedures, we have specifically set the scaling coefficient to 0.3 and have conducted the meta-learning process for 2000 steps.
The learning rate is set to 1e-3 for CLIP models and 1e-4 for Flan-T5 models.
Following this, we assess the performance of the model on the test dataset.
The changes in the average model performance during the meta-learning phase of the Concrete task arithmetic are depicted in Figure~\ref{fig:average_performance_of_concrete_task_arithmetic}~\footnote{In the case of the Flan-T5-Large meta-learning, we imposed an early stopping mechanism as the model nearly reached convergence around 1200 steps.}, where we observe a consistent and gradual improvement in model's average performance.

\begin{table}[!tbp]
  \caption{
    \textbf{CLIP-ViT-B/32 with task arithmetic.}
    We show the performance of the merged model with different scaling coefficients on various datasets.
  }
  \label{table:clip-vit-b-32_task_arithmetic}
  \resizebox{\linewidth}{!}{%
    \centering
    \begin{tabular}{c|ccccccccc}
      \toprule
      \textbf{Scaling Coeff.} & \textbf{SUN397}                                                 & \textbf{Cars}                                                   & \textbf{RESISC45}                                               & \textbf{EuroSAT}                                                & \textbf{SVHN}                                                   & \textbf{GTSRB}                                                  & \textbf{MNIST}                                                  & \textbf{DTD}                                                    & \textbf{Avg.}                                                   \\
      \midrule
      0.0                     & \cellcolor[HTML]{ece51b}\textcolor[HTML]{000000}{63.2}          & \cellcolor[HTML]{dde318}\textcolor[HTML]{000000}{59.6}          & \cellcolor[HTML]{86d549}\textcolor[HTML]{000000}{60.2}          & \cellcolor[HTML]{306a8e}\textcolor[HTML]{f1f1f1}{45.0}          & \cellcolor[HTML]{440154}\textcolor[HTML]{f1f1f1}{31.6}          & \cellcolor[HTML]{32648e}\textcolor[HTML]{f1f1f1}{32.6}          & \cellcolor[HTML]{440154}\textcolor[HTML]{f1f1f1}{48.2}          & \cellcolor[HTML]{8bd646}\textcolor[HTML]{000000}{44.4}          & \cellcolor[HTML]{1fa188}\textcolor[HTML]{f1f1f1}{48.1}          \\
      0.1                     & \cellcolor[HTML]{fde725}\textcolor[HTML]{000000}{\textbf{65.0}} & \cellcolor[HTML]{fde725}\textcolor[HTML]{000000}{\textbf{63.0}} & \cellcolor[HTML]{efe51c}\textcolor[HTML]{000000}{70.3}          & \cellcolor[HTML]{a2da37}\textcolor[HTML]{000000}{71.2}          & \cellcolor[HTML]{1fa188}\textcolor[HTML]{f1f1f1}{59.4}          & \cellcolor[HTML]{25ab82}\textcolor[HTML]{f1f1f1}{48.6}          & \cellcolor[HTML]{4ac16d}\textcolor[HTML]{000000}{83.8}          & \cellcolor[HTML]{dde318}\textcolor[HTML]{000000}{49.9}          & \cellcolor[HTML]{addc30}\textcolor[HTML]{000000}{63.9}          \\
      0.2                     & \cellcolor[HTML]{f4e61e}\textcolor[HTML]{000000}{63.8}          & \cellcolor[HTML]{f6e620}\textcolor[HTML]{000000}{62.1}          & \cellcolor[HTML]{fde725}\textcolor[HTML]{000000}{\textbf{72.0}} & \cellcolor[HTML]{fde725}\textcolor[HTML]{000000}{\textbf{78.1}} & \cellcolor[HTML]{aadc32}\textcolor[HTML]{000000}{74.4}          & \cellcolor[HTML]{c8e020}\textcolor[HTML]{000000}{65.1}          & \cellcolor[HTML]{c8e020}\textcolor[HTML]{000000}{94.0}          & \cellcolor[HTML]{fde725}\textcolor[HTML]{000000}{\textbf{52.2}} & \cellcolor[HTML]{fde725}\textcolor[HTML]{000000}{\textbf{70.2}} \\
      0.3                     & \cellcolor[HTML]{9bd93c}\textcolor[HTML]{000000}{55.3}          & \cellcolor[HTML]{a8db34}\textcolor[HTML]{000000}{54.9}          & \cellcolor[HTML]{cae11f}\textcolor[HTML]{000000}{66.7}          & \cellcolor[HTML]{f6e620}\textcolor[HTML]{000000}{77.4}          & \cellcolor[HTML]{f8e621}\textcolor[HTML]{000000}{80.2}          & \cellcolor[HTML]{fde725}\textcolor[HTML]{000000}{\textbf{69.7}} & \cellcolor[HTML]{f4e61e}\textcolor[HTML]{000000}{97.3}          & \cellcolor[HTML]{dfe318}\textcolor[HTML]{000000}{50.1}          & \cellcolor[HTML]{efe51c}\textcolor[HTML]{000000}{69.0}          \\
      0.4                     & \cellcolor[HTML]{1f9f88}\textcolor[HTML]{f1f1f1}{36.7}          & \cellcolor[HTML]{2eb37c}\textcolor[HTML]{f1f1f1}{41.0}          & \cellcolor[HTML]{4ec36b}\textcolor[HTML]{000000}{53.8}          & \cellcolor[HTML]{67cc5c}\textcolor[HTML]{000000}{66.4}          & \cellcolor[HTML]{f8e621}\textcolor[HTML]{000000}{\textbf{80.6}} & \cellcolor[HTML]{d2e21b}\textcolor[HTML]{000000}{66.0}          & \cellcolor[HTML]{fde725}\textcolor[HTML]{000000}{\textbf{98.1}} & \cellcolor[HTML]{70cf57}\textcolor[HTML]{000000}{42.5}          & \cellcolor[HTML]{84d44b}\textcolor[HTML]{000000}{60.7}          \\
      0.5                     & \cellcolor[HTML]{38598c}\textcolor[HTML]{f1f1f1}{18.2}          & \cellcolor[HTML]{2e6f8e}\textcolor[HTML]{f1f1f1}{23.2}          & \cellcolor[HTML]{218e8d}\textcolor[HTML]{f1f1f1}{38.7}          & \cellcolor[HTML]{1f958b}\textcolor[HTML]{f1f1f1}{53.9}          & \cellcolor[HTML]{dae319}\textcolor[HTML]{000000}{77.8}          & \cellcolor[HTML]{6ccd5a}\textcolor[HTML]{000000}{57.4}          & \cellcolor[HTML]{fbe723}\textcolor[HTML]{000000}{97.8}          & \cellcolor[HTML]{22a884}\textcolor[HTML]{f1f1f1}{34.5}          & \cellcolor[HTML]{24aa83}\textcolor[HTML]{f1f1f1}{50.2}          \\
      0.6                     & \cellcolor[HTML]{472a7a}\textcolor[HTML]{f1f1f1}{8.0}           & \cellcolor[HTML]{46327e}\textcolor[HTML]{f1f1f1}{9.5}           & \cellcolor[HTML]{375b8d}\textcolor[HTML]{f1f1f1}{25.1}          & \cellcolor[HTML]{2e6e8e}\textcolor[HTML]{f1f1f1}{45.8}          & \cellcolor[HTML]{98d83e}\textcolor[HTML]{000000}{73.0}          & \cellcolor[HTML]{1f9e89}\textcolor[HTML]{f1f1f1}{45.5}          & \cellcolor[HTML]{e7e419}\textcolor[HTML]{000000}{96.2}          & \cellcolor[HTML]{27808e}\textcolor[HTML]{f1f1f1}{27.0}          & \cellcolor[HTML]{26818e}\textcolor[HTML]{f1f1f1}{41.3}          \\
      0.7                     & \cellcolor[HTML]{471365}\textcolor[HTML]{f1f1f1}{3.5}           & \cellcolor[HTML]{470d60}\textcolor[HTML]{f1f1f1}{2.8}           & \cellcolor[HTML]{453581}\textcolor[HTML]{f1f1f1}{16.7}          & \cellcolor[HTML]{3c4f8a}\textcolor[HTML]{f1f1f1}{39.9}          & \cellcolor[HTML]{50c46a}\textcolor[HTML]{000000}{67.1}          & \cellcolor[HTML]{2d708e}\textcolor[HTML]{f1f1f1}{35.0}          & \cellcolor[HTML]{addc30}\textcolor[HTML]{000000}{91.9}          & \cellcolor[HTML]{375a8c}\textcolor[HTML]{f1f1f1}{20.2}          & \cellcolor[HTML]{34608d}\textcolor[HTML]{f1f1f1}{34.6}          \\
      0.8                     & \cellcolor[HTML]{450559}\textcolor[HTML]{f1f1f1}{1.4}           & \cellcolor[HTML]{450457}\textcolor[HTML]{f1f1f1}{1.2}           & \cellcolor[HTML]{481b6d}\textcolor[HTML]{f1f1f1}{11.4}          & \cellcolor[HTML]{46307e}\textcolor[HTML]{f1f1f1}{34.7}          & \cellcolor[HTML]{20a386}\textcolor[HTML]{f1f1f1}{60.0}          & \cellcolor[HTML]{433e85}\textcolor[HTML]{f1f1f1}{25.2}          & \cellcolor[HTML]{46c06f}\textcolor[HTML]{f1f1f1}{83.3}          & \cellcolor[HTML]{46307e}\textcolor[HTML]{f1f1f1}{14.0}          & \cellcolor[HTML]{424186}\textcolor[HTML]{f1f1f1}{28.9}          \\
      0.9                     & \cellcolor[HTML]{440256}\textcolor[HTML]{f1f1f1}{0.7}           & \cellcolor[HTML]{440154}\textcolor[HTML]{f1f1f1}{0.9}           & \cellcolor[HTML]{460a5d}\textcolor[HTML]{f1f1f1}{8.4}           & \cellcolor[HTML]{481a6c}\textcolor[HTML]{f1f1f1}{31.1}          & \cellcolor[HTML]{27808e}\textcolor[HTML]{f1f1f1}{52.7}          & \cellcolor[HTML]{471164}\textcolor[HTML]{f1f1f1}{17.7}          & \cellcolor[HTML]{277f8e}\textcolor[HTML]{f1f1f1}{69.6}          & \cellcolor[HTML]{481668}\textcolor[HTML]{f1f1f1}{10.4}          & \cellcolor[HTML]{482374}\textcolor[HTML]{f1f1f1}{24.0}          \\
      1.0                     & \cellcolor[HTML]{440154}\textcolor[HTML]{f1f1f1}{0.4}           & \cellcolor[HTML]{440154}\textcolor[HTML]{f1f1f1}{0.7}           & \cellcolor[HTML]{440154}\textcolor[HTML]{f1f1f1}{6.7}           & \cellcolor[HTML]{440154}\textcolor[HTML]{f1f1f1}{27.7}          & \cellcolor[HTML]{39558c}\textcolor[HTML]{f1f1f1}{44.6}          & \cellcolor[HTML]{440154}\textcolor[HTML]{f1f1f1}{15.2}          & \cellcolor[HTML]{460a5d}\textcolor[HTML]{f1f1f1}{49.4}          & \cellcolor[HTML]{440154}\textcolor[HTML]{f1f1f1}{7.8}           & \cellcolor[HTML]{440154}\textcolor[HTML]{f1f1f1}{19.1}          \\
      \bottomrule
    \end{tabular}
  }
\end{table}

\begin{table}[!tbp]
  \caption{
    \textbf{CLIP-ViT-B/32 with Ties-Merging.}
    We show the performance of the merged model with different scaling coefficients on various datasets.
  }
  \label{table:clip-vit-b-32_ties_merging}
  \resizebox{\linewidth}{!}{%
    \centering
    \begin{tabular}{c|ccccccccc}
      \toprule
      \textbf{Scaling Coeff.} & \textbf{SUN397}                                                 & \textbf{Cars}                                                   & \textbf{RESISC45}                                               & \textbf{EuroSAT}                                                & \textbf{SVHN}                                                   & \textbf{GTSRB}                                                  & \textbf{MNIST}                                                  & \textbf{DTD}                                                    & \textbf{Avg.}                                                   \\
      \midrule
      0.0                     & \cellcolor[HTML]{dde318}\textcolor[HTML]{000000}{63.2}          & \cellcolor[HTML]{c5e021}\textcolor[HTML]{000000}{59.6}          & \cellcolor[HTML]{60ca60}\textcolor[HTML]{000000}{60.2}          & \cellcolor[HTML]{355f8d}\textcolor[HTML]{f1f1f1}{45.0}          & \cellcolor[HTML]{440154}\textcolor[HTML]{f1f1f1}{31.6}          & \cellcolor[HTML]{440154}\textcolor[HTML]{f1f1f1}{32.6}          & \cellcolor[HTML]{440154}\textcolor[HTML]{f1f1f1}{48.2}          & \cellcolor[HTML]{4ac16d}\textcolor[HTML]{000000}{44.4}          & \cellcolor[HTML]{31668e}\textcolor[HTML]{f1f1f1}{48.1}          \\
      0.1                     & \cellcolor[HTML]{f4e61e}\textcolor[HTML]{000000}{65.3}          & \cellcolor[HTML]{e7e419}\textcolor[HTML]{000000}{62.6}          & \cellcolor[HTML]{bddf26}\textcolor[HTML]{000000}{68.7}          & \cellcolor[HTML]{46c06f}\textcolor[HTML]{f1f1f1}{63.9}          & \cellcolor[HTML]{2e6d8e}\textcolor[HTML]{f1f1f1}{51.9}          & \cellcolor[HTML]{34618d}\textcolor[HTML]{f1f1f1}{45.0}          & \cellcolor[HTML]{25ab82}\textcolor[HTML]{f1f1f1}{79.2}          & \cellcolor[HTML]{98d83e}\textcolor[HTML]{000000}{49.0}          & \cellcolor[HTML]{37b878}\textcolor[HTML]{f1f1f1}{60.7}          \\
      0.2                     & \cellcolor[HTML]{fde725}\textcolor[HTML]{000000}{\textbf{66.5}} & \cellcolor[HTML]{fde725}\textcolor[HTML]{000000}{\textbf{64.9}} & \cellcolor[HTML]{ece51b}\textcolor[HTML]{000000}{73.0}          & \cellcolor[HTML]{b0dd2f}\textcolor[HTML]{000000}{72.1}          & \cellcolor[HTML]{37b878}\textcolor[HTML]{f1f1f1}{70.0}          & \cellcolor[HTML]{2fb47c}\textcolor[HTML]{f1f1f1}{58.8}          & \cellcolor[HTML]{9dd93b}\textcolor[HTML]{000000}{91.6}          & \cellcolor[HTML]{dae319}\textcolor[HTML]{000000}{52.3}          & \cellcolor[HTML]{b2dd2d}\textcolor[HTML]{000000}{68.6}          \\
      0.3                     & \cellcolor[HTML]{f1e51d}\textcolor[HTML]{000000}{65.0}          & \cellcolor[HTML]{f8e621}\textcolor[HTML]{000000}{64.3}          & \cellcolor[HTML]{fde725}\textcolor[HTML]{000000}{\textbf{74.7}} & \cellcolor[HTML]{f4e61e}\textcolor[HTML]{000000}{76.8}          & \cellcolor[HTML]{a8db34}\textcolor[HTML]{000000}{81.3}          & \cellcolor[HTML]{c5e021}\textcolor[HTML]{000000}{69.4}          & \cellcolor[HTML]{dfe318}\textcolor[HTML]{000000}{96.5}          & \cellcolor[HTML]{fde725}\textcolor[HTML]{000000}{\textbf{54.3}} & \cellcolor[HTML]{fde725}\textcolor[HTML]{000000}{\textbf{72.8}} \\
      0.4                     & \cellcolor[HTML]{b5de2b}\textcolor[HTML]{000000}{59.2}          & \cellcolor[HTML]{c8e020}\textcolor[HTML]{000000}{59.8}          & \cellcolor[HTML]{dfe318}\textcolor[HTML]{000000}{71.7}          & \cellcolor[HTML]{fde725}\textcolor[HTML]{000000}{\textbf{77.6}} & \cellcolor[HTML]{e2e418}\textcolor[HTML]{000000}{86.3}          & \cellcolor[HTML]{fde725}\textcolor[HTML]{000000}{\textbf{72.9}} & \cellcolor[HTML]{f4e61e}\textcolor[HTML]{000000}{98.2}          & \cellcolor[HTML]{e2e418}\textcolor[HTML]{000000}{52.8}          & \cellcolor[HTML]{f6e620}\textcolor[HTML]{000000}{72.3}          \\
      0.5                     & \cellcolor[HTML]{40bd72}\textcolor[HTML]{f1f1f1}{46.5}          & \cellcolor[HTML]{69cd5b}\textcolor[HTML]{000000}{51.1}          & \cellcolor[HTML]{8ed645}\textcolor[HTML]{000000}{64.5}          & \cellcolor[HTML]{b2dd2d}\textcolor[HTML]{000000}{72.3}          & \cellcolor[HTML]{f8e621}\textcolor[HTML]{000000}{88.3}          & \cellcolor[HTML]{eae51a}\textcolor[HTML]{000000}{71.6}          & \cellcolor[HTML]{fbe723}\textcolor[HTML]{000000}{98.8}          & \cellcolor[HTML]{81d34d}\textcolor[HTML]{000000}{47.8}          & \cellcolor[HTML]{a0da39}\textcolor[HTML]{000000}{67.6}          \\
      0.6                     & \cellcolor[HTML]{277f8e}\textcolor[HTML]{f1f1f1}{29.3}          & \cellcolor[HTML]{21a585}\textcolor[HTML]{f1f1f1}{39.9}          & \cellcolor[HTML]{2ab07f}\textcolor[HTML]{f1f1f1}{52.9}          & \cellcolor[HTML]{27ad81}\textcolor[HTML]{f1f1f1}{60.0}          & \cellcolor[HTML]{fde725}\textcolor[HTML]{000000}{\textbf{88.8}} & \cellcolor[HTML]{a2da37}\textcolor[HTML]{000000}{67.3}          & \cellcolor[HTML]{fde725}\textcolor[HTML]{000000}{\textbf{99.0}} & \cellcolor[HTML]{25ab82}\textcolor[HTML]{f1f1f1}{41.0}          & \cellcolor[HTML]{2eb37c}\textcolor[HTML]{f1f1f1}{59.8}          \\
      0.7                     & \cellcolor[HTML]{3f4788}\textcolor[HTML]{f1f1f1}{15.3}          & \cellcolor[HTML]{2a778e}\textcolor[HTML]{f1f1f1}{28.4}          & \cellcolor[HTML]{29798e}\textcolor[HTML]{f1f1f1}{39.2}          & \cellcolor[HTML]{2b758e}\textcolor[HTML]{f1f1f1}{49.2}          & \cellcolor[HTML]{f6e620}\textcolor[HTML]{000000}{88.0}          & \cellcolor[HTML]{4cc26c}\textcolor[HTML]{000000}{61.5}          & \cellcolor[HTML]{fde725}\textcolor[HTML]{000000}{98.9}          & \cellcolor[HTML]{287c8e}\textcolor[HTML]{f1f1f1}{34.3}          & \cellcolor[HTML]{277f8e}\textcolor[HTML]{f1f1f1}{51.8}          \\
      0.8                     & \cellcolor[HTML]{481f70}\textcolor[HTML]{f1f1f1}{7.1}           & \cellcolor[HTML]{3f4889}\textcolor[HTML]{f1f1f1}{17.7}          & \cellcolor[HTML]{404688}\textcolor[HTML]{f1f1f1}{27.5}          & \cellcolor[HTML]{3c508b}\textcolor[HTML]{f1f1f1}{42.4}          & \cellcolor[HTML]{e5e419}\textcolor[HTML]{000000}{86.4}          & \cellcolor[HTML]{20928c}\textcolor[HTML]{f1f1f1}{53.0}          & \cellcolor[HTML]{fbe723}\textcolor[HTML]{000000}{98.7}          & \cellcolor[HTML]{3a538b}\textcolor[HTML]{f1f1f1}{28.8}          & \cellcolor[HTML]{3b518b}\textcolor[HTML]{f1f1f1}{45.2}          \\
      0.9                     & \cellcolor[HTML]{460a5d}\textcolor[HTML]{f1f1f1}{3.3}           & \cellcolor[HTML]{481d6f}\textcolor[HTML]{f1f1f1}{9.4}           & \cellcolor[HTML]{481f70}\textcolor[HTML]{f1f1f1}{20.0}          & \cellcolor[HTML]{482173}\textcolor[HTML]{f1f1f1}{35.3}          & \cellcolor[HTML]{cae11f}\textcolor[HTML]{000000}{84.2}          & \cellcolor[HTML]{365d8d}\textcolor[HTML]{f1f1f1}{44.4}          & \cellcolor[HTML]{f6e620}\textcolor[HTML]{000000}{98.4}          & \cellcolor[HTML]{482173}\textcolor[HTML]{f1f1f1}{23.2}          & \cellcolor[HTML]{482475}\textcolor[HTML]{f1f1f1}{39.8}          \\
      1.0                     & \cellcolor[HTML]{440154}\textcolor[HTML]{f1f1f1}{1.6}           & \cellcolor[HTML]{440154}\textcolor[HTML]{f1f1f1}{4.5}           & \cellcolor[HTML]{440154}\textcolor[HTML]{f1f1f1}{15.0}          & \cellcolor[HTML]{440154}\textcolor[HTML]{f1f1f1}{31.0}          & \cellcolor[HTML]{a8db34}\textcolor[HTML]{000000}{81.3}          & \cellcolor[HTML]{472d7b}\textcolor[HTML]{f1f1f1}{37.7}          & \cellcolor[HTML]{f1e51d}\textcolor[HTML]{000000}{97.9}          & \cellcolor[HTML]{440154}\textcolor[HTML]{f1f1f1}{20.0}          & \cellcolor[HTML]{440154}\textcolor[HTML]{f1f1f1}{36.1}          \\
      \bottomrule
    \end{tabular}
  }
\end{table}

\begin{table}[!tbp]
  \caption{
    \textbf{CLIP-ViT-B/32 with Concrete Task Arithmetic.}
    We show the performance of the merged model with increasing test-time adaptation steps on various datasets, see Algorithm~\ref{alg:meta-learn_mask} and \ref{alg:concrete_task_arithmetic}.
    The performance of the merged model improves consistently as the number of test-time adaptation steps increases, which demonstrates the effectiveness of our Concrete Task Arithmetic to resolve task interference.
  }
  \label{table:clip-vit-b-32_concrete_task_arithmetic}
  \resizebox{\linewidth}{!}{%
    \centering
    \begin{tabular}{c|ccccccccc}
      \toprule
      \textbf{Step} & \textbf{SUN397}                                        & \textbf{Cars}                                          & \textbf{RESISC45}                                      & \textbf{EuroSAT}                                       & \textbf{SVHN}                                          & \textbf{GTSRB}                                         & \textbf{MNIST}                                         & \textbf{DTD}                                           & \textbf{Avg.}                                          \\\midrule
      0             & \cellcolor[HTML]{440154}\textcolor[HTML]{f1f1f1}{55.2} & \cellcolor[HTML]{440154}\textcolor[HTML]{f1f1f1}{54.7} & \cellcolor[HTML]{440154}\textcolor[HTML]{f1f1f1}{66.7} & \cellcolor[HTML]{440154}\textcolor[HTML]{f1f1f1}{77.8} & \cellcolor[HTML]{440154}\textcolor[HTML]{f1f1f1}{80.4} & \cellcolor[HTML]{440154}\textcolor[HTML]{f1f1f1}{69.4} & \cellcolor[HTML]{440154}\textcolor[HTML]{f1f1f1}{97.3} & \cellcolor[HTML]{440154}\textcolor[HTML]{f1f1f1}{49.9} & \cellcolor[HTML]{440154}\textcolor[HTML]{f1f1f1}{68.9} \\
      200           & \cellcolor[HTML]{287d8e}\textcolor[HTML]{f1f1f1}{58.3} & \cellcolor[HTML]{21a585}\textcolor[HTML]{f1f1f1}{59.3} & \cellcolor[HTML]{25848e}\textcolor[HTML]{f1f1f1}{70.9} & \cellcolor[HTML]{297b8e}\textcolor[HTML]{f1f1f1}{85.2} & \cellcolor[HTML]{424086}\textcolor[HTML]{f1f1f1}{82.4} & \cellcolor[HTML]{404688}\textcolor[HTML]{f1f1f1}{72.0} & \cellcolor[HTML]{31688e}\textcolor[HTML]{f1f1f1}{97.7} & \cellcolor[HTML]{297a8e}\textcolor[HTML]{f1f1f1}{51.0} & \cellcolor[HTML]{2c738e}\textcolor[HTML]{f1f1f1}{72.1} \\
      400           & \cellcolor[HTML]{27ad81}\textcolor[HTML]{f1f1f1}{59.8} & \cellcolor[HTML]{7fd34e}\textcolor[HTML]{000000}{61.0} & \cellcolor[HTML]{2eb37c}\textcolor[HTML]{f1f1f1}{72.7} & \cellcolor[HTML]{26ad81}\textcolor[HTML]{f1f1f1}{88.9} & \cellcolor[HTML]{2c728e}\textcolor[HTML]{f1f1f1}{84.4} & \cellcolor[HTML]{34608d}\textcolor[HTML]{f1f1f1}{73.2} & \cellcolor[HTML]{20a486}\textcolor[HTML]{f1f1f1}{98.0} & \cellcolor[HTML]{35b779}\textcolor[HTML]{f1f1f1}{51.7} & \cellcolor[HTML]{1fa187}\textcolor[HTML]{f1f1f1}{73.7} \\
      600           & \cellcolor[HTML]{3fbc73}\textcolor[HTML]{f1f1f1}{60.3} & \cellcolor[HTML]{a0da39}\textcolor[HTML]{000000}{61.4} & \cellcolor[HTML]{6ccd5a}\textcolor[HTML]{000000}{73.9} & \cellcolor[HTML]{63cb5f}\textcolor[HTML]{000000}{91.4} & \cellcolor[HTML]{20928c}\textcolor[HTML]{f1f1f1}{85.8} & \cellcolor[HTML]{23898e}\textcolor[HTML]{f1f1f1}{75.3} & \cellcolor[HTML]{35b779}\textcolor[HTML]{f1f1f1}{98.1} & \cellcolor[HTML]{58c765}\textcolor[HTML]{000000}{51.9} & \cellcolor[HTML]{44bf70}\textcolor[HTML]{f1f1f1}{74.8} \\
      800           & \cellcolor[HTML]{81d34d}\textcolor[HTML]{000000}{61.2} & \cellcolor[HTML]{fde725}\textcolor[HTML]{000000}{62.5} & \cellcolor[HTML]{86d549}\textcolor[HTML]{000000}{74.3} & \cellcolor[HTML]{95d840}\textcolor[HTML]{000000}{92.9} & \cellcolor[HTML]{27ad81}\textcolor[HTML]{f1f1f1}{87.0} & \cellcolor[HTML]{21a685}\textcolor[HTML]{f1f1f1}{76.8} & \cellcolor[HTML]{5ec962}\textcolor[HTML]{000000}{98.2} & \cellcolor[HTML]{46c06f}\textcolor[HTML]{f1f1f1}{51.8} & \cellcolor[HTML]{7ad151}\textcolor[HTML]{000000}{75.6} \\
      1000          & \cellcolor[HTML]{a5db36}\textcolor[HTML]{000000}{61.6} & \cellcolor[HTML]{e7e419}\textcolor[HTML]{000000}{62.2} & \cellcolor[HTML]{b0dd2f}\textcolor[HTML]{000000}{74.9} & \cellcolor[HTML]{c5e021}\textcolor[HTML]{000000}{94.1} & \cellcolor[HTML]{48c16e}\textcolor[HTML]{f1f1f1}{87.9} & \cellcolor[HTML]{52c569}\textcolor[HTML]{000000}{78.5} & \cellcolor[HTML]{5ec962}\textcolor[HTML]{000000}{98.2} & \cellcolor[HTML]{b5de2b}\textcolor[HTML]{000000}{52.3} & \cellcolor[HTML]{a8db34}\textcolor[HTML]{000000}{76.2} \\
      1200          & \cellcolor[HTML]{b8de29}\textcolor[HTML]{000000}{61.8} & \cellcolor[HTML]{e7e419}\textcolor[HTML]{000000}{62.2} & \cellcolor[HTML]{c5e021}\textcolor[HTML]{000000}{75.2} & \cellcolor[HTML]{dae319}\textcolor[HTML]{000000}{94.7} & \cellcolor[HTML]{81d34d}\textcolor[HTML]{000000}{89.0} & \cellcolor[HTML]{8ed645}\textcolor[HTML]{000000}{79.8} & \cellcolor[HTML]{c8e020}\textcolor[HTML]{000000}{98.4} & \cellcolor[HTML]{84d44b}\textcolor[HTML]{000000}{52.1} & \cellcolor[HTML]{c8e020}\textcolor[HTML]{000000}{76.6} \\
      1400          & \cellcolor[HTML]{dde318}\textcolor[HTML]{000000}{62.2} & \cellcolor[HTML]{dde318}\textcolor[HTML]{000000}{62.1} & \cellcolor[HTML]{cde11d}\textcolor[HTML]{000000}{75.3} & \cellcolor[HTML]{e5e419}\textcolor[HTML]{000000}{95.0} & \cellcolor[HTML]{a0da39}\textcolor[HTML]{000000}{89.5} & \cellcolor[HTML]{d5e21a}\textcolor[HTML]{000000}{81.1} & \cellcolor[HTML]{c8e020}\textcolor[HTML]{000000}{98.4} & \cellcolor[HTML]{58c765}\textcolor[HTML]{000000}{51.9} & \cellcolor[HTML]{e7e419}\textcolor[HTML]{000000}{77.0} \\
      1600          & \cellcolor[HTML]{efe51c}\textcolor[HTML]{000000}{62.4} & \cellcolor[HTML]{b2dd2d}\textcolor[HTML]{000000}{61.6} & \cellcolor[HTML]{eae51a}\textcolor[HTML]{000000}{75.7} & \cellcolor[HTML]{f4e61e}\textcolor[HTML]{000000}{95.4} & \cellcolor[HTML]{cde11d}\textcolor[HTML]{000000}{90.2} & \cellcolor[HTML]{e5e419}\textcolor[HTML]{000000}{81.4} & \cellcolor[HTML]{fde725}\textcolor[HTML]{000000}{98.5} & \cellcolor[HTML]{9dd93b}\textcolor[HTML]{000000}{52.2} & \cellcolor[HTML]{f6e620}\textcolor[HTML]{000000}{77.2} \\
      1800          & \cellcolor[HTML]{fde725}\textcolor[HTML]{000000}{62.6} & \cellcolor[HTML]{c5e021}\textcolor[HTML]{000000}{61.8} & \cellcolor[HTML]{fde725}\textcolor[HTML]{000000}{76.0} & \cellcolor[HTML]{f4e61e}\textcolor[HTML]{000000}{95.4} & \cellcolor[HTML]{e7e419}\textcolor[HTML]{000000}{90.6} & \cellcolor[HTML]{f4e61e}\textcolor[HTML]{000000}{81.7} & \cellcolor[HTML]{fde725}\textcolor[HTML]{000000}{98.5} & \cellcolor[HTML]{58c765}\textcolor[HTML]{000000}{51.9} & \cellcolor[HTML]{fde725}\textcolor[HTML]{000000}{77.3} \\
      2000          & \cellcolor[HTML]{f6e620}\textcolor[HTML]{000000}{62.5} & \cellcolor[HTML]{89d548}\textcolor[HTML]{000000}{61.1} & \cellcolor[HTML]{fde725}\textcolor[HTML]{000000}{76.0} & \cellcolor[HTML]{fde725}\textcolor[HTML]{000000}{95.7} & \cellcolor[HTML]{fde725}\textcolor[HTML]{000000}{91.0} & \cellcolor[HTML]{fde725}\textcolor[HTML]{000000}{81.9} & \cellcolor[HTML]{fde725}\textcolor[HTML]{000000}{98.5} & \cellcolor[HTML]{58c765}\textcolor[HTML]{000000}{51.9} & \cellcolor[HTML]{fde725}\textcolor[HTML]{000000}{77.3} \\
      \bottomrule
    \end{tabular}
  }
\end{table}

Tables~\ref{table:clip-vit-b-32_task_arithmetic}, \ref{table:clip-vit-b-32_ties_merging} and \ref{table:clip-vit-b-32_concrete_task_arithmetic} show the performance of CLIP-ViT-B/32 with task arithmetic, Ties-Merging, and Concrete Task Arithmetic, respectively.
Each table shows how performance varies across a set of benchmark datasets with different scaling coefficients or adaptation steps.

For both Task Arithmetic and Ties-Merging, we observed an initial increase in performance across most tasks as the scaling coefficient increased. The peak performance was typically achieved around a scaling coefficient of 0.3, after which the performance began to decline rapidly on most tasks.
This decline can be attributed to the competition and interference among various tasks.
As the scaling coefficient increases, the model tends to overfit certain tasks, leading to a decrease in performance on other tasks. This phenomenon was particularly noticeable on the SVHN and MNIST datasets.

In contrast, for Concrete Task Arithmetic, we noticed a gradual improvement in model performance with an increase in the number of adaptation steps.
These results indicate that performing model merging under the common subspace learned through meta-learning can help resolve interference among downstream tasks.


\subsection{AdaMerging-Based Methods}
\label{appendix:adamerging_based_methods}

AdaMerging-based methods form a distinct category within the landscape of model merging strategies, where the primary focus is on refinement.
Unlike task arithmetic-based methods that rely on a uniform scaling coefficient, AdaMerging-based strategies introduce a fine-grained approach by assigning unique weights to each task or different layers within the task vector.
This nuanced weighting mechanism allows for a more personalized integration of task-specific knowledge into the pre-trained model.
We describe this process mathematically as:
\begin{equation}
  \label{eq:adamerging}
  \theta = \theta_0 + \sum_{i=1}^{n} \lambda_i \mathcal{A}(\tau_i) \quad \text{or}\quad \theta^l = \theta_0^l + \sum_{i=1}^{n} \lambda^{l}_{i} \mathcal{A}(\tau^{l}_{i}).
\end{equation}
Here, $\mathcal{A}$ represents a transformation applied to the task vector $\tau_i$,
such as masking and rescaling in our Concrete AdaMerging (Algorithm~\ref{alg:concrete_adamerging}), as well as the elimination of redundant parameter values and resolution of sign conflicts in AdaMerging++~\cite{yangAdaMergingAdaptiveModel2024}.

Fundamentally, AdaMerging recognizes the diversity of tasks and their varying degrees of relevance to the underlying model.
This can be particularly advantageous when dealing with tasks of varying complexity or dissimilar data distributions.
For layers, the approach allows the model to capitalize on the hierarchical representation of features at different levels of abstraction.
In practice, this translates into a meticulous process of weight allocation that considers the specific contribution of each task, or the particular influence of each layer when merging with a pre-trained model's existing knowledge base by performing test-time adaptation on unlabeled test samples.

The versatility of AdaMerging-based methods lends itself well to complex multi-task learning scenarios.
It has the potential to finely balance task-specific learning without risking the dilution of the model’s core competencies.
Moreover, by treating different tasks and layers with individual consideration, AdaMerging-based methods can effectively restrain the negative interference often witnessed when a model is adjusted to new tasks that are substantially different from the ones it was originally trained on.

In the upcoming paragraphs, we will explore the experimental results obtained from the utilization of AdaMerging and our Concrete AdaMerging.
This exploration aims to offer a detailed understanding of their practical implications and the extent of their applicability.

\begin{figure}[!tb]
  \begin{center}
    \begin{subfigure}[b]{0.4\textwidth}
      \centering
      \includegraphics[height=4.5cm]{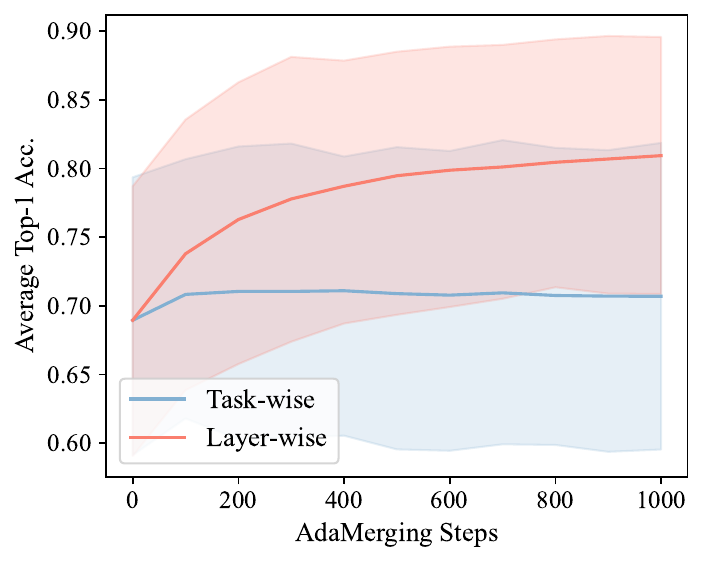}
      \caption{CLIP-ViT-B/32}
      \label{fig:vit-b-32_adamerging}
    \end{subfigure}
    \hspace{0.5cm}
    \begin{subfigure}[b]{0.45\textwidth}
      \centering
      \includegraphics[height=4.5cm]{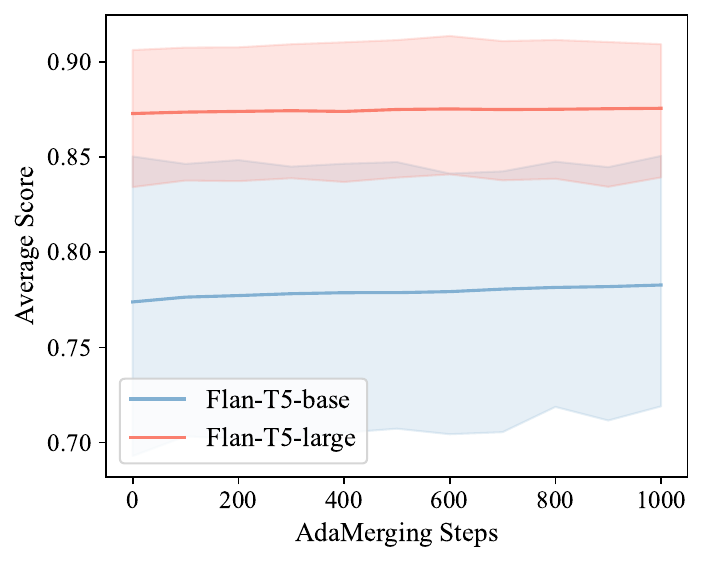}
      \caption{Flan-T5 models with layer-wise AdaMerging}
      \label{fig:flan-t5_layer_wise_adamerging}
    \end{subfigure}
    \caption{
      \textbf{AdaMerging.}
      The average performance of AdaMerging merged models as the learning step increases.
    }
    \label{fig:average_performance_of_adamerging}
  \end{center}
\end{figure}

Figure~\ref{fig:average_performance_of_adamerging} shows the average performance of AdaMerging merged models as the learning step increases.
Specifically, Figure~\ref{fig:vit-b-32_adamerging} shows the results of CLIP-ViT-B/32 with both task-wise AdaMerging and layer-wise AdaMerging, and Figure~\ref{fig:flan-t5_layer_wise_adamerging} shows the results of Flan-T5-base and Flan-T5-large models with layer-wise AdaMerging.

We observe that the average performance of the model increases as the learning step increases.
This suggests that the AdaMerging merged model can benefit from the test-time adaptation process and improve its performance on the test dataset.
In the case of CLIP-ViT-B/32, the average performance of the model with layer-wise AdaMerging is higher than the model with task-wise AdaMerging.
This is because the layer-wise AdaMerging allows the model to adapt to different tasks at different layers, which is more flexible than the task-wise AdaMerging.
In the case of Flan-T5-base and Flan-T5-large, the average performance of Flan-T5-large is higher than Flan-T5-base, which indicates that the model with more parameters can benefit more from the AdaMerging.

As for our Concrete AdaMerging, as explained in Algorithm~\ref{alg:concrete_adamerging}, we first use Algorithm~\ref{alg:meta-learn_mask} to determine the Concrete mask.
Then, we use the Concrete mask to mask the task vector and rescale it to obtain the final task vector to perform normal AdaMerging. We conduct the meta-learning process for 2000 steps and the AdaMerging process for 1000 steps.
We use Adam optimizer to optimize the parameters.
The learning rate settings are $\alpha = 1, \beta = 0.001$ for CLIP models and $\alpha = 0.1, \beta = 0.0001$ for Flan-T5 models.
During the test-time adaptation phase, we set the learning rate to $\beta$.

\begin{figure}[!tb]
  \begin{center}
    \begin{subfigure}[b]{0.4\textwidth}
      \centering
      \includegraphics[height=4.5cm]{images/clip-vit-b-32_concrete_adamerging.pdf}
      \caption{Meta-learn the Concrete mask}
      \label{fig:clip-vit-b-32_concrete_adamerging}
    \end{subfigure}
    \hspace{0.5cm}
    \begin{subfigure}[b]{0.4\textwidth}
      \centering
      \includegraphics[height=4.5cm]{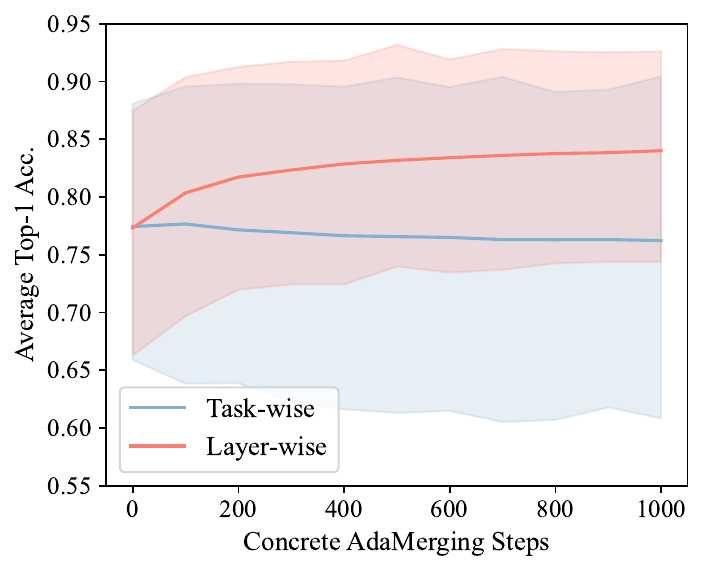}
      \caption{AdaMerging with the Concrete mask}
      \label{fig:clip-vit-b-32_concrete_adamerging_tta}
    \end{subfigure}
    \caption{
      \textbf{CLIP-ViT-B/32 with Concrete AdaMerging.}
      Here we show the whole process of applying Concrete AdaMerging to CLIP-ViT-B/32, the y-axis is shared by these two subfigures:
      (a) shows the performance of the merged model during the meta-learning phase of the Concrete AdaMerging, see Algorithm~\ref{alg:meta-learn_mask};
      (b) shows the performance of the model during the test-time adaptation phase of the Concrete AdaMerging, see Algorithm~\ref{alg:concrete_adamerging}.
    }
    \label{fig:vit-b-32_concrete_adamerging}
  \end{center}
\end{figure}

Figure~\ref{fig:vit-b-32_concrete_adamerging} shows the whole process of applying Concrete AdaMerging to CLIP-ViT-B/32.
Specifically, Figure~\ref{fig:clip-vit-b-32_concrete_adamerging} shows the changes in the average model performance during the meta-learning phase of the Concrete AdaMerging. We observe a consistent and gradual improvement in the model's average performance.
Figure~\ref{fig:clip-vit-b-32_concrete_adamerging_tta} shows the changes in the average model performance during the test-time adaptation phase of the Concrete AdaMerging. We observe that the average performance of the model increases as the learning step increases, which indicates that Concrete AdaMerging can benefit from the test-time adaptation process and improve its performance on the test dataset.

\begin{table}[!tbp]
  \caption{
    \textbf{CLIP-ViT-B/32 with Layer-wise AdaMerging.}
    We show the performance of the merged model with increasing test-time adaptation steps on various datasets.
  }
  \label{table:clip-vit-b-32_layer_wise_adamerging}
  \resizebox{\linewidth}{!}{%
    \centering
    \begin{tabular}{c|ccccccccc}
      \toprule
      \textbf{TTA Step} & \textbf{SUN397}                                                 & \textbf{Cars}                                                   & \textbf{RESISC45}                                               & \textbf{EuroSAT}                                                & \textbf{SVHN}                                                   & \textbf{GTSRB}                                                  & \textbf{MNIST}                                                  & \textbf{DTD}                                                    & \textbf{Avg.}                                                   \\
      \midrule
      0                 & \cellcolor[HTML]{440154}\textcolor[HTML]{f1f1f1}{55.3}          & \cellcolor[HTML]{440154}\textcolor[HTML]{f1f1f1}{54.9}          & \cellcolor[HTML]{440154}\textcolor[HTML]{f1f1f1}{66.7}          & \cellcolor[HTML]{440154}\textcolor[HTML]{f1f1f1}{77.4}          & \cellcolor[HTML]{440154}\textcolor[HTML]{f1f1f1}{80.2}          & \cellcolor[HTML]{440154}\textcolor[HTML]{f1f1f1}{69.7}          & \cellcolor[HTML]{2a788e}\textcolor[HTML]{f1f1f1}{97.3}          & \cellcolor[HTML]{440154}\textcolor[HTML]{f1f1f1}{50.1}          & \cellcolor[HTML]{440154}\textcolor[HTML]{f1f1f1}{69}            \\
      100               & \cellcolor[HTML]{218f8d}\textcolor[HTML]{f1f1f1}{59.7}          & \cellcolor[HTML]{2c728e}\textcolor[HTML]{f1f1f1}{60.4}          & \cellcolor[HTML]{2f6c8e}\textcolor[HTML]{f1f1f1}{72.2}          & \cellcolor[HTML]{4ac16d}\textcolor[HTML]{000000}{88.3}          & \cellcolor[HTML]{32b67a}\textcolor[HTML]{f1f1f1}{84.4}          & \cellcolor[HTML]{365d8d}\textcolor[HTML]{f1f1f1}{76.7}          & \cellcolor[HTML]{fde725}\textcolor[HTML]{000000}{97.6}          & \cellcolor[HTML]{481d6f}\textcolor[HTML]{f1f1f1}{51.0}          & \cellcolor[HTML]{29798e}\textcolor[HTML]{f1f1f1}{73.8}          \\
      200               & \cellcolor[HTML]{3dbc74}\textcolor[HTML]{f1f1f1}{61.4}          & \cellcolor[HTML]{22a785}\textcolor[HTML]{f1f1f1}{63.6}          & \cellcolor[HTML]{218e8d}\textcolor[HTML]{f1f1f1}{74.4}          & \cellcolor[HTML]{e2e418}\textcolor[HTML]{000000}{92.0}          & \cellcolor[HTML]{95d840}\textcolor[HTML]{000000}{85.6}          & \cellcolor[HTML]{1fa287}\textcolor[HTML]{f1f1f1}{83.5}          & \cellcolor[HTML]{7ad151}\textcolor[HTML]{000000}{97.5}          & \cellcolor[HTML]{404688}\textcolor[HTML]{f1f1f1}{52.4}          & \cellcolor[HTML]{25ac82}\textcolor[HTML]{f1f1f1}{76.3}          \\
      300               & \cellcolor[HTML]{73d056}\textcolor[HTML]{000000}{62.3}          & \cellcolor[HTML]{4ec36b}\textcolor[HTML]{000000}{65.4}          & \cellcolor[HTML]{21a585}\textcolor[HTML]{f1f1f1}{75.9}          & \cellcolor[HTML]{f1e51d}\textcolor[HTML]{000000}{92.4}          & \cellcolor[HTML]{f6e620}\textcolor[HTML]{000000}{86.5}          & \cellcolor[HTML]{63cb5f}\textcolor[HTML]{000000}{87.9}          & \cellcolor[HTML]{7ad151}\textcolor[HTML]{000000}{97.5}          & \cellcolor[HTML]{2a768e}\textcolor[HTML]{f1f1f1}{54.4}          & \cellcolor[HTML]{58c765}\textcolor[HTML]{000000}{77.8}          \\
      400               & \cellcolor[HTML]{addc30}\textcolor[HTML]{000000}{63.1}          & \cellcolor[HTML]{7fd34e}\textcolor[HTML]{000000}{66.7}          & \cellcolor[HTML]{44bf70}\textcolor[HTML]{f1f1f1}{77.7}          & \cellcolor[HTML]{f6e620}\textcolor[HTML]{000000}{92.5}          & \cellcolor[HTML]{d8e219}\textcolor[HTML]{000000}{86.2}          & \cellcolor[HTML]{a5db36}\textcolor[HTML]{000000}{90.5}          & \cellcolor[HTML]{440154}\textcolor[HTML]{f1f1f1}{97.1}          & \cellcolor[HTML]{1f948c}\textcolor[HTML]{f1f1f1}{55.8}          & \cellcolor[HTML]{84d44b}\textcolor[HTML]{000000}{78.7}          \\
      500               & \cellcolor[HTML]{d2e21b}\textcolor[HTML]{000000}{63.6}          & \cellcolor[HTML]{a2da37}\textcolor[HTML]{000000}{67.5}          & \cellcolor[HTML]{7cd250}\textcolor[HTML]{000000}{79.3}          & \cellcolor[HTML]{f1e51d}\textcolor[HTML]{000000}{92.4}          & \cellcolor[HTML]{fde725}\textcolor[HTML]{000000}{\textbf{86.6}} & \cellcolor[HTML]{cde11d}\textcolor[HTML]{000000}{91.9}          & \cellcolor[HTML]{2a788e}\textcolor[HTML]{f1f1f1}{97.3}          & \cellcolor[HTML]{31b57b}\textcolor[HTML]{f1f1f1}{57.3}          & \cellcolor[HTML]{b0dd2f}\textcolor[HTML]{000000}{79.5}          \\
      600               & \cellcolor[HTML]{f1e51d}\textcolor[HTML]{000000}{64.0}          & \cellcolor[HTML]{c0df25}\textcolor[HTML]{000000}{68.1}          & \cellcolor[HTML]{84d44b}\textcolor[HTML]{000000}{79.5}          & \cellcolor[HTML]{f1e51d}\textcolor[HTML]{000000}{92.4}          & \cellcolor[HTML]{fde725}\textcolor[HTML]{000000}{\textbf{86.6}} & \cellcolor[HTML]{e7e419}\textcolor[HTML]{000000}{92.8}          & \cellcolor[HTML]{414487}\textcolor[HTML]{f1f1f1}{97.2}          & \cellcolor[HTML]{65cb5e}\textcolor[HTML]{000000}{58.5}          & \cellcolor[HTML]{c8e020}\textcolor[HTML]{000000}{79.9}          \\
      700               & \cellcolor[HTML]{f8e621}\textcolor[HTML]{000000}{64.1}          & \cellcolor[HTML]{d2e21b}\textcolor[HTML]{000000}{68.5}          & \cellcolor[HTML]{98d83e}\textcolor[HTML]{000000}{80.0}          & \cellcolor[HTML]{f1e51d}\textcolor[HTML]{000000}{92.4}          & \cellcolor[HTML]{f6e620}\textcolor[HTML]{000000}{86.5}          & \cellcolor[HTML]{eae51a}\textcolor[HTML]{000000}{92.9}          & \cellcolor[HTML]{22a884}\textcolor[HTML]{f1f1f1}{97.4}          & \cellcolor[HTML]{81d34d}\textcolor[HTML]{000000}{59.0}          & \cellcolor[HTML]{d2e21b}\textcolor[HTML]{000000}{80.1}          \\
      800               & \cellcolor[HTML]{f1e51d}\textcolor[HTML]{000000}{64.0}          & \cellcolor[HTML]{ece51b}\textcolor[HTML]{000000}{69.1}          & \cellcolor[HTML]{dae319}\textcolor[HTML]{000000}{81.5}          & \cellcolor[HTML]{f6e620}\textcolor[HTML]{000000}{92.5}          & \cellcolor[HTML]{e2e418}\textcolor[HTML]{000000}{86.3}          & \cellcolor[HTML]{f4e61e}\textcolor[HTML]{000000}{93.3}          & \cellcolor[HTML]{22a884}\textcolor[HTML]{f1f1f1}{97.4}          & \cellcolor[HTML]{a5db36}\textcolor[HTML]{000000}{59.6}          & \cellcolor[HTML]{eae51a}\textcolor[HTML]{000000}{80.5}          \\
      900               & \cellcolor[HTML]{f8e621}\textcolor[HTML]{000000}{64.1}          & \cellcolor[HTML]{f6e620}\textcolor[HTML]{000000}{69.3}          & \cellcolor[HTML]{fbe723}\textcolor[HTML]{000000}{82.3}          & \cellcolor[HTML]{fde725}\textcolor[HTML]{000000}{\textbf{92.7}} & \cellcolor[HTML]{c2df23}\textcolor[HTML]{000000}{86.0}          & \cellcolor[HTML]{fbe723}\textcolor[HTML]{000000}{93.6}          & \cellcolor[HTML]{7ad151}\textcolor[HTML]{000000}{97.5}          & \cellcolor[HTML]{cae11f}\textcolor[HTML]{000000}{60.2}          & \cellcolor[HTML]{f4e61e}\textcolor[HTML]{000000}{80.7}          \\
      1000              & \cellcolor[HTML]{fde725}\textcolor[HTML]{000000}{\textbf{64.2}} & \cellcolor[HTML]{fde725}\textcolor[HTML]{000000}{\textbf{69.5}} & \cellcolor[HTML]{fde725}\textcolor[HTML]{000000}{\textbf{82.4}} & \cellcolor[HTML]{f6e620}\textcolor[HTML]{000000}{92.5}          & \cellcolor[HTML]{f6e620}\textcolor[HTML]{000000}{86.5}          & \cellcolor[HTML]{fde725}\textcolor[HTML]{000000}{\textbf{93.7}} & \cellcolor[HTML]{fde725}\textcolor[HTML]{000000}{\textbf{97.6}} & \cellcolor[HTML]{fde725}\textcolor[HTML]{000000}{\textbf{61.1}} & \cellcolor[HTML]{fde725}\textcolor[HTML]{000000}{\textbf{80.9}} \\
      \bottomrule
    \end{tabular}
  }
\end{table}

\begin{table}[!tbp]
  \caption{
    \textbf{CLIP-ViT-B/32 with Layer-wise Concrete AdaMerging.}
    We show the performance of the merged model with increasing test-time adaptation steps on various datasets.
  }
  \label{table:clip-vit-b-32_layer_wise_concrete_adamerging_tta}
  \resizebox{\linewidth}{!}{%
    \centering
    \begin{tabular}{c|ccccccccc}
      \toprule
      \textbf{TTA Step} & \textbf{SUN397}                                                 & \textbf{Cars}                                                   & \textbf{RESISC45}                                               & \textbf{EuroSAT}                                                & \textbf{SVHN}                                                   & \textbf{GTSRB}                                                  & \textbf{MNIST}                                                  & \textbf{DTD}                                                    & \textbf{Avg.}                                                   \\
      \midrule
      0                 & \cellcolor[HTML]{440154}\textcolor[HTML]{f1f1f1}{62.2}          & \cellcolor[HTML]{440154}\textcolor[HTML]{f1f1f1}{62.8}          & \cellcolor[HTML]{440154}\textcolor[HTML]{f1f1f1}{77.2}          & \cellcolor[HTML]{440154}\textcolor[HTML]{f1f1f1}{92.9}          & \cellcolor[HTML]{440154}\textcolor[HTML]{f1f1f1}{85.8}          & \cellcolor[HTML]{440154}\textcolor[HTML]{f1f1f1}{84.3}          & \cellcolor[HTML]{440154}\textcolor[HTML]{f1f1f1}{98.3}          & \cellcolor[HTML]{440154}\textcolor[HTML]{f1f1f1}{55.0}          & \cellcolor[HTML]{440154}\textcolor[HTML]{f1f1f1}{77.3}          \\
      100               & \cellcolor[HTML]{34608d}\textcolor[HTML]{f1f1f1}{63.9}          & \cellcolor[HTML]{2b758e}\textcolor[HTML]{f1f1f1}{65.6}          & \cellcolor[HTML]{2b758e}\textcolor[HTML]{f1f1f1}{81.2}          & \cellcolor[HTML]{2eb37c}\textcolor[HTML]{f1f1f1}{94.9}          & \cellcolor[HTML]{23a983}\textcolor[HTML]{f1f1f1}{89.3}          & \cellcolor[HTML]{1fa187}\textcolor[HTML]{f1f1f1}{91.4}          & \cellcolor[HTML]{5cc863}\textcolor[HTML]{000000}{98.6}          & \cellcolor[HTML]{31678e}\textcolor[HTML]{f1f1f1}{57.9}          & \cellcolor[HTML]{25838e}\textcolor[HTML]{f1f1f1}{80.3}          \\
      200               & \cellcolor[HTML]{228c8d}\textcolor[HTML]{f1f1f1}{64.9}          & \cellcolor[HTML]{27ad81}\textcolor[HTML]{f1f1f1}{67.3}          & \cellcolor[HTML]{29af7f}\textcolor[HTML]{f1f1f1}{83.7}          & \cellcolor[HTML]{c0df25}\textcolor[HTML]{000000}{95.7}          & \cellcolor[HTML]{46c06f}\textcolor[HTML]{f1f1f1}{89.9}          & \cellcolor[HTML]{70cf57}\textcolor[HTML]{000000}{94.0}          & \cellcolor[HTML]{21918c}\textcolor[HTML]{f1f1f1}{98.5}          & \cellcolor[HTML]{1f988b}\textcolor[HTML]{f1f1f1}{59.7}          & \cellcolor[HTML]{32b67a}\textcolor[HTML]{f1f1f1}{81.7}          \\
      300               & \cellcolor[HTML]{1f998a}\textcolor[HTML]{f1f1f1}{65.2}          & \cellcolor[HTML]{48c16e}\textcolor[HTML]{f1f1f1}{67.9}          & \cellcolor[HTML]{5cc863}\textcolor[HTML]{000000}{84.9}          & \cellcolor[HTML]{a8db34}\textcolor[HTML]{000000}{95.6}          & \cellcolor[HTML]{58c765}\textcolor[HTML]{000000}{90.1}          & \cellcolor[HTML]{b0dd2f}\textcolor[HTML]{000000}{95.2}          & \cellcolor[HTML]{21918c}\textcolor[HTML]{f1f1f1}{98.5}          & \cellcolor[HTML]{46c06f}\textcolor[HTML]{f1f1f1}{61.2}          & \cellcolor[HTML]{5cc863}\textcolor[HTML]{000000}{82.3}          \\
      400               & \cellcolor[HTML]{3aba76}\textcolor[HTML]{f1f1f1}{66.0}          & \cellcolor[HTML]{6ece58}\textcolor[HTML]{000000}{68.4}          & \cellcolor[HTML]{8bd646}\textcolor[HTML]{000000}{85.7}          & \cellcolor[HTML]{eae51a}\textcolor[HTML]{000000}{95.9}          & \cellcolor[HTML]{8bd646}\textcolor[HTML]{000000}{90.6}          & \cellcolor[HTML]{cae11f}\textcolor[HTML]{000000}{95.7}          & \cellcolor[HTML]{5cc863}\textcolor[HTML]{000000}{98.6}          & \cellcolor[HTML]{77d153}\textcolor[HTML]{000000}{62.0}          & \cellcolor[HTML]{90d743}\textcolor[HTML]{000000}{82.9}          \\
      500               & \cellcolor[HTML]{5ec962}\textcolor[HTML]{000000}{66.4}          & \cellcolor[HTML]{86d549}\textcolor[HTML]{000000}{68.7}          & \cellcolor[HTML]{aadc32}\textcolor[HTML]{000000}{86.2}          & \cellcolor[HTML]{c0df25}\textcolor[HTML]{000000}{95.7}          & \cellcolor[HTML]{d2e21b}\textcolor[HTML]{000000}{91.2}          & \cellcolor[HTML]{dfe318}\textcolor[HTML]{000000}{96.1}          & \cellcolor[HTML]{5cc863}\textcolor[HTML]{000000}{98.6}          & \cellcolor[HTML]{9dd93b}\textcolor[HTML]{000000}{62.5}          & \cellcolor[HTML]{b0dd2f}\textcolor[HTML]{000000}{83.2}          \\
      600               & \cellcolor[HTML]{7cd250}\textcolor[HTML]{000000}{66.7}          & \cellcolor[HTML]{98d83e}\textcolor[HTML]{000000}{68.9}          & \cellcolor[HTML]{c0df25}\textcolor[HTML]{000000}{86.5}          & \cellcolor[HTML]{d5e21a}\textcolor[HTML]{000000}{95.8}          & \cellcolor[HTML]{eae51a}\textcolor[HTML]{000000}{91.4}          & \cellcolor[HTML]{e5e419}\textcolor[HTML]{000000}{96.2}          & \cellcolor[HTML]{5cc863}\textcolor[HTML]{000000}{98.6}          & \cellcolor[HTML]{cae11f}\textcolor[HTML]{000000}{63.1}          & \cellcolor[HTML]{c5e021}\textcolor[HTML]{000000}{83.4}          \\
      700               & \cellcolor[HTML]{93d741}\textcolor[HTML]{000000}{66.9}          & \cellcolor[HTML]{c0df25}\textcolor[HTML]{000000}{69.3}          & \cellcolor[HTML]{d2e21b}\textcolor[HTML]{000000}{86.8}          & \cellcolor[HTML]{d5e21a}\textcolor[HTML]{000000}{95.8}          & \textbf{\cellcolor[HTML]{fde725}\textcolor[HTML]{000000}{91.6}} & \cellcolor[HTML]{eae51a}\textcolor[HTML]{000000}{96.3}          & \textbf{\cellcolor[HTML]{fde725}\textcolor[HTML]{000000}{98.7}} & \cellcolor[HTML]{e2e418}\textcolor[HTML]{000000}{63.4}          & \cellcolor[HTML]{d8e219}\textcolor[HTML]{000000}{83.6}          \\
      800               & \cellcolor[HTML]{d0e11c}\textcolor[HTML]{000000}{67.4}          & \cellcolor[HTML]{d2e21b}\textcolor[HTML]{000000}{69.5}          & \cellcolor[HTML]{dfe318}\textcolor[HTML]{000000}{87.0}          & \cellcolor[HTML]{d5e21a}\textcolor[HTML]{000000}{95.8}          & \textbf{\cellcolor[HTML]{fde725}\textcolor[HTML]{000000}{91.6}} & \cellcolor[HTML]{efe51c}\textcolor[HTML]{000000}{96.4}          & \textbf{\cellcolor[HTML]{fde725}\textcolor[HTML]{000000}{98.7}} & \cellcolor[HTML]{f1e51d}\textcolor[HTML]{000000}{63.6}          & \cellcolor[HTML]{ece51b}\textcolor[HTML]{000000}{83.8}          \\
      900               & \cellcolor[HTML]{dde318}\textcolor[HTML]{000000}{67.5}          & \cellcolor[HTML]{e5e419}\textcolor[HTML]{000000}{69.7}          & \cellcolor[HTML]{e7e419}\textcolor[HTML]{000000}{87.1}          & \cellcolor[HTML]{eae51a}\textcolor[HTML]{000000}{95.9}          & \cellcolor[HTML]{eae51a}\textcolor[HTML]{000000}{91.4}          & \cellcolor[HTML]{f8e621}\textcolor[HTML]{000000}{96.6}          & \textbf{\cellcolor[HTML]{fde725}\textcolor[HTML]{000000}{98.7}} & \cellcolor[HTML]{f8e621}\textcolor[HTML]{000000}{63.7}          & \cellcolor[HTML]{ece51b}\textcolor[HTML]{000000}{83.8}          \\
      1000              & \textbf{\cellcolor[HTML]{fde725}\textcolor[HTML]{000000}{67.8}} & \textbf{\cellcolor[HTML]{fde725}\textcolor[HTML]{000000}{70.0}} & \textbf{\cellcolor[HTML]{fde725}\textcolor[HTML]{000000}{87.5}} & \textbf{\cellcolor[HTML]{fde725}\textcolor[HTML]{000000}{96.0}} & \textbf{\cellcolor[HTML]{fde725}\textcolor[HTML]{000000}{91.6}} & \textbf{\cellcolor[HTML]{fde725}\textcolor[HTML]{000000}{96.7}} & \textbf{\cellcolor[HTML]{fde725}\textcolor[HTML]{000000}{98.7}} & \textbf{\cellcolor[HTML]{fde725}\textcolor[HTML]{000000}{63.8}} & \textbf{\cellcolor[HTML]{fde725}\textcolor[HTML]{000000}{84.0}} \\
      \bottomrule
    \end{tabular}
  }
\end{table}

\begin{table}[!tbp]
  \caption{
    \textbf{Individual Performance of CLIP-ViT-B/32 with learned Concrete mask.}
    Pre-trained and fine-tuned models are used as baselines.
  }
  \label{table:individual_clip-vit-b-32_with_concrete_mask}
  \resizebox{\linewidth}{!}{%
    \centering
    \begin{tabular}{c|ccccccccc}
      \toprule
      \textbf{Model} & \textbf{SUN397} & \textbf{Cars} & \textbf{RESISC45} & \textbf{EuroSAT} & \textbf{SVHN} & \textbf{GTSRB} & \textbf{MNIST} & \textbf{DTD} & \textbf{Avg.} \\
      \midrule
      Pre-trained    & 63.2            & 59.6          & 60.2              & 45.0             & 31.6          & 32.6           & 48.3           & 44.4         & 48.1          \\
      Fine-tuned     & 75.3            & 77.7          & 96.1              & 99.9             & 97.5          & 98.7           & 99.7           & 79.4         & 90.5          \\\midrule
      Concrete TA    & 75.2            & 78.5          & 96.0              & 99.9             & 97.3          & 98.7           & 99.7           & 78.9         & 90.5          \\
      TW Concrete AM & 75.3            & 78.5          & 96.0              & 99.9             & 97.3          & 98.6           & 99.7           & 79.0         & 90.5          \\
      LW Concrete AM & 75.4            & 78.5          & 96.1              & 99.9             & 97.4          & 98.8           & 99.7           & 79.4         & 90.6          \\
      \bottomrule
    \end{tabular}
  }
\end{table}

\begin{table}[!tbp]
  \caption{
    \textbf{Individual Performance of CLIP-ViT-L/14 with learned Concrete mask.}
    Pre-trained and fine-tuned models are used as baselines.
  }
  \label{table:individual_clip-vit-l-14_with_concrete_mask}
  \resizebox{\linewidth}{!}{%
    \centering
    \begin{tabular}{c|ccccccccc}
      \toprule
      \textbf{Model} & \textbf{SUN397} & \textbf{Cars} & \textbf{RESISC45} & \textbf{EuroSAT} & \textbf{SVHN} & \textbf{GTSRB} & \textbf{MNIST} & \textbf{DTD} & \textbf{Avg.} \\
      \midrule
      Pre-trained    & 68.2            & 77.9          & 71.3              & 61.3             & 58.4          & 50.6           & 76.4           & 55.4         & 64.9          \\
      Individual     & {82.3}          & {92.4}        & {97.4}            & {99.9}           & {98.1}        & {99.2}         & {99.7}         & 84.1         & {94.1}        \\\midrule
      Concrete TA    & 82.1            & 92.5          & 97.4              & 99.9             & 98.1          & 99.3           & 99.7           & 84.4         & 94.2          \\
      LW Concrete AM & 82.3            & 92.5          & 97.3              & 99.9             & 98.1          & 99.2           & 99.7           & 84.4         & 94.2          \\
      \bottomrule
    \end{tabular}
  }
\end{table}

\begin{table}[!tb]
  \caption{\textbf{Individual Performance of LoRA fine-tuned Flan-T5-base with learned Concrete mask.}
    Pre-trained and fine-tuned models are used as baselines.
  }
  \label{table:individual_flan-t5-base_with_concrete_mask}
  \resizebox{\linewidth}{!}{%
    \centering
    \begin{tabular}{lccccccccc}
      \toprule
      \textbf{Method} & \textbf{CoLA} & \textbf{MNLI} & \textbf{MRPC} & \textbf{QNLI} & \textbf{QQP} & \textbf{RTE} & \textbf{SST2} & \textbf{STSB} & \textbf{Avg.} \\
      \midrule
      Pre-trained     & 69.1          & 56.5          & 76.2          & 88.4          & 82.1         & 80.1         & 91.2          & 62.2          & 75.7          \\
      Individual      & 69.1          & 82.7          & 85.5          & 90.9          & 84.0         & 84.4         & 92.9          & 87.4          & 84.6          \\
      \midrule
      Concrete TA     & 69.1          & 82.8          & 85.8          & 90.8          & 83.9         & 84.5         & 92.9          & 87.4          & 84.7          \\
      LW Concrete AM  & 69.1          & 82.9          & 86.0          & 90.9          & 83.9         & 84.1         & 92.9          & 87.4          & 84.6          \\
      \bottomrule
    \end{tabular}
  }
\end{table}

\begin{table}[!tb]
  \caption{\textbf{Individual Performance of LoRA fine-tuned Flan-T5-large with learned Concrete mask.}
    Pre-trained and fine-tuned models are used as baselines.
  }
  \label{table:individual_flan-t5-large_with_concrete_mask}
  \resizebox{\linewidth}{!}{%
    \centering
    \begin{tabular}{lccccccccc}
      \toprule
      \textbf{Method} & \textbf{CoLA} & \textbf{MNLI} & \textbf{MRPC} & \textbf{QNLI} & \textbf{QQP} & \textbf{RTE} & \textbf{SST2} & \textbf{STSB} & \textbf{Avg.} \\
      \midrule
      Pre-trained     & 73.7          & 56.6          & 82.4          & 91.1          & 85.5         & 85.6         & 94.3          & 87.5          & 82.1          \\
      Individual      & 80.2          & 88.5          & 89.2          & 94.4          & 87.2         & 91.7         & 95.2          & 90.9          & 89.6          \\
      \midrule
      Concrete TA     & 80.1          & 88.4          & 89.2          & 94.4          & 87.2         & 91.3         & 95.2          & 91.0          & 89.6          \\
      LW Concrete AM  & 80.2          & 88.4          & 89.5          & 94.4          & 87.2         & 91.3         & 95.2          & 91.0          & 89.6          \\
      \bottomrule
    \end{tabular}
  }
\end{table}

In Table~\ref{table:clip-vit-b-32_layer_wise_adamerging} and Table~\ref{table:clip-vit-b-32_layer_wise_concrete_adamerging_tta}, we show the performance of merged CLIP-ViT-B/32 across various downstream tasks with layer-wise AdaMerging and layer-wise Concrete AdaMerging, respectively.
We observed that the performance of both Layer-wise AdaMerging and Layer-wise Concrete AdaMerging on various downstream tasks consistently improved with the progression of test-time adaptation.
Notably, given the same number of update steps, Layer-wise Concrete AdaMerging generally outperformed Layer-wise AdaMerging.
This suggests that performing model fusion on the common subspace learned during meta-training can effectively resolve the interference among tasks.
These findings validate the effectiveness of our proposed method.
Furthermore, they highlight the potential of leveraging learned common subspaces for model fusion, especially in scenarios where task interference could pose a significant challenge.


\subsection{Individual Performance with Concrete Masks}
\label{appendix:individual_performance_with_concrete_masks}

In this subsection, we show the individual performance of task-specific models with learned Concrete masks.
To be specific, after the meta-learning phase of finding the Concrete mask, we use the Concrete mask to mask the task vector and rescale it to obtain the final task vector, subsequently, we add the final task vector to the pre-trained model and evaluate the performance.

Tables~\ref{table:individual_clip-vit-b-32_with_concrete_mask}, \ref{table:individual_clip-vit-l-14_with_concrete_mask}, \ref{table:individual_flan-t5-base_with_concrete_mask} and~\ref{table:individual_flan-t5-large_with_concrete_mask} show the individual performance of CLIP-ViT-B/32, CLIP-ViT-L/14, LoRA fine-tuned Flan-T5-base and LoRA fine-tuned Flan-T5-large with learned Concrete masks, respectively.
The Concrete masks are learned during the meta-learning phase of model fusion experiments using Concrete Task Arithmetic, task-wise Concrete AdaMerging and layer-wise Concrete AdaMerging.
We observe that the individual performance of the task-specific models with learned Concrete masks is comparable to the fine-tuned models.
This suggests that the Concrete masks learned during the meta-learning phase are effective in capturing the task-specific information.


\section{Generalization Experiment Details}
\label{appendix:generalization_experiment_details}
\begin{table}[!tbp]
  \caption{Generalization results on two unseen tasks when merging ViT-B/32 models on six tasks.}
  \label{table:generalization_results_clip-vit-b-32_2}
  \resizebox{\linewidth}{!}{
    \begin{tabular}{l|ccccccc|ccc}
      \toprule
      \multirow{2}{*}{\textbf{Method}} & \multicolumn{7}{c|}{\textbf{Seen Tasks}} & \multicolumn{3}{c}{\textbf{Unseen Tasks}}                                                                                                                                 \\
                                       & SUN397                                   & Cars                                      & GTSRB         & EuroSAT       & DTD           & MNIST         & \textbf{Avg.} & RESISC45      & SVHN          & \textbf{Avg.} \\
      \midrule
      Task Arithmetic                  & 63.8                                     & 63.9                                      & 75.2          & 87.3          & 56.6          & 95.7          & 73.8          & 52.5          & 49.9          & 51.2          \\
      Ties-Merging                     & \textbf{67.8}                            & \textbf{67.2}                             & 67.8          & 78.9          & 56.2          & 92.8          & 71.8          & \textbf{58.4} & 49.3          & 53.9          \\
      \textbf{Concrete TA}             & 66.4                                     & 65.7                                      & \textbf{90.0} & \textbf{96.4} & \textbf{57.2} & \textbf{98.1} & \textbf{79.0} & 54.3          & \textbf{58.9} & \textbf{56.6} \\
      \midrule
      AdaMerging                       & 67.1                                     & 67.8                                      & 94.8          & 94.4          & 59.6          & 98.2          & 80.3          & 50.2          & 60.9          & 55.5          \\
      AdaMerging++                     & 68.9                                     & 69.6                                      & 91.6          & 94.3          & 61.9          & 98.7          & 80.8          & \textbf{52.0} & \textbf{64.9} & \textbf{58.5} \\
      \textbf{Concrete AM}             & \textbf{69.6}                            & \textbf{71.0}                             & \textbf{97.6} & \textbf{97.3} & \textbf{68.7} & \textbf{99.0} & \textbf{83.9} & 48.1          & 62.3          & 55.2          \\
      \bottomrule
    \end{tabular}
  }
\end{table}

We also carry out generalization experiments on CLIP-ViT-B/32.
For this analysis, we designate two tasks as unseen, while the remaining six tasks are considered seen.
The experiments involve merging the CLIP-ViT-B/32 models fine-tuned on the six seen tasks, and we evaluate the performance of the merged model on both the seen and unseen tasks.
The outcomes from two independent runs of the generalization experiments are detailed in Tables~\ref{table:generalization_results_clip-vit-b-32} and \ref{table:generalization_results_clip-vit-b-32_2}.

As shown in Tables~\ref{table:generalization_results_clip-vit-b-32} and \ref{table:generalization_results_clip-vit-b-32_2}, both Concrete Task Arithmetic and Concrete AdaMerging demonstrate superior performance over a range of alternative techniques for a majority of the tasks assessed.
In the first set of experiments, Concrete Task Arithmetic outperformed other task arithmetic-based methods in seven out of eight tasks, while in both sets, Concrete AdaMerging surpassed other methods in six out of eight tasks.
On unseen tasks, Concrete Task Arithmetic significantly outperforms other task arithmetic-based methods. This indicates that the learned Concrete mask can effectively capture the shared information between different tasks.
This is helpful for transferring knowledge from seen tasks to unseen tasks.

\begin{table}[!tbp]
  \caption{Experimental settings for model generalization on ViT-B/32 models.}
  \label{table:generalization_experiment_settings}
  \resizebox{\linewidth}{!}{
    \centering
    \begin{tabular}{l|cccccc|cc}
      \toprule
      \textbf{Experiment ID}                                                  & \multicolumn{6}{c|}{\textbf{Seen Tasks}} & \multicolumn{2}{c}{\textbf{Unseen Tasks}}                                                          \\\midrule
      Experiment 1 (Table~\ref{table:generalization_results_clip-vit-b-32})   & SUN397                                   & Cars                                      & RESISC45 & DTD     & SVHN & GTSRB & MNIST    & EuroSAT \\
      Experiment 2 (Table~\ref{table:generalization_results_clip-vit-b-32_2}) & SUN397                                   & Cars                                      & GTSRB    & EuroSAT & DTD  & MNIST & RESISC45 & SVHN    \\
      \bottomrule
    \end{tabular}
  }
\end{table}

\begin{figure}[!tb]
  \begin{center}
    \begin{subfigure}[b]{0.4\textwidth}
      \centering
      \includegraphics[height=4.5cm]{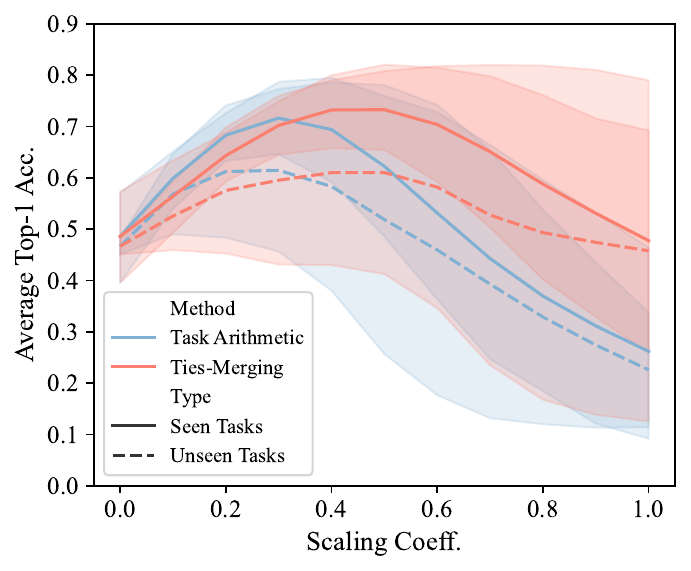}
      \caption{Experiment 1}
      \label{fig:clip-vit-b-32_generalization_exp1_task_arithmetic_and_ties_merging}
    \end{subfigure}
    \hspace{0.5cm}
    \begin{subfigure}[b]{0.4\textwidth}
      \centering
      \includegraphics[height=4.5cm]{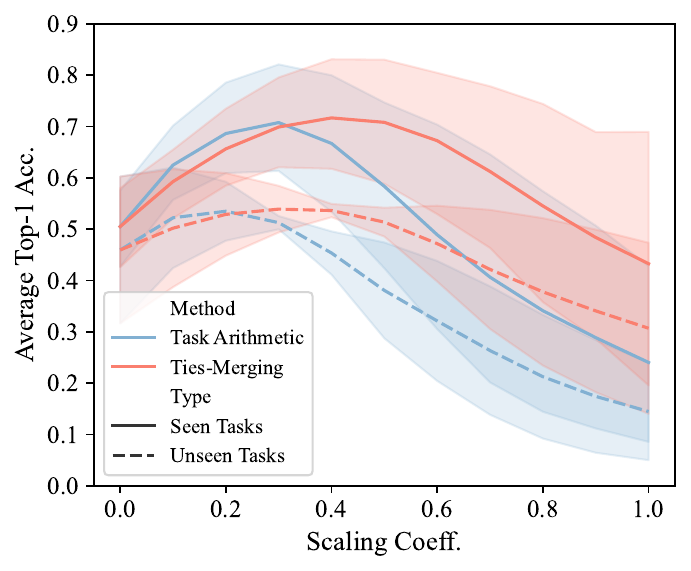}
      \caption{Experiment 2}
      \label{fig:clip-vit-b-32_generalization_exp2_task_arithmetic_and_ties_merging}
    \end{subfigure}
    \caption{
      \textbf{Generalization experiments on CLIP-ViT-B/32 using Task Arithmetic and Ties-Merging.}
      Comparison of Average Top-1 Accuracy for Task Arithmetic and Ties-Merging on both seen and unseen tasks, as a function of the scaling coefficient.
      Different line styles represent different task types (seen or unseen), and different colors represent different methods.
    }
    \label{fig:clip-vit-b-32_generalization_exp_task_arithmetic_and_ties_merging}
  \end{center}
\end{figure}

\begin{figure}[!tb]
  \begin{center}
    \begin{subfigure}[b]{0.4\textwidth}
      \centering
      \includegraphics[height=4.5cm]{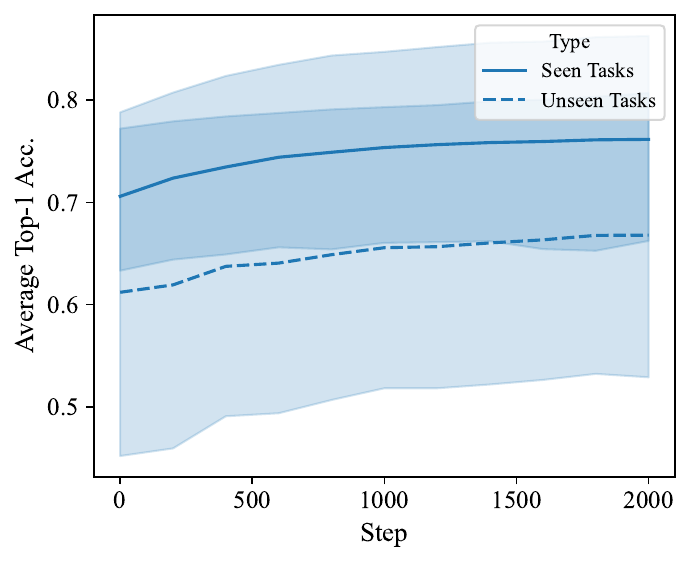}
      \caption{Experiment 1}
      \label{fig:clip-vit-b-32_generalization_exp1_concrete_task_arithmetic}
    \end{subfigure}
    \hspace{0.5cm}
    \begin{subfigure}[b]{0.4\textwidth}
      \centering
      \includegraphics[height=4.5cm]{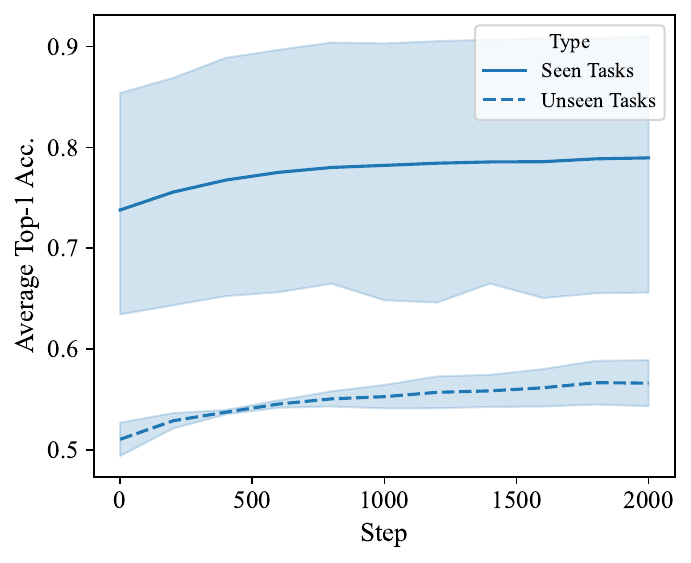}
      \caption{Experiment 2}
      \label{fig:clip-vit-b-32_generalization_exp2_concrete_task_arithmetic}
    \end{subfigure}
    \caption{
      \textbf{Generalization experiments on CLIP-ViT-B/32 using Concrete Task Arithmetic.}
      This figure illustrates the average Top-1 accuracy of seen and unseen tasks over different meta-learning steps.
      The x-axis represents the step, and the y-axis represents the average Top-1 accuracy. The line plot shows the performance of the model on seen tasks and unseen tasks (Table~\ref{table:generalization_results_clip-vit-b-32}).
    }
    \label{fig:clip-vit-b-32_generalization_exp_concrete_task_arithmetic}
  \end{center}
\end{figure}

\begin{figure}[htbp]
  \begin{center}
    \begin{subfigure}[b]{0.45\textwidth}
      \centering
      \includegraphics[height=4.5cm]{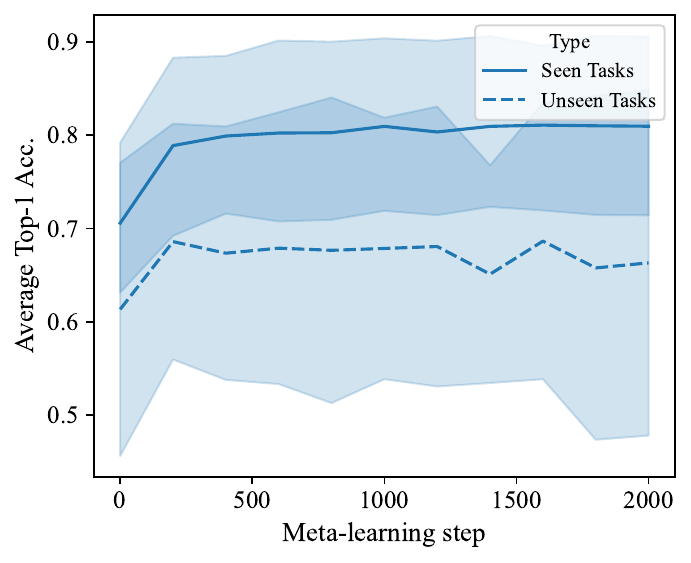}
      \caption{Experiment 1: meta-learn the Concrete mask}
      \label{clip-vit-b-32_generalization_exp1_layer_wise_concrete_adamerging}
    \end{subfigure}
    \hspace{0.5cm}
    \begin{subfigure}[b]{0.45\textwidth}
      \centering
      \includegraphics[height=4.5cm]{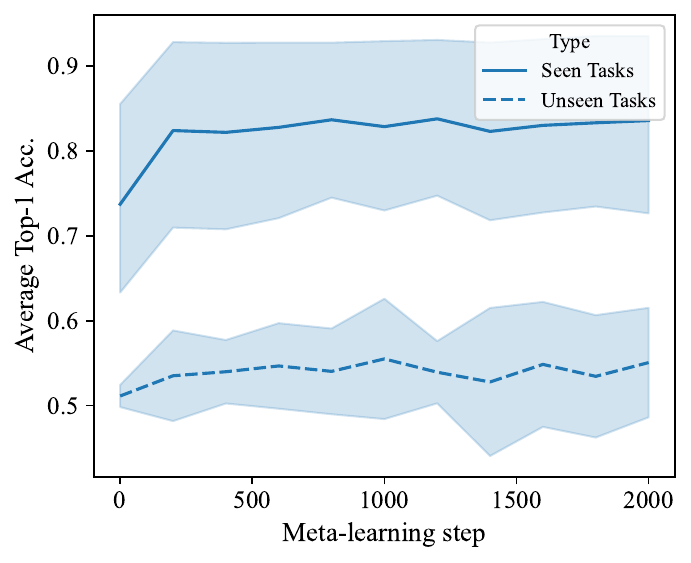}
      \caption{Experiment 2: meta-learn the Concrete mask}
      \label{clip-vit-b-32_generalization_exp2_layer_wise_concrete_adamerging}
    \end{subfigure}
    \begin{subfigure}[b]{0.45\textwidth}
      \centering
      \includegraphics[height=4.5cm]{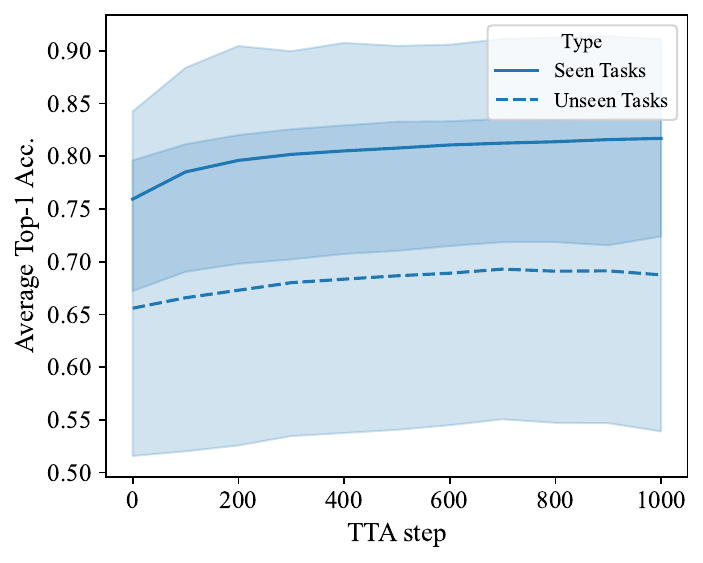}
      \caption{Experiment 1: AdaMerging TTA}
      \label{clip-vit-b-32_generalization_exp1_layer_wise_concrete_adamerging_tta}
    \end{subfigure}
    \hspace{0.5cm}
    \begin{subfigure}[b]{0.45\textwidth}
      \centering
      \includegraphics[height=4.5cm]{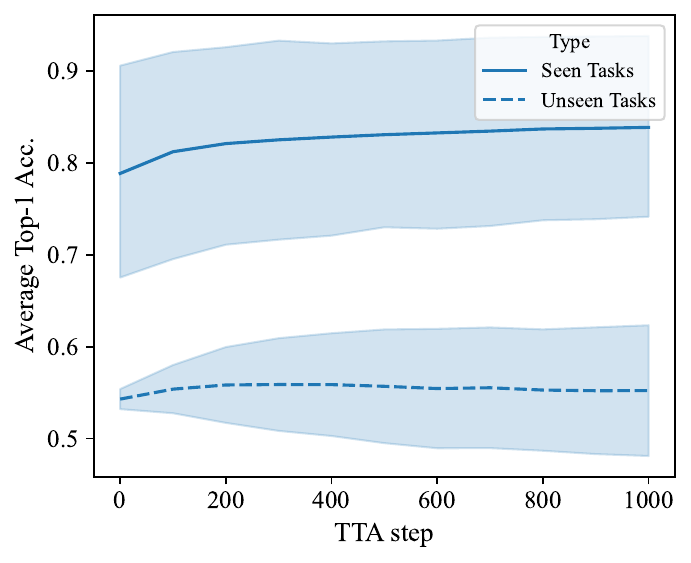}
      \caption{Experiment 2: AdaMerging TTA}
      \label{fig:clip-vit-b-32_generalization_exp2_layer_wise_concrete_adamerging_tta}
    \end{subfigure}
    \caption{
      \textbf{Generalization experiments on CLIP-ViT-B/32 using Concrete AdaMerging.}
      Figures (a) and (b): \textit{Performance comparison of seen and unseen tasks over meta-learning steps}.
      These two figures illustrate the average Top-1 accuracy of seen and unseen tasks over different meta-learning steps.
      The line plots indicate that the average performance of the model increases consistently on both seen and unseen tasks.
      Figures (c) and (d): \textit{Performance comparison of seen and unseen tasks over AdaMerging TTA Steps}.
      These two figures illustrate the average Top-1 accuracy of seen and unseen tasks over different AdaMerging Test-Time Adaptation (TTA) steps.
      The line plots indicate that the average performance of the model on seen tasks improves consistently, while the performance on unseen tasks remains largely unchanged in terms of the mean.
      However, there is an increase in variance in Experiment 2, suggesting that the performance may be better on some possible unseen tasks and worse on others.
    }
    \label{fig:clip-vit-b-32_generalization_exp_layer_wise_concrete_adamerging}
  \end{center}
\end{figure}

\textbf{Task Arithmetic and Ties-Merging.}
Figure~\ref{fig:clip-vit-b-32_generalization_exp_task_arithmetic_and_ties_merging} shows the performance of Task Arithmetic and Ties-Merging on both seen and unseen tasks, as a function of the scaling coefficient.
In Table~\ref{table:generalization_results_clip-vit-b-32}, we report the average top-1 accuracy of the merged model with the scaling coefficient set to $0.3$.
It is evident that the merged model performs optimally on both seen and unseen tasks when the scaling coefficient is within the range of $0.2$ to $0.5$.

\textbf{Concrete Task Arithmetic.}
Figure~\ref{fig:clip-vit-b-32_generalization_exp_concrete_task_arithmetic} shows the performance of Concrete Task Arithmetic on both seen and unseen tasks over different meta-learning steps.
The scaling coefficient is initialized to $0.3$ and is constant during the meta-learning process, we optimize the merged model by updating the Concrete mask.
The learning rate is set to 0.001 constantly during the meta-learning process.

As we can see, the performance of the merged model on both seen tasks and unseen tasks improves as the number of meta-learning steps increases.
This indicates that the Concrete mask can effectively capture the shared information between different tasks, and the shared information is helpful for transfering knowledge from seen tasks to unseen tasks.

\textbf{Concrete AdaMerging.}
Figure~\ref{fig:clip-vit-b-32_generalization_exp_layer_wise_concrete_adamerging} shows the performance of Concrete AdaMerging on both seen and unseen tasks over different meta-learning steps and AdaMerging test-time adaptation steps.
The steps of meta-learning the Concrete mask range from 0 to 2000, and the steps of AdaMerging test-time adaptation range from 0 to 1000.
We set the learning rate $\alpha=1, \beta=0.001$ for meta-learning the Concrete mask, and 0.001 for AdaMerging test-time adaptation.
The layer-wise weights are initialized to $0.3$ at the beginning of every meta-learning inner loop.

For both seen and unseen tasks, the model's average Top-1 accuracy appears to improve as it progresses through various meta-learning steps, implying that the model benefits from the meta-learning process.
In contrast, when AdaMerging is applied, the performance on unseen tasks does not exhibit a clear upward trend. Instead, the mean performance level remains relatively stable.
Notably, in Experiment 2, there is an observed increase in performance variance during the AdaMerging test-time adaptation steps.
This increasing variance suggests that while some unseen tasks may benefit from the methods applied, resulting in improved performance, other tasks might be negatively impacted, exhibiting decreased performance.

\textbf{Discussion about the Concrete methods.}
Based on the observations from Figures~\ref{fig:clip-vit-b-32_generalization_exp_concrete_task_arithmetic} and \ref{fig:clip-vit-b-32_generalization_exp_layer_wise_concrete_adamerging}, it can be concluded that the Concrete Task Arithmetic demonstrates good generalization capabilities on unseen tasks. However, the impact of the Concrete AdaMerging technique on seen and unseen tasks varies.
Seen tasks benefit significantly from both the meta-learning approach and the AdaMerging test-time adaptation steps, resulting in consistent improvements in model performance. Conversely, for unseen tasks, although the average performance does not show a decline, the results are more heterogeneous.
In summary, these findings suggest that our learned Concrete masks generalize more stably to unseen tasks than learned layer-wise weights.


\begin{figure}[tbp]
  \captionsetup[subfigure]{font=footnotesize}
  \centering
    \begin{subfigure}[b]{0.22\linewidth}
      \centering
      \includegraphics[width=2.4cm,height=2cm]{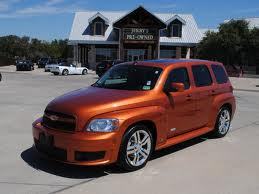}
      \caption{Clean}
    \end{subfigure}
    \begin{subfigure}[b]{0.22\linewidth}
      \centering
      \includegraphics[width=2.4cm,height=2cm]{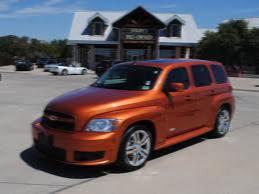}
      \caption{Motion Blur}
    \end{subfigure}
    \begin{subfigure}[b]{0.22\linewidth}
      \centering
      \includegraphics[width=2.4cm,height=2cm]{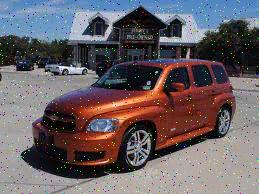}
      \caption{Impulse Noise}
    \end{subfigure}
    \begin{subfigure}[b]{0.22\linewidth}
      \centering
      \includegraphics[width=2.4cm,height=2cm]{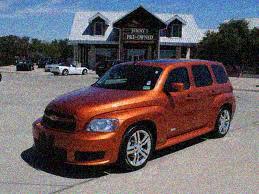}
      \caption{Gaussian Noise}
    \end{subfigure}
    \\
    \begin{subfigure}[b]{0.22\linewidth}
      \centering
      \includegraphics[width=2.4cm,height=2cm]{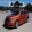}
      \caption{Pixelate}
    \end{subfigure}
    \begin{subfigure}[b]{0.22\linewidth}
      \centering
      \includegraphics[width=2.4cm,height=2cm]{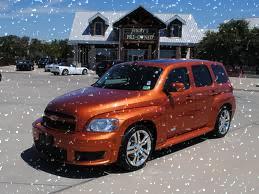}
      \caption{Spatter}
    \end{subfigure}
    \begin{subfigure}[b]{0.22\linewidth}
      \centering
      \includegraphics[width=2.4cm,height=2cm]{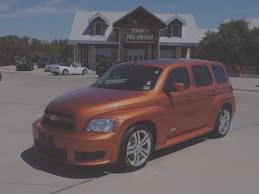}
      \caption{Contrast}
    \end{subfigure}
    \begin{subfigure}[b]{0.22\linewidth}
      \centering
      \includegraphics[width=2.4cm,height=2cm]{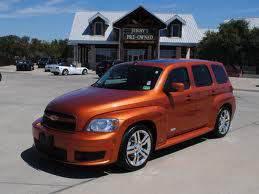}
      \caption{JPEG}
    \end{subfigure}
    \caption{
      Example images corrupted by eight distortion types for out-of-distribution evaluation.
    }
    \label{fig:distorted_images}
\end{figure}

\begin{table}[htbp]
  \caption{Ablations of the test data distribution on ViT-B/16.}
  \label{table:ablation_data_distribution_vit_b_16}
  \resizebox{\linewidth}{!}{%
    \centering
    \begin{tabular}{l|cccc|cccc}
      \toprule
      \textbf{Method}      & \textbf{Cars}                                           & \textbf{EuroSAT}                                          & \textbf{RESISC45} & \textbf{GTSRB} & \textbf{Cars} & \textbf{EuroSAT} & \textbf{RESISC45} & \textbf{GTSRB} \\
      \midrule
                           & \multicolumn{4}{c|}{{Clean Test Set}}                   & \multicolumn{4}{c}{{Corrupted Test Set (Motion Blur)}}                                                                                                                 \\
      Task Arithmetic      & 75.3                                                    & 96.3                                                      & 85.3              & 80.5           & 73.5          & 70.9             & 83.9              & 72.2           \\
      \textbf{Concrete TA} & \textbf{77.2}                                           & \textbf{97.0}                                             & \textbf{86.7}     & \textbf{84.9}  & \textbf{75.5} & \textbf{73.6}    & \textbf{85.3}     & \textbf{75.4}  \\
      AdaMerging           & 83.4                                                    & 97.2                                                      & 88.6              & 97.5           & 81.3          & \textbf{75.9}    & 87.4              & 95.6           \\
      Concrete AdaMerging  & 84.8                                                    & \textbf{97.8}                                             & \textbf{91.2}     & \textbf{97.9}  & \textbf{83.3} & 75.3             & \textbf{89.1}     & \textbf{95.7}  \\
      \midrule
                           & \multicolumn{4}{c|}{Corrupted Test Set (Impulse Noise)} & \multicolumn{4}{c}{Corrupted Test Set (Gaussian Noise)}                                                                                                                \\
      Task Arithmetic      & 70.4                                                    & \textbf{59.5}                                             & 75.2              & \textbf{54.0}  & 72.2          & \textbf{60.8}    & 78.5              & \textbf{51.0}  \\
      \textbf{Concrete TA} & \textbf{72.8}                                           & 44.9                                                      & \textbf{77.7}     & 49.6           & \textbf{74.5} & 54.6             & \textbf{79.8}     & 47.8           \\
      AdaMerging           & 77.6                                                    & \textbf{42.1}                                             & 81.9              & \textbf{90.2}  & 79.1          & \textbf{58.9}    & 81.2              & \textbf{74.5}  \\
      Concrete AdaMerging  & \textbf{81.1}                                           & 27.4                                                      & \textbf{87.5}     & 72.8           & \textbf{82.6} & 55.2             & \textbf{85.9}     & 64.3           \\
      \midrule
                           & \multicolumn{4}{c|}{Corrupted Test Set (Pixelate)}      & \multicolumn{4}{c}{Corrupted Test Set (Spatter)}                                                                                                                       \\
      Task Arithmetic      & \textbf{3.8}                                            & 38.0                                                      & 24.8              & 71.3           & 72.1          & \textbf{58.4}    & 79.9              & \textbf{60.1}  \\
      \textbf{Concrete TA} & 2.3                                                     & \textbf{38.4}                                             & 24.8              & \textbf{74.4}  & \textbf{73.8} & 48.3             & \textbf{80.9}     & 54.8           \\
      AdaMerging           & \textbf{4.1}                                            & \textbf{46.4}                                             & \textbf{23.6}     & 91.3           & 79.3          & \textbf{60.9}    & 85.8              & \textbf{93.7}  \\
      Concrete AdaMerging  & 0.9                                                     & 39.9                                                      & 16.3              & \textbf{92.3}  & \textbf{80.9} & 52.8             & \textbf{87.3}     & 92.8           \\
      \midrule
                           & \multicolumn{4}{c|}{Corrupted Test Set (Contrast)}      & \multicolumn{4}{c}{Corrupted Test Set (JPEG Compression)}                                                                                                              \\
      Task Arithmetic      & 73.4                                                    & 62.5                                                      & 81.3              & 76.9           & 75.1          & 73.1             & 84.8              & 64.7           \\
      \textbf{Concrete TA} & \textbf{75.3}                                           & \textbf{66.4}                                             & \textbf{82.8}     & \textbf{79.6}  & \textbf{76.7} & \textbf{75.8}    & \textbf{85.8}     & \textbf{65.4}  \\
      AdaMerging           & 81.4                                                    & 68.1                                                      & 85.8              & 96.8           & 81.9          & 76.0             & 87.3              & \textbf{91.0}  \\
      Concrete AdaMerging  & \textbf{83.5}                                           & \textbf{70.9}                                             & \textbf{88.9}     & \textbf{97.0}  & \textbf{83.7} & \textbf{77.1}    & \textbf{89.8}     & 89.8           \\
      \bottomrule
    \end{tabular}
  }
\end{table}

\section{Test Data Distribution Ablations}
\label{appendix:test_data_ablations}

We carry out a range of ablation studies on the data distribution to evaluate the robustness of our approach when dealing with out-of-distribution (OOD) test data. In numerous practical situations, the test data to which we aim to generalize might not come from the same distribution as the training data. This difference can create substantial obstacles to performance and reliability.

In order to assess the robustness of our method and its ability to perform under various conditions, we intentionally introduce eight different kinds of noise into the test data.
These specific types of noise were selected to cover a wide spectrum of potential distortions that the model could potentially face in real-world scenarios.
By deliberately corrupting the test data, we are able to gauge the capability of our method to manage and adjust to such distortions. This gives us a more thorough understanding of how our method performs when the distribution of test data differs from that of the training data.
Figure~\ref{fig:distorted_images} offers visual illustrations of the images that have been distorted as a result of this process. This visual representation aids in understanding the extent and nature of the distortions applied to the test data, providing a clearer picture of the challenging conditions under which the model is expected to perform.

We set the learning rate to $3\times 10^{-4}$ and employ the Adam optimizer for test-time adaptation on the test dataset, following the details outlined in Section~\ref{subsec:concrete_subspace_learning} of the main paper. The model undergoes training for 2000 steps, and we present the performance results across all tasks in Tables~\ref{table:ablation_data_distribution_vit_b_32} and \ref{table:ablation_data_distribution_vit_b_16}.
The two tables present the results of ablation studies on the test data distribution for two models: ViT-B/32 and ViT-B/16.
It can be observed that our method surpasses Task Arithmetic in four out of eight types of corruption. In the remaining four types of corruption, our method is on par with Task Arithmetic.
This comparative superiority demonstrates the robustness of our method and its potential to be a reliable solution in practical applications where data may not always be clean or come from the same distribution as the training set.


\section{Implementation Details}

In this section, we provide more details about the implementation of our method.

\subsection{Preprocessed Examples of GLUE Benchmark}
\label{appendix:preprocessed_examples_of_glue_benchmark}

Flan-T5 models embody the encoder-decoder architecture specific to Transformer models, operating within a text-to-text framework. 
Therefore, we reformat the initial inputs into a structure that aligns with this text-to-text paradigm. 
In this subsection, we present a range of instances demonstrating how the data has been preprocessed to fit 
the requirements of the GLUE (General Language Understanding Evaluation) benchmark~\cite{wangGLUEMultiTaskBenchmark2018}. 

We report exact match accuracy for all tasks except for STSB, where we report Spearman's $\rho$.

\subsubsection{CoLA}

We assign label 0 as ``\textit{unacceptable}'' and label 1 as ``\textit{acceptable}''.

Original inputs:
\begin{itemize}
  \item \textit{sentence:} Our friends won't buy this analysis, let alone the next one we propose.
  \item \textit{label:} 1
\end{itemize}

Preprocessed:
\begin{itemize}
  \item \textit{input:} Indicate if the following sentence is grammatically correct or not: 
  ``\textbf{Our friends won't buy this analysis, let alone the next one we propose.}''. 
    Answer `acceptable' or `unacceptable'.
  \item \textit{target:} acceptable
\end{itemize}

\subsubsection{MNLI}

We assign label 0 as ``\textit{entailment}'', label 1 as ``\textit{neutral}'' and label 2 as ``\textit{contradiction}''.

Original inputs:
\begin{itemize}
  \item \textit{hypothesis:} Product and geography are what make cream skimming work.
  \item \textit{premise:} Conceptually cream skimming has two basic dimensions - product and geography.
  \item \textit{label:} 1
\end{itemize}

Preprocessed:
\begin{itemize}
  \item \textit{input:} Does the premise: `\textbf{Product and geography are what make cream skimming work.}' logically imply, contradict, 
  or is neutral to the hypothesis: `\textbf{Conceptually cream skimming has two basic dimensions - product and geography.}'? 
  Answer with `entailment', `contradiction', or `neutral'.
  \item \textit{target:} neutral
\end{itemize}

\subsubsection{MRPC}

We assign label 0 as ``\textit{no}'' and label 1 as ``\textit{yes}''.

Original inputs:
\begin{itemize}
  \item \textit{sentence1:} Amrozi accused his brother, whom he called ``the witness'', of deliberately distorting his evidence.
  \item \textit{sentence2:} Referring to him as only ``the witness'', Amrozi accused his brother of deliberately distorting his evidence.
  \item \textit{label:} 1
\end{itemize}

Preprocessed:
\begin{itemize}
  \item \textit{input:} Are the following sentences `\textbf{Amrozi accused his brother, whom he called ``the witness'', of deliberately distorting his evidence.}' 
  and `\textbf{Referring to him as only ``the witness'', Amrozi accused his brother of deliberately distorting his evidence.}' conveying the same meaning? Answer with `yes' or `no'.
  \item \textit{target:} yes
\end{itemize}

\subsubsection{QNLI}

We assign label 0 as ``\textit{yes}'' and label 1 as ``\textit{no}''.

Original inputs:
\begin{itemize}
  \item \textit{question:} What kind of test does the doctor perform?
  \item \textit{sentence:} The doctor performs a test to see if you have strep throat.
  \item \textit{label:} 1
\end{itemize}

Preprocessed:
\begin{itemize}
  \item \textit{input:} Given the context: `\textbf{The doctor performs a test to see if you have strep throat.}', 
  does the question `\textbf{What kind of test does the doctor perform?}' have an answer based on the information provided? Answer with `yes' or `no'.
  \item \textit{target:} yes
\end{itemize}

\subsubsection{QQP}

We assign label 0 as ``\textit{no}'' and label 1 as ``\textit{yes}''.

Original inputs:
\begin{itemize}
  \item \textit{question1:} How is the life of a math student? Could you describe your own experiences?
  \item \textit{question2:} Which level of preparation is enough for the exam jlpt5?
  \item \textit{label:} 0
\end{itemize}

Preprocessed:
\begin{itemize}
  \item \textit{input:} Do the questions `\textbf{How is the life of a math student? Could you describe your own experiences?}' 
  and `\textbf{Which level of preparation is enough for the exam jlpt5?}' have the same intent? 
  Answer with `yes' or `no'.
  \item \textit{target:} no
\end{itemize}

\subsubsection{RTE}

We assign label 0 as ``yes'' and label 1 as ``no''.

Original inputs:
\begin{itemize}
  \item \textit{sentence1:} No Weapons of Mass Destruction Found in Iraq Yet.
  \item \textit{sentence2:} Weapons of Mass Destruction Found in Iraq.
  \item \textit{label:} 1
\end{itemize}

Preprocessed:
\begin{itemize}
  \item \textit{input:} Does the text: `\textbf{No Weapons of Mass Destruction Found in Iraq Yet.}' 
  entail that `\textbf{Weapons of Mass Destruction Found in Iraq.}' is true? 
  Provide `yes' or `no'.
  \item \textit{target:} no
\end{itemize}

\subsubsection{SST2}

We assign label 0 as ``\textit{negative}'' and label 1 as ``\textit{positive}''.

Original inputs:
\begin{itemize}
  \item \textit{sentence:} hide new secretions from the parental units
  \item \textit{label:} 0
\end{itemize}

Preprocessed:
\begin{itemize}
  \item \textit{input:} Given the sentence `\textbf{hide new secretions from the parental units}', determine the sentiment. Is it positive or negative?
  \item \textit{target:} negative
\end{itemize}

\subsubsection{STSB}

We format the `\texttt{label}' argument as a string with one decimal place, the corresponding Python code is `\texttt{"\{:.1f\}".format(label)}`.

Original inputs:
\begin{itemize}
  \item \textit{sentence1:} A plane is taking off.
  \item \textit{sentence2:} An air plane is taking off.
  \item \textit{label:} 5
\end{itemize}

Preprocessed:
\begin{itemize}
  \item \textit{input:} Consider the sentences `\textbf{A plane is taking off.}' and `\textbf{An air plane is taking off.}'. 
  On a scale from 1 (completely different) to 5 (completely similar), rate the similarity.
  \item \textit{target:} 5.0
\end{itemize}

\end{document}